\documentclass{article}

\PassOptionsToPackage{numbers, compress}{natbib} 

\usepackage[final]{neurips_2024}

\usepackage[utf8]{inputenc} 
\usepackage[T1]{fontenc}    
\usepackage{hyperref}       
\usepackage{url}            
\usepackage{booktabs}       
\usepackage{amsfonts}       
\usepackage{nicefrac}       
\usepackage{microtype}      
\usepackage{xcolor}         

\usepackage{verbatim}
\usepackage{graphicx}
\usepackage{multirow}
\usepackage{algorithm}
\usepackage{algpseudocode}
\usepackage{amsthm}
\usepackage[most]{tcolorbox}
\usepackage{subcaption} 
\usepackage{tabularx} 
\usepackage{adjustbox}
\usepackage{multicol, multirow, makecell}
\usepackage{wrapfig}
\usepackage{xspace}

\definecolor{promptcolor}{RGB}{0,0,255} 
\definecolor{responsecolor}{RGB}{0,128,0} 

\newcommand{\x}{\mathbf{x}}

\newcommand{\ba}{\mathbf{a}}
\newcommand{\K}{\mathbf{K}} 
\newcommand{\Khat}{\mathbf{\hat{K}}} 
\newcommand{\bk}{\mathbf{k}} 
\newcommand{\bkhat}{\mathbf{\hat{k}}} 

\newcommand{\q}{\mathbf{q}}
\newcommand{\qhat}{\mathbf{\hat{q}}}
\newcommand{\V}{\mathbf{V}}

\newcommand{\bv}{\mathbf{v}}
\newcommand{\bP}{\mathbf{P}}
\newcommand{\bR}{\mathbf{R}}
\newcommand{\bL}{\mathbf{L}}
\newcommand{\bigO}{\mathcal{O}}
\newcommand{\method}{Loki\xspace}
\newtheorem{theorem}{Theorem}[section]

\newtheorem{lemma}[theorem]{Lemma}

\newcommand{\prompt}[1]{%
    \begin{tcolorbox}[colframe=promptcolor, colback=promptcolor!10, sharp corners, boxrule=0.5pt, arc=2mm]
        \begin{tcolorbox}[colframe=promptcolor, colback=white, rounded corners, boxrule=0.5pt, arc=2mm]
            \centering\small\textbf{Prompt}
        \end{tcolorbox}
        #1
    \end{tcolorbox}
}

\newcommand{\response}[2]{%
    \begin{tcolorbox}[colframe=responsecolor, colback=responsecolor!10, sharp corners, boxrule=0.5pt, arc=2mm]
        \begin{tcolorbox}[colframe=responsecolor, colback=white, rounded corners, boxrule=0.5pt, arc=2mm]
            \centering\small\textbf{#2}
        \end{tcolorbox}
        #1
    \end{tcolorbox}
}

\newtcolorbox{researchquestion}{
    colback=blue!5!white,
    colframe=blue!75!black,
    boxrule=0mm,
    arc=0mm,
    left=10pt,
    right=10pt,
    boxsep=1pt,
    toptitle=1pt,
    bottomtitle=1pt,
}

\title{Loki: Low-rank Keys for Efficient Sparse Attention}

 \author{%
   Prajwal Singhania,
   Siddharth Singh,
   Shwai He,
   Soheil Feizi,
   Abhinav Bhatele\\
   \AND
   \vspace{-0.18in}\\
   Department of Computer Science, University of Maryland\\
   College Park, MD 20742\\
   \texttt{prajwal@umd.edu},
   \texttt{bhatele@cs.umd.edu}
 }

\begin{document}

\maketitle

\begin{abstract}
Inference on large language models (LLMs) can be expensive in terms of the
compute and memory costs involved, especially when long sequence lengths are
used. In particular, the self-attention mechanism used in LLM inference
contributes significantly to these costs, which has sparked an interest in
approximating the self-attention computation to reduce such costs. In this
work, we propose to approximate self-attention by focusing on the
dimensionality of {\em key} vectors computed in the attention block. Our
analysis reveals that key vectors lie in a significantly lower-dimensional
space, consistently across several datasets and models. Exploiting this
observation, we propose \emph{\method}, a novel sparse attention method that
ranks and selects tokens in the KV-cache based on attention scores computed in
low-dimensional space. Our evaluations show that \method is able to speed up
the attention computation due to reduced data movement (load/store) and compute
costs while maintaining the efficacy of the models better than other popular
approximation methods.

\end{abstract}

\section{Introduction}
\label{sec:intro}
As large language models (LLMs) grow in size, deploying them for efficient
inference presents substantial challenges, largely due to computation and
memory access bottlenecks in the self-attention block~\citep{transformer},
especially when handling long sequences. These challenges stem from the
autoregressive nature of attention, which generates the output one token at a
time. At each step, the entire preceding state, stored in the key-value (KV)
cache, must be fetched from memory, which can sometimes exceed the size of the
model parameters itself~\citep{woosuk2023vllm}. This frequent KV-cache access
from GPU DRAM to registers becomes costly, as it scales quadratically with the
output sequence length. In addition, matrix multiplications in the
attention layers also have a quadratic scaling cost with sequence length,
compounding the overall computational burden.

Several strategies~\citep{zhang2023hH2O, ribar2023sparq, liu2023scissorhands}
have been proposed to address this challenge by reducing the computational
complexity and/or memory demands associated with the self-attention mechanism.
One promising category of approaches focuses on approximating attention,
employing techniques such as quantization or using a subset of the tokens in
the KV-cache~\citep{ge2024model} (sparse attention).

\begin{figure}[t]
  \centering
  \includegraphics[height=1.75in]{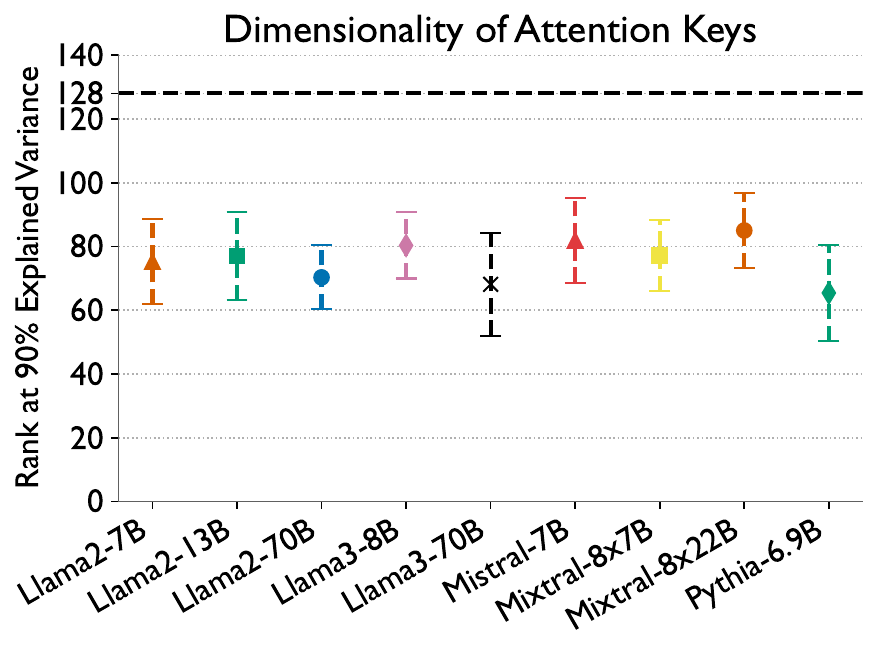}
  \hfill
  \includegraphics[height=1.75in]{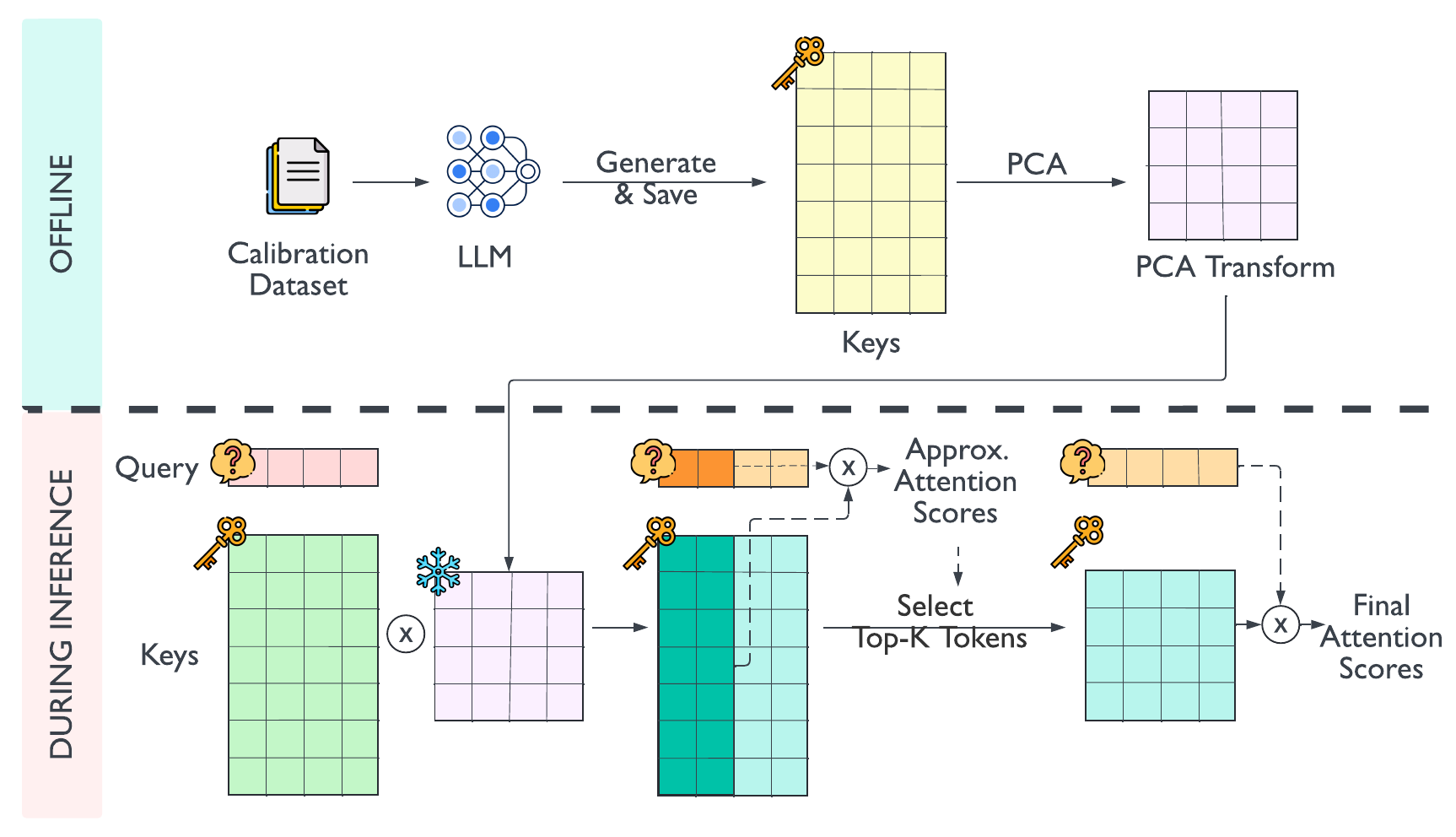}
  \caption{Rank at which 90\% of the variance is explained, averaged across all
layers and heads for different models. Full rank is represented by the black
dashed line (left). Overview of \method (right).}
  \label{fig:overview}
\end{figure}

In contrast to other sparse attention approaches that either permanently prune
tokens from the key-value cache~\citep{zhang2023hH2O} or impose a fixed
sparsity pattern~\citep{xiao2023efficient}, our proposed method dynamically
selects key tokens at each generation step based on approximate attention
scores and avoids deletions.  This approach is inspired by a critical
observation: across a range of LLMs and datasets, key tensors consistently
occupy a significantly lower-dimensional space than the full attention head
dimension. For instance, in Figure~\ref{fig:overview} (left), we show that
across various LLMs~\citep{dubey2024llama3herdmodels, jiang2024mixtral}, 90\%
of the variance explained by PCA is captured at an effective key vector rank of
around 80, despite the key tensor dimension being much larger (128).

Based on this observation, we introduce \method, a sparse attention method that
leverages the low-dimensional structure of key vectors to reduce data movement
and computation costs without significantly impacting model quality. First, we
apply PCA to keys generated from a calibration dataset, storing all principal
components but using only the top $d$ (25-50\%) to compute approximate
attention scores during inference. This dimensionality reduction, informed by
our previous observation that key vectors have low effective rank, allows us to
efficiently identify the top-$k$ (12.5-25\%) most relevant tokens using the
approximate scores. For these selected keys, we then revert to the full
dimensionality to compute the final attention scores, ensuring both efficiency
and accuracy. Figure~\ref{fig:overview} (right) illustrates our approach.

Our theoretical complexity analysis demonstrates that \method can provide
significant speedups in the attention step. However, actually realizing these
gains requires an efficient implementation of our method to minimize data
movement in the additional operations introduced on top of the original self
attention algorithm. Thus, we implement optimized sparse matrix multiplication kernels
for \method~in Triton, leading to a speedup of up to 45\% over the 
standard HuggingFace Transformer's~\citep{wolf-etal-2020-transformers} attention 
implementation (\emph{vanilla} attention) for Llama2-13B.
For this setting, the average degradation in model
accuracy (measured across 6 different benchmarks and 8 different models) is
only 6.8\%.

Our contributions can be summarized as follows:
\vspace{-0.1in}
\begin{itemize}
    \item Detailed analysis showing the intrinsic low-dimensionality of keys in
self-attention, its variation across layers for different models, and
consistency across different datasets.
    \item \method: a sparse attention method that exploits the aforementioned
low dimensionality of keys to make the attention computation faster without sacrificing
model quality.
    \item Optimized kernels for efficient implementation of \method~in PyTorch.
    \item Evaluation of
\method\footnote{\url{https://github.com/hpcgroup/loki}}~on multiple LLMs and
downstream tasks, showing that it can achieve significant speedups with minimal
degradation in model quality.
\end{itemize} 

\section{Background and Related Work}
\label{sec:bg}
The attention mechanism~\citep{transformer} is at the core of the transformer
architecture. Consider a single attention query head with head dimension
$D$, processing an input token sequence of length $S$. During auto-regressive
generation, the output of the attention head is calculated as:
\begin{equation}
    \label{eq:attention}
    \boldsymbol{y} = \mathrm{softmax}\Bigl(\dfrac{\q\K^\top}{\sqrt{D}}\Bigr) \cdot \boldsymbol{V}
\end{equation}
where $\q \in \mathbb{R}^{1\times D}$ is the query, and $\K \in \mathbb{R}^{S
\times D}$ and $\V \in \mathbb{R}^{S \times D}$ are the key and value caches
respectively. Additionally, newer transformer models add Rotary Position
Embeddings (RoPE)~\citep{su2023roformer} to the keys and query, before
computing the attention scores.  Since every query attends to all past keys,
the mechanism has a quadratic complexity $\bigO(S^2)$ in number of input $+$
generated tokens.

\subsection{Related Work}

Numerous studies have explored the low-rank structures in transformers for
various purposes. Linformer~\citep{wang2020linformer} demonstrated that the
attention score matrix is low-rank and proposed alternative low-rank attention
formulations during training for linear computational complexity.
LoRA~\citep{lora} showed that parameter updates to a transformer model during
fine-tuning reside in a low-dimensional subspace. To the best of our knowledge,
our work is the first to study the intrinsic low dimensionality of the
attention keys themselves and demonstrate the generalizability of this
low-dimensional structure across different models (for natural language data).

Sparse-transformers~\citep{child2019generating} was one of the first works to
introduce a sparse-attention method employing strided sparsity patterns in the
attention mechanism. Reformer~\citep{kitaev2020reformer} used locally-sensitive
hashing to compute attention scores in a sparse manner.
Performer~\citep{choromanski2022rethinking} used positive orthogonal random
features to approximate the attention mechanism. Unlike these methods, which
require training or fine-tuning, our approach operates entirely post-training
without any fine-tuning.

Another category of sparse attention methods employ token eviction policies to
permanently delete tokens from the KV-cache based on some heuristic.
StreamingLLM~\citep{xiao2023efficient} uses initial tokens and a rolling
KV-cache for processing infinite-length sequences. Zhang et
al.~\citep{zhang2023hH2O} retain only "Heavy Hitters" tokens in the KV-cache
based on accumulated attention scores. Scissorhands~\citep{liu2023scissorhands}
prioritizes important tokens based on the "Persistence of Importance
Hypothesis". Ge et al.~\citep{ge2023model} propose an adaptive eviction policy
for each transformer layer. These methods are effective in reducing the memory
and compute footprint of the attention but suffer from permanent loss of
information leading to a non-trivial degradation in model quality. Our method
does not involve any permanent loss of information with the trade-off of not
reducing the memory footprint. Quantization-based approximate
approaches~\citep{jacob2017quantization, nagel2021white} are complementary to
our work and can be applied in tandem.

SparQ Attention~\citep{ribar2023sparq} is a recent work that inspires our
approach. They use high-magnitude query dimensions and corresponding key
dimensions for approximate attention scoring, followed by computing the full
attention scores for the top-$k$ keys. However, their method requires costly
non-contiguous column indexing of the key vectors. Further, they store two
copies of the past keys for efficiency, increasing memory use by 50\%. In
contrast, \method~avoids the extra memory and leverages the natural ordering of
principal components, allowing for a more efficient slicing operation.

A concurrent work, InfiniGen~\citep{lee2024infinigen}, accelerates attention by
pre-fetching top-$k$ keys from CPU to GPU memory, using SVD-based low-rank
approximation of the attention scores. While their low-rank approximation is
similar to ~\method, our work provides deeper analysis of the intrinsic
low-rank structure of attention keys and focuses on speeding up attention
computation without CPU offloading. Importantly, their results affirm the
benefits of the low-dimensional nature of attention keys applied in other
contexts.

\section{Dimensionality Analysis of Attention Keys}
\label{sec:dimanalysis}
As noted in Section~\ref{sec:intro}, \method, our proposed method for
sparse self-attention, is based on the observation that key tensors
consistently reside in a lower-dimensional space than the full attention head
dimension suggests. Here, we present empirical evidence supporting this claim
by performing PCA on the keys generated in several language models and
datasets. 

\subsection{Models and Datasets Used}

To investigate the dimensionality of attention keys, we run 11
transformer-based models: Llama-2 7B/13B/70B~\citep{touvron2023llama}, Llama-3
8B/70B~\citep{dubey2024llama3herdmodels},
TinyLlama-1.1B~\citep{zhang2024tinyllama},
Pythia-6.9B~\citep{biderman2023pythia}, Mistral-7B~\citep{jiang2023mistral},
Mixtral-8x7B/8x22B~\citep{jiang2024mixtral}, and
Phi3-Mini-4K~\citep{microsoft2024phi3} on three popular English language
datasets: WikiText-2~\citep{wikitext-103} (Validation Split),
C4~\citep{raffel2023exploring} (Custom Split), and
BookCorpus~\citep{zhu2015aligning} (Custom Split).  Custom splits are used for
datasets where the validation split is not available.  We run perplexity
evaluation on these datasets and save the generated attention keys, before and
after the application of rotary embeddings~\citep{su2023roformer}, referred to
as \emph{pre-rotary} and \emph{post-rotary} keys, respectively throughout the
paper. We then perform PCA on all the keys generated for each layer and head
individually. 

The metric we use in our analysis is the rank at which $v$\% of the variance is
explained by the principal components. We calculate this metric for each layer
and head of the models as follows:
\begin{equation}
    Rank_{l,h}@v = \min \left\{ d \in \mathbb{Z}^+ : \sum_{j=1}^{d} \lambda_{l,h}^j \geq v/100 \right\}
\end{equation}
where, $\lambda_{l,h}^j$ is the $j^{th}$ normalized eigenvalue of the
covariance matrix of the keys for layer, $l$ and head, $h$. We
average this metric ranks across all heads of layer, $l$ and refer to
it as $Rank_{l}@v$.

\subsection{Findings and Discussion}

Figure \ref{fig:overview} (left) shows the average $Rank_{l}@90$ across all
layers for models with full key dimensionality of 128. We can see that the
average rank is significantly lower than the full dimensionality of the keys
for all models. Diving deeper, we present a layer-wise analysis for a few
models: Llama2-7B, Llama3-70B, Mixtral-8x7B, and Phi3-Mini-4K in
Figure~\ref{fig:rank90}. The results for the other models are similar and can
be found in Appendix \ref{appendix_dimanalysis}. 

\begin{figure}[h]
  \centering
    \includegraphics[width=0.98\textwidth]{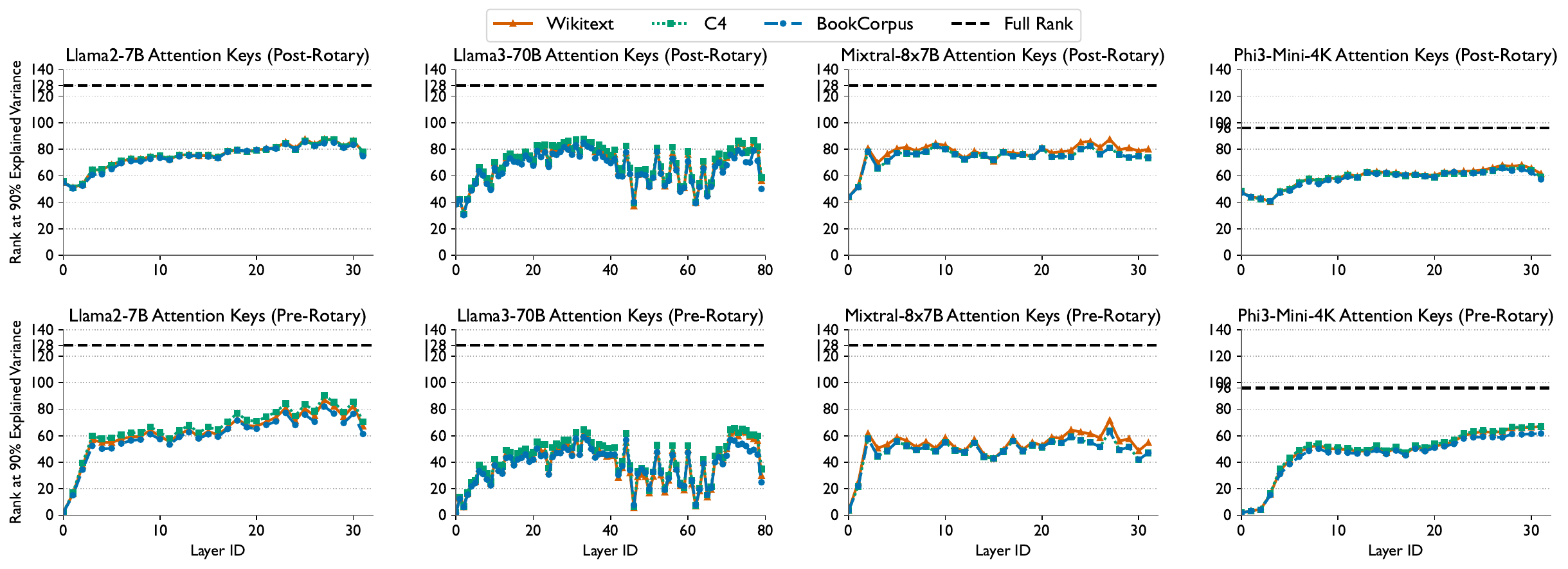}

    \caption{Rank at which 90\% of the variance is explained for pre-rotary and
    post-rotary keys produced by each layer averaged across all heads
    ($Rank_{l}@90$) for different models. We observe that all models exhibit
    significantly low rank (full dimensionality is 128 or 96 represented by the black dashed line) consistently across all
    datasets.} \label{fig:rank90}
\end{figure}

We observe that the dimensionality of the keys (both pre-rotary and
post-rotary) is significantly lower than the full dimensionality of the keys
across all calibration datasets. Furthermore, the $Rank_{l}@90$ for a
particular layer is consistent across datasets, for all combinations of models
and datasets. This indicates that the lower-dimensional structure of the keys
is consistent when calculated using different calibration datasets. Another
trend we observe is that the initial layers of most models have a very low
rank, as compared to the later layers, and this trend is particularly prominent
for the pre-rotary keys. Lastly, we also observe that for most models, the
average of $Rank_{l}@90$ across all layers is lower for pre-rotary keys as
compared to post-rotary keys, indicating that the rotary embeddings increase
the dimensionality of the keys.  Further analysis on the variation of the rank
across different heads within a layer and across different layers within a
model can be found in Appendix \ref{appendix_dimanalysis}.

These results indicate the existence of the following properties: (1) The keys
produced by the attention layers of transformer models lie in a significantly
lower-dimensional space. (2) The lower-dimensional structure of the keys is
consistent across different calibration datasets. (3) Rotary embeddings increase
the dimensionality of the keys for most models. We now use the first two
properties to propose \method, an efficient sparse-attention method.

\section{\method: Low-Dimensional Key Representations}
\label{sec:loki}
We now describe our proposed algorithm for sparse attention -- \method.
\method leverages low dimensional projections of the keys in the KV-cache to
efficiently and accurately select the top-$k$ (12.5-25\%) most relevant tokens
for self attention.  Before discussing our approach, let us first look at some
theoretical properties of attention in the PCA-transformed space of the key
tensors. 

\subsection{Properties of Attention in the PCA-transformed Space}
\label{subsec:properties}

We begin by proving two lemmas that provide the rationale for our approach to
compute attention in the PCA-transformed space. 

\begin{lemma}
  Let $D$ be the dimension of an attention head and $\bP \in \mathbb{R}^{D \times D}$ be the PCA projection matrix
  of key tensors calibrated offline on a dataset. Assuming we are generating the $S^{\mathit{th}}$ token in the sequence,
   let $\q_S \in \mathbb{R}^{1 \times D}$ be the query vector for the $S^{th}$ token,
  $\K_{:S} \in \mathbb{R}^{S \times D}$ be the key vectors, including the past
  $(S-1)$ keys and the current key. Then, the attention scores computed using
  the PCA-transformed query and keys are equivalent to the attention
  scores computed using the original query and keys.
  \label{lemma:projection}
\end{lemma}
\vspace{-0.2in}
\begin{proof}
  Let $\qhat_S = \q_S\bP$ and $\Khat_{:S} = \K_{:S}\bP$ be the PCA
  transformed query and key vectors. 
  Focusing on the dot product term in the attention computation (Equation \ref{eq:attention}), we have:
  \begin{align*}
    \q_S\K_{:S}^{T} &= \q_S(\Khat_{:S}\bP^{T})^{T} \text{ [inverting the PCA transform]}\\
    &= \q_S((\bP^{T})^{T}\Khat_{:S}^{T}) = (\q_S\bP)\Khat_{:S}^{T} = \qhat_{S}\Khat_{:S}^{T}
  \end{align*}
It is important to note here that Lemma \ref{lemma:projection} holds for 
any orthogonal $\bP$.
  \end{proof}

\begin{lemma}
  Let $\Khat_{:S,:d} \in \mathbb{R}^{S\times d}$ ($d < D$) be the reduced dimension key vectors 
  obtained by projecting the key vectors onto the first $d$ principal components of $\bP$ . 
  Then, the attention scores computed using $\Khat_{:S,:d}$ are a good approximation of the
  the actual attention scores.
  \label{lemma:reconstruction}
\end{lemma}
\vspace{-0.2in}
\begin{proof}
  Let $\textbf{R}_{:d} \in \mathbb{R}^{d \times D}$ be an orthogonal transformation 
  that transforms the keys into the reduced dimension space as $\bL_{:S,:d} = \K_{:S}\textbf{R}_{:d}$. 
  Our objective is to minimize the following expression:
  \begin{align}
    \min_{\bR_{:d}}||\q_S\K_{:S}^{T} - \q_{S}(\bL_{:S,:d}\bR_{:d}^{T})^{T}||_2^2
  \end{align}
  Using Cauchy-Schwarz inequality, we have:
  \begin{align}
    ||\q_S\K_{:S}^{T} - \q_S(\bL_{:S,:d}\bR_{:d}^{T})^{T}||_2^2 &\leq ||\q_S||_2^2||\K_{:S}^{T} - (\bL_{:S,:d}\bR_{:d}^{T})^{T}||_2^2
  \end{align}
  We change our objective to minimize the upper bound in the RHS instead of the original objective. 
  We know that PCA minimizes the reconstruction error (2nd term in the RHS) among all 
  the orthogonal transformations. Thus, it follows that the optimal value of $\bR^{*}_{:d} = \bP_{:d}$, and 
  $\bL^{*}_{:S,:d}=\Khat_{:S,:d}$
\end{proof}

Since we minimize an upper bound when proving Lemma \ref{lemma:reconstruction},
it is possible that some other transformation might give a better approximation
to the dot product. Thus, in our experiments, we use PCA transforms computed on
both the pre-rotary and post-rotary keys as candidate transformations.

Based on these lemmas and the inherent low-dimensional nature of key tensors in
attention, we now introduce the workings of the \method~algorithm. 

\subsection{PCA-based Top-K Algorithm}

Loki implements a PCA-based Top-K Attention approach. Previous works have shown
that attention scores for a query are highly concentrated on a small subset of
keys~\citep{xiao2024efficient, sun2024massive}.  This observation has motivated
several methods that compute attention using only the top-$k$ most relevant
keys. However, these previous works either compute the exact attention scores
and then select the top-$k$ keys~\citep{topkattention} or compute non-exact
scores but have significantly higher memory
requirements~\citep{ribar2023sparq}.  \method~alleviates these issues by
computing approximate attention scores (for ranking the keys) in the reduced
lower-dimensional space, without any significant increase in memory
requirements. Algorithm \ref{alg:topkpca} shows our \method~method. Line 5 of
the algorithm computes the approximate attention scores using $d$ principal
dimensions of the query and key vectors. Lines 6-7 select the top-$k$ keys
based on the approximate attention scores. Line 8 computes the exact attention
scores using the selected top-$k$ keys, directly in the transformed space
(Lemma \ref{lemma:projection}).

\begin{algorithm}[h]
  \caption{\method: PCA-based Top-K Attention}
  \label{alg:topkpca}
  \begin{algorithmic}[1]
  \Require At the $S^{th}$ step - Input: $\x_S \in
  \mathbb{R}^{1 \times D}$, KV-cache: $\Khat_{:S-1}, \V_{:S-1} \in \mathbb{R}^{(S - 1)
  \times D}$, Projection Matrix: $\bP \in \mathbb{R}^{D \times D}$, Configuration parameters (reduced dimensionality, top-$k$): $d$, $k$ 
  \Function {\method-Attention}{$\x_S, \Khat_{:S-1}, \V_{:S-1}, \bP, d, k$} 
  \State $\q_S, \bk_S, \bv_S \gets computeQKV(\x_S)$ 
  \State $\qhat_S \gets \q_S\bP$, $\bkhat_S \gets \bk_S\bP$ 
  \State $\Khat_{:S} \gets concat(\Khat_{:S-1}, \bkhat_S)$, $\V_{:S} \gets concat(\V_{:S-1}, \bv_S)$ 
  \State $\ba_{approx} \gets \qhat_{S,:d}(\Khat_{:S,:d})^{T}$ 
  \State $indices \gets topk(\ba_{approx}, k)$ 
  \State $\Khat_{:S}^{\prime} \gets \Khat_{:S}[indices]$, $\V_{:S}^{\prime} \gets \V_{:S}[indices]$
  \State $\ba_{exact} \gets softmax(\frac{\qhat_S\Khat_{:S}^{T}}{\sqrt{D}})$ 
  \State \Return $\ba_{exact}\V_{:S}^{\prime}$
  \EndFunction
  \end{algorithmic}
\end{algorithm}

\vspace{0.08in}
\noindent \textbf{Compute and Memory Analysis:} For vanilla attention, the
complexity of computing $\q_S\K_{:S}^{T}$ is $\bigO(DS)$ and the complexity of
multiplying the values with the attention scores is $\bigO(DS)$. For \method,
the complexity of calculating the approximate attention scores (Line 5) is
$\bigO(dS)$. The complexity of selecting the top-$k$ keys (Lines 6-7) is
approximately $\bigO(Slog(S) + k)$ (sorting followed by selection). The
complexity of calculating the exact attention scores and multiplying with the
values (Line 8-9) is $\bigO(2Dk)$. Additionally, the complexity of projections
into the PCA space (Line 3) is $\bigO(2D^2)$. Assuming the complexity of
selecting the top-$k$ keys is small compared to the other operations, the
overall complexity of the algorithm is $\bigO(dS + 2Dk + 2D^2)$. Then, we have:
\begin{align} speedup = \frac{2DS}{dS + 2Dk + 2D^2}  = \frac{1}{d/2D + k/S +
D/S} \approx \frac{1}{d_f/2 + k_f} \hspace{1em}(\text{given }D << S)
\end{align} where, $d_{f} = d/D$ and $k_{f} = k/S$. The memory requirement of
the KV-cache is the same as the original attention, with a small overhead of
storing the PCA transformation matrix.

\subsection{Implementation in Triton} \label{sec:kernel}

Performing \method~efficiently involves complex indexing operations within the
KV-cache (lines 5 and 7 of Algorithm~\ref{alg:topkpca}).  Standard PyTorch
operations create temporary, dense copies of the KV-cache data in memory,
leading to slowdowns due to expensive memory access. To alleviate this issue,
we develop optimized kernels in Triton~\citep{triton:openai} for the three
matrix multiplication operations in \method.  Our kernels can directly access
relevant subsets of the KV-cache (both feature and sequence dimensions) and
perform computations within GPU registers. This eliminates the need for
creating dense copies, significantly improving performance. Our approach builds
on SparQ~\citep{ribar2023sparq}, which introduced similar kernels for top-$k$
attention calculations. However, we identified and addressed inefficiencies in
the SparQ kernels, which resulted in speedups of nearly $2-3\times$ in certain
scenarios.  (see Appendix~\ref{append:comp_pcatopk}).

\section{Experimental Setup}
\label{sec:setup}
We evaluate \method on the basis of 
perplexity using the WikiText-2~\citep{wikitext-103} dataset (test split), and on the basis of downstream task performance for  
short contexts using the LM-harness
benchmark~\citep{eval-harness} and long contexts
using LongBench~\citep{bai2023longbench}. For the short-context evaluation, we
choose the same tasks and associated metrics as the HuggingFace OpenLLM
leaderboard~\citep{open-llm-leaderboard}. For the LongBench tasks, we evaluate
on all the English language tasks. 

We compare our method against three methods -- full attention without any
approximations, the exact TopK approach which computes the exact attention
scores and then uses the top-$k$ tokens to compute the final output, and
H$_2$O~\citep{zhang2023hH2O}, a popular token-eviction method. For these
comparisons, we show the results with a budget size of $k_f$ = 0.25 and 0.125.
For our method, we additionally use $d_f$ = 0.25 and 0.125. This configuration
of our represents a 2.6x theoretical speedup. Table \ref{ref:budget} provides
an overview of the methods compared and the associated budget terms. H$_2$O's
budget was split equally between the heavy hitter and recent tokens, as per the
author's recommendations. For H$_2$O, we were unable to run the GSM8K task as
the the author's ML benchmarking code was too memory intensive to run for that
task. For the aforementioned experiments, we generate PCA transforms using the
WikiText-103 dataset. For the LongBench tasks, we compare our method with the
full attention baseline as we were unable to run H$_2$O due to memory
constraints. 

For the generalizability study, we compare the results of our method with PCA
transforms from different calibration datasets:
WikiText-103~\citep{wikitext-103}, C4~\citep{raffel2023exploring}, and
BookCorpus~\citep{zhu2015aligning}. Additionally, we also benchmark our triton
based implementation of \method~by running an attention microbenchmark on a
Llama2-13B-like setup (same hidden size and number of heads) for various prompt
and generation lengths, and demonstrate speedups over vanilla attention.

\begin{table}[t]
  \small
  \centering
  \caption{Explanation of key-budget and dimensionality (Dim.) for different approaches, along with the expected speedup and memory savings.}
  \begin{adjustbox}{width=\linewidth}
  \begin{tabular}{lccp{7cm}cc}
      \toprule
      Method & Budget & Dim. & Description & Speedup & Memory Savings \\
      \midrule
      Exact Top-K & $k_f$ & Full & $k_f$ fraction of keys selected using exact attention scores & No & No \\
      H$_2$O & $k_f$ & Full & $k_f$ fraction of keys \& values selected using H$_2$O policy& $\frac{1}{k_f}$ & $\frac{1}{k_f}$ \\
      \method~& $k_f$ & $d_f$ & $k_f$ fraction of keys \&values selected using attention scores computed 
      with $d_f$ fraction of full dimensionality & $\frac{1}{(d_f/2) + k_f}$ & No \\
      \bottomrule
  \end{tabular}
  \end{adjustbox}
  \label{ref:budget}
\end{table}

All experiments are run on NVIDIA A100 GPUs with 40 and 80 GB of memory on the
Perlmutter~\citep{perlmutter} supercomputer. For larger models, we use
AxoNN~\citep{singh:ipdps2022,singh:arxiv2024} to shard the model across
multiple GPUs.

\section{Results}
\label{sec:results}
We now present the comparisons of \method with full attention and other sparse
attention methods, including a comparison of the computation times.

\subsection{Comparison with Full Attention}

\begin{figure}[h]
  \centering
    \includegraphics[width=\textwidth]{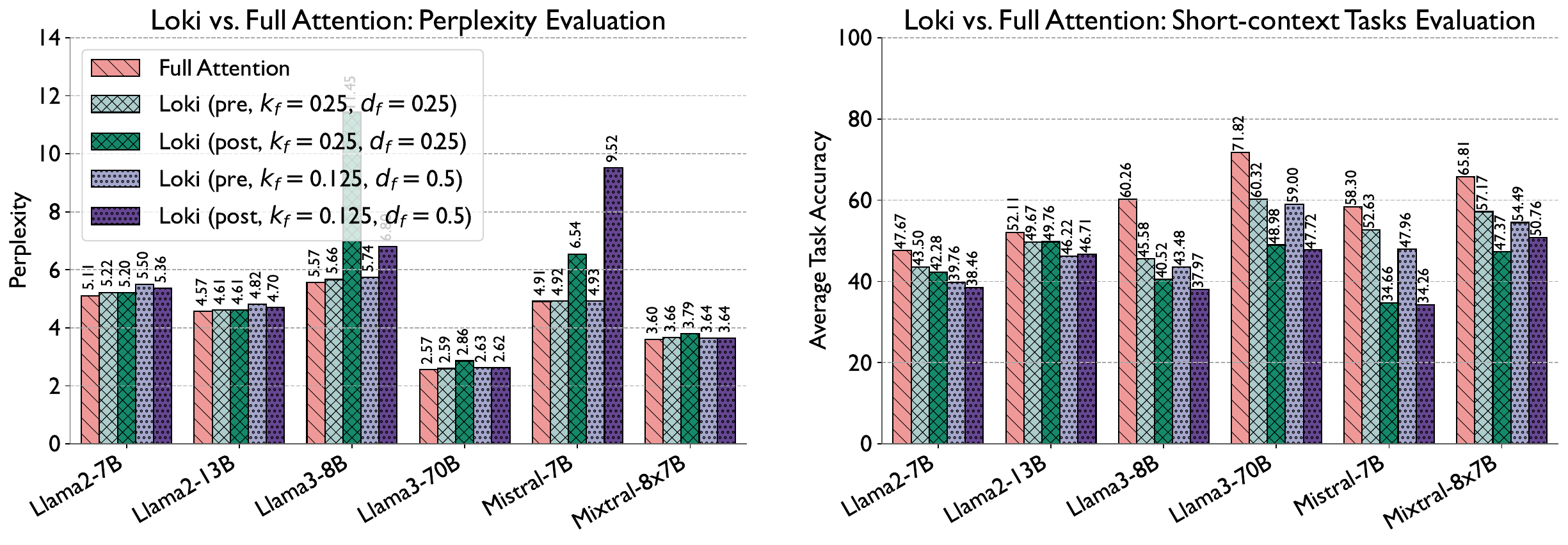}
    \caption{Evaluation of \method~on perplexity (left plot) and short-context tasks (right plot) for different models. 
    Task accuracy is an average across all short-context tasks mentioned in \ref{sec:setup}.}
    \label{fig:all_model_eval}
\end{figure}

Let us begin our discussion with Figure \ref{fig:all_model_eval}, showing the
perplexity (left) and short-context downstream task evaluation (right) results
for \method~on different models.
We focus on the Llama2-7B model, comparing pre-rotary (light green/purple)
and post-rotary (dark green/purple) PCA transforms for different $k_f$ and
$d_f$ values. For Llama2-7B, we see that the performance of both candidate
transforms is similar. This trend is consistent across all the models except
for Llama3-8B/70B and Mistral-7B, where the post-rotary PCA transform performs
significantly worse than the pre-rotary one. For Llama3-8B, perplexity jumps
from about 5 for the full attention to over 10, a significant decline not seen
with the pre-rotary transform. Mistral-7B shows a similar pattern. This is a
surprising observation since attention scores are calculated from post-rotary
keys in the original attention mechanism. A possible explanation is that 
post-rotary PCA captures token distributions tied to specific positions in 
the calibration dataset, while pre-rotary PCA may generalize better by using 
less positional information. Nevertheless, at least one of the PCA
transformations performs well for every model. For subsequent results, we only
show the better-performing transformation for each model. 

\begin{figure}[h]
  \centering
    \includegraphics[width=\linewidth]{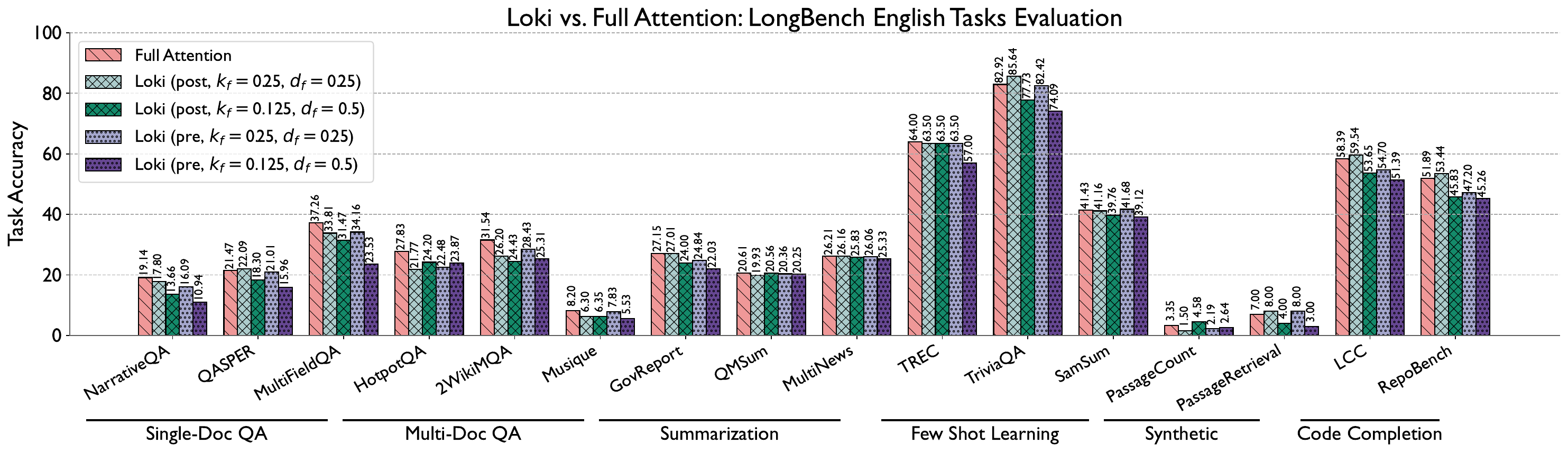}
    \caption{Evaluation of \method~on LongBench tasks for the Llama2-7B-Chat model.}
    \label{fig:long_context_eval}
\end{figure}

Figure \ref{fig:long_context_eval} shows the performance of \method~on the
LongBench tasks for the Llama2-7B-Chat model. We see that for all tasks, either
one of the two candidate transforms performs similarly to full attention. For
Summarization, Few Shot Learning, Synthetic, and Code Completion task
categories, the best performing \method~configuration is at par or better than
the full attention model. For the Single-Doc QA and Multi-Doc QA task
categories, ~\method~performs slightly worse than the full attention model,
with the biggest drop in performance observed for HotpotQA of around 3\%.
Comparing different $(k_f, d_f)$ settings, we see that using $k_f$ = 0.25 and
$d_f$ = 0.25 (green), is better than using $k_f$ = 0.125 and $d_f$ = 0.5
(purple) for all models and tasks (short-context and long-context). These two
settings balance speed and performance well, with the first being superior for
accuracy.

\subsection{Comparison with Other Sparse Attention Methods}
\label{subsec:comparison_other_sparse}

\begin{figure}[h]
  \centering
    \includegraphics[width=\textwidth]{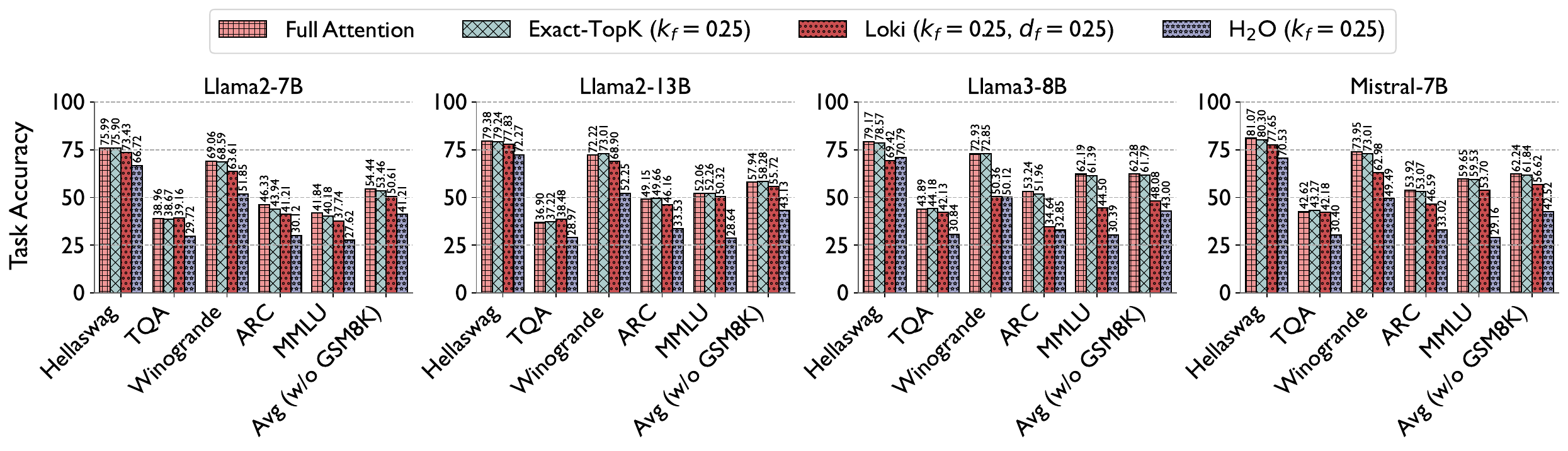}
    \caption{Downstream task performance for \method~and other approaches for different models (higher is better).
     GSM8K is excluded, as we were unable to run H$_2$O for this task.}
    \label{fig:tasks_baseline_eval}
\end{figure}

Next, we compare the performance of \method~with other methods, using $k_f$
= 0.25 for all methods and $d_f$ = 0.25 for ours. Table \ref{tab:perplexity}
shows the perplexity results for Llama2-7B/13B, Llama3-8B, and Mistral-7B.
\method's perplexity drop is within 0.1 of full attention across all models, a
threshold considered acceptable for attention mechanism
approximations~\citep{Yao2023ZeroQuantV2EP}. In contrast, H$_2$O's perplexity
drop nears 0.2 for all models.  Figure \ref{fig:tasks_baseline_eval} confirms
this trend on short-context evaluation. \method~performs similar to full
attention for all models, except Llama3-8B, where the performance is notably
worse, but still better than H$_2$O.  Importantly, on the challenging MMLU
task, \method~degrades less than H$_2$O.

\begin{table}[h]
  \centering
  \caption{Perplexity evaluation of \method and other approaches for different models (lower is better).}
  \begin{adjustbox}{width=\textwidth, center}
  \begin{tabular}{@{}lccccccc@{}}
      \toprule
      Method & $k_f$ & $d_f$ & Speedup & Llama2-7B & Llama2-13B & Llama3-8B & Mistral-7B \\
      \toprule
      Full Attention & - & - & No & 5.1101 & 4.5680 & 5.5696 & 4.9140 \\
      Exact-TopK & 0.25 & - & No & 5.1809 & 4.5926 & 5.5716 & 4.9171 \\
      \midrule
      H$_2$O & 0.25 & - & Yes & 5.2810 & 4.7009 & 5.7056 & 5.0805 \\
      \method~& 0.25 & 0.25 & Yes & \textbf{5.2017} & \textbf{4.6102} & \textbf{5.6648} & \textbf{4.9233} \\
      \bottomrule
  \end{tabular}
  \end{adjustbox}
  \label{tab:perplexity}
\end{table}

It is important to note here that \method~is designed to be compatible with other
sparse attention methods. For instance, token-eviction methods like H$_2$O delete tokens to
save KV-cache memory, whereas \method~reduces memory bandwidth by selecting the
top-$k$ tokens without deletion, making them orthogonal. A combined approach
could involve using H$_2$O to delete tokens, then applying \method~to select
top-$k$ tokens from the remaining cache. Similarly, \method~is theoretically
orthogonal to quantization methods.

\begin{figure}[h]
  \centering
  \includegraphics[height=1.35in]{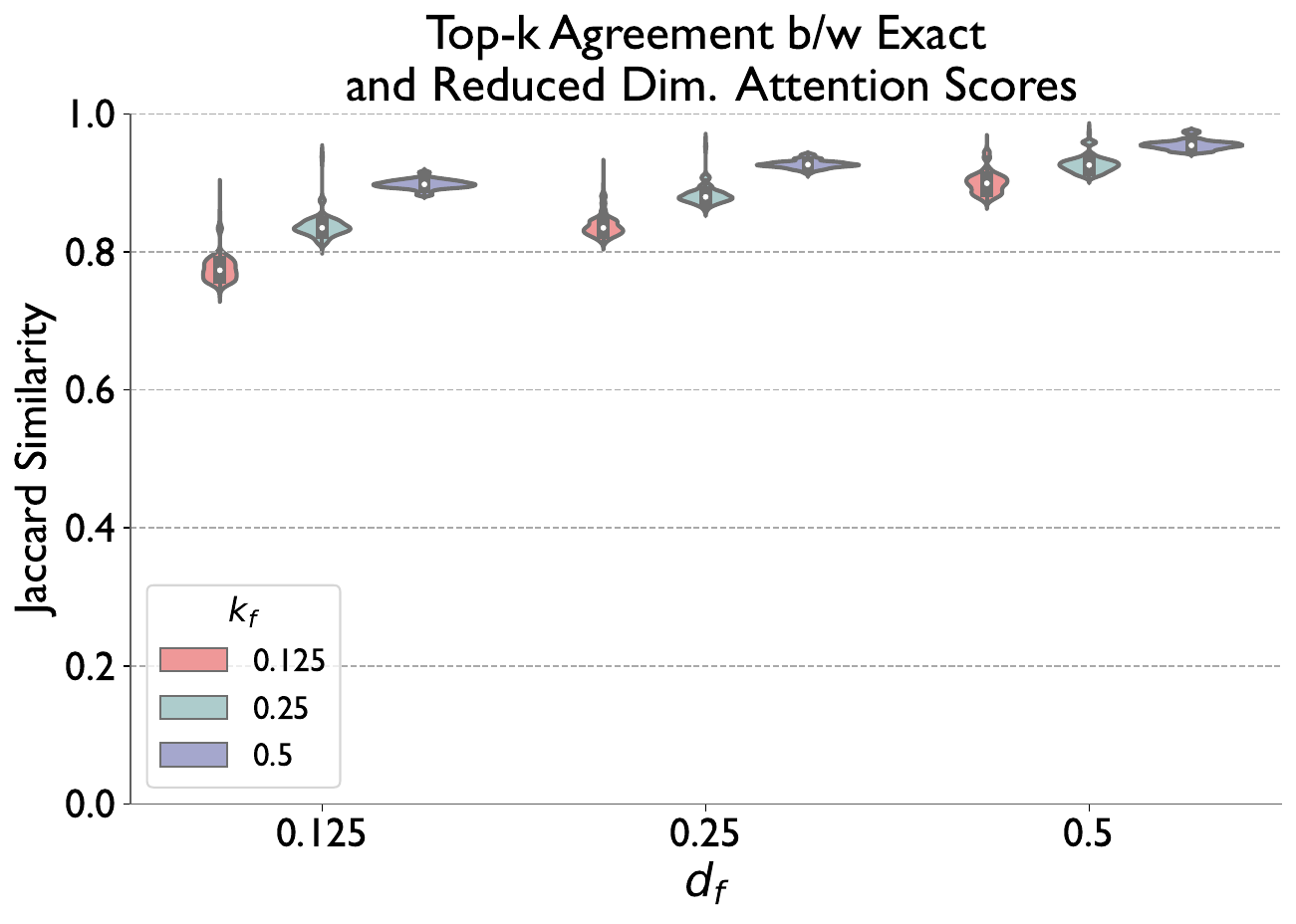}
  \hspace{0.1in}
  \includegraphics[height=1.35in]{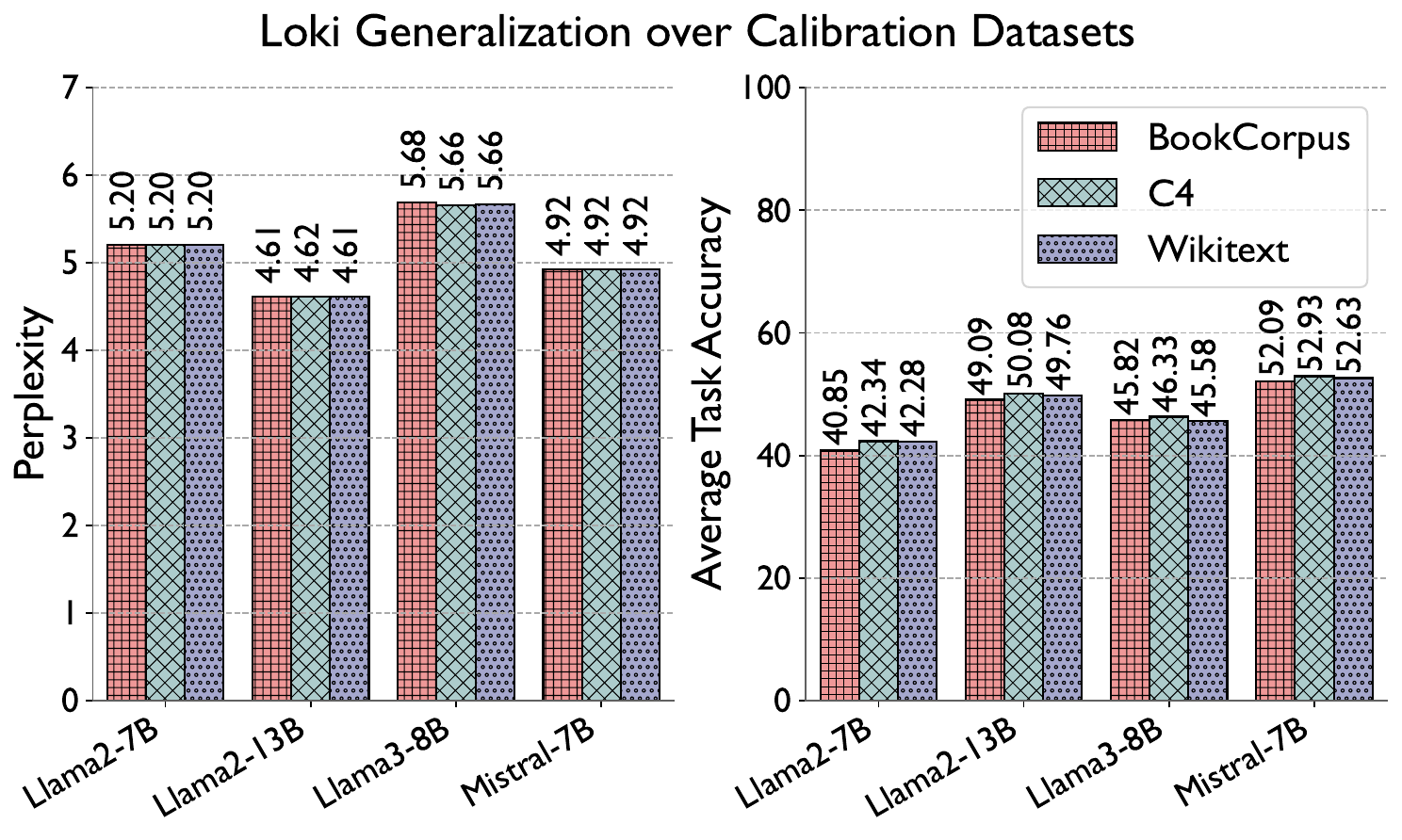}
  \includegraphics[height=1.35in]{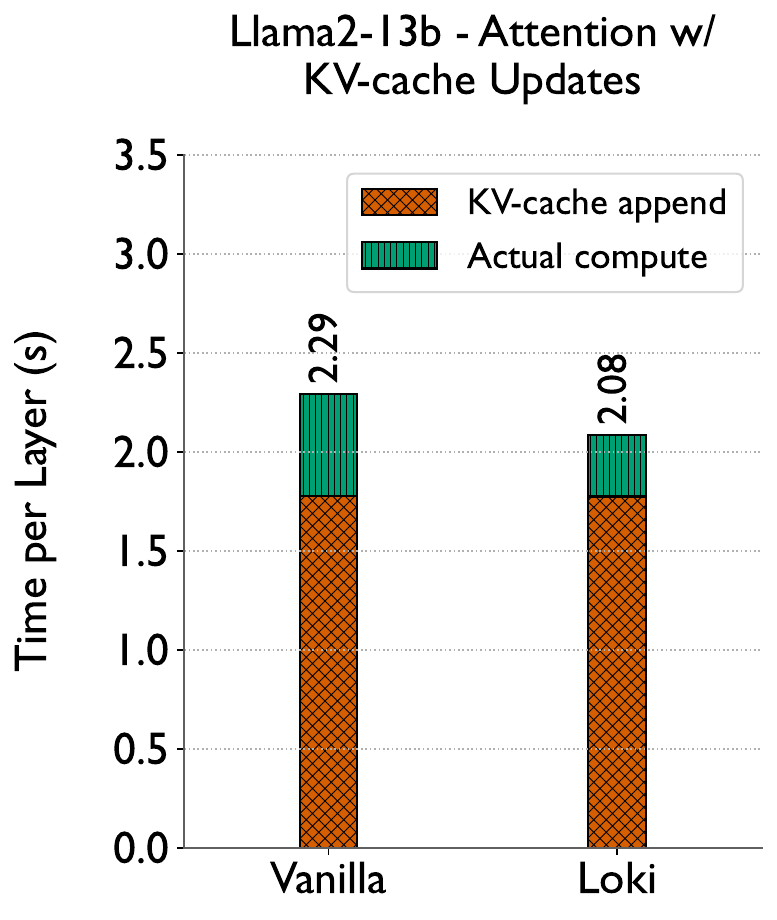}
  \caption{Top-$k$ agreement between \method~and Exact-TopK methods for Llama2-7B (left plot). 
  Performance of \method~using transformations derived from different calibration datasets (middle plots).
  Benchmarking vanilla attention and \method~for Llama2-13B using huggingface transformers with 
  cache append times (right plot, prompt length = 3072, generation length = 512).
  }
  \label{fig:topkagreement}
\end{figure}

Comparing \method~with Exact-TopK, we find similar performance for
Llama2-7B/13B and Mistral-7B. Exact-TopK represents the upper performance bound
for \method~if it could perfectly select the top-$k$ tokens. To understand why
\method~works well, we examined the top-$k$ agreement between ~\method's
reduced dimensional attention scores and exact attention scores. Figure
\ref{fig:topkagreement} shows the Jaccard similarity between the top-$k$ tokens
selected by both methods across all layers and heads for Llama2-7B. For the
settings: ($k_f = 0.25$, $d_f = 0.25$) and ($k_f = 0.125$, $d_f = 0.5$),
evaluated in Figure \ref{fig:all_model_eval}, the Jaccard similarity is around
0.9, validating that the \method~is able to select the top-$k$ tokens with high
accuracy.

\subsection{Generalizability}

We now turn our attention to the generalizability of the PCA transformations
used in our method. Figure \ref{fig:topkagreement} (middle) shows the
performance of \method~using PCA transformations derived from different
calibration datasets ($k_f = 0.25, d_f = 0.25$).  We see that the performance
of \method~is consistent across different calibration datasets, indicating that
the PCA transformations used in our method are generalizable. This is an
important observation as it shows that the PCA keys can be generated using a
variety of calibration datasets and still achieve good performance. 

\subsection{Computational Efficiency}

\begin{figure}[h]
  \includegraphics[height=1.75in]{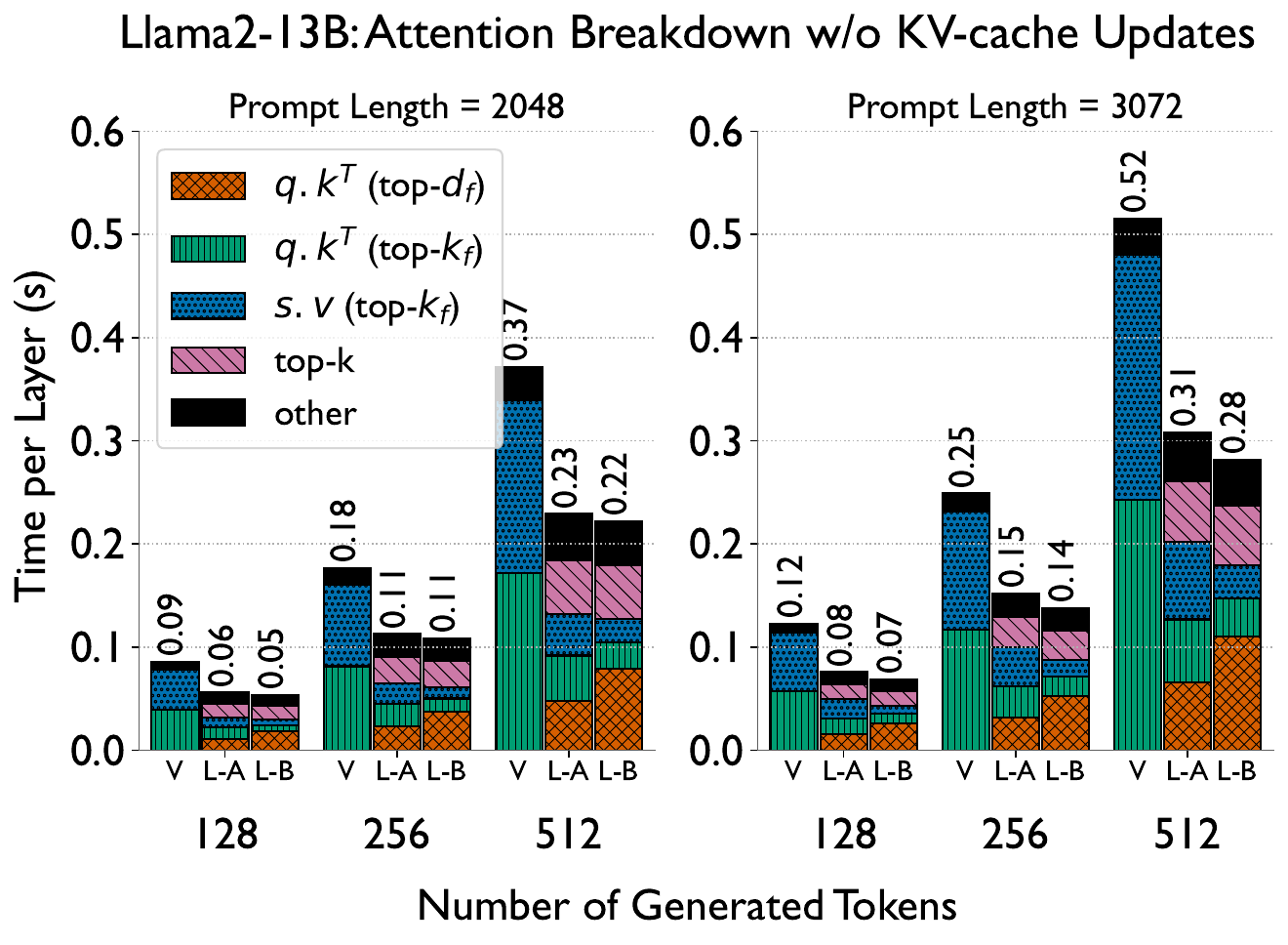}
  \includegraphics[height=1.75in]{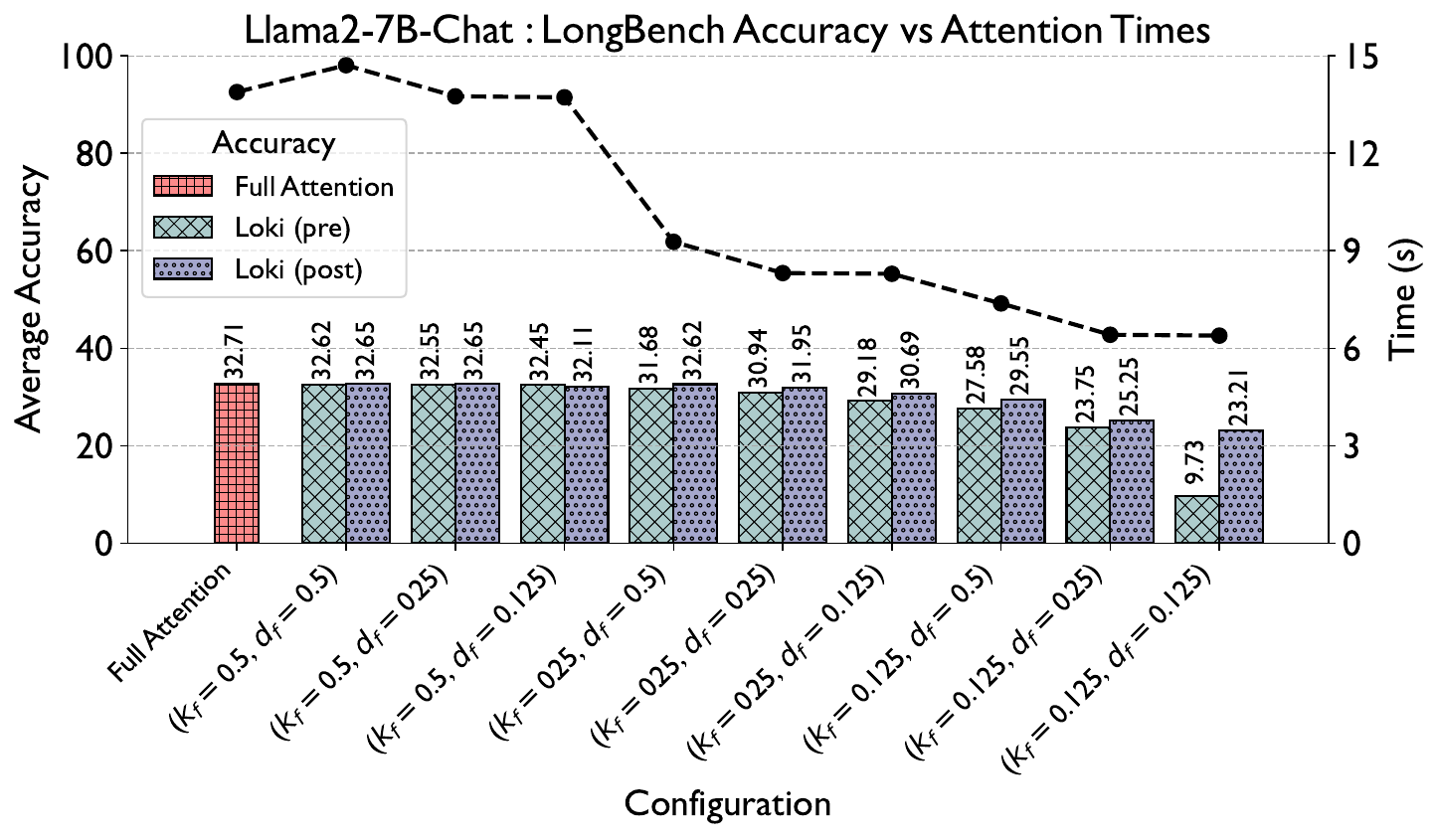}
  \caption{Time per layer for vanilla attention (V) and \method (\mbox{L-A:
${k_f=0.25,d_f=0.25}$}; \mbox{L-B: ${k_f=0.125,d_f=0.25}$}) for Llama2-13B
using huggingface transformers (left two plots). LongBench average accuracy for
different \method configurations, alongside micro-benchmark attention times (right plot, all
layers, prompt length = 3500 \& generation length = 512). We choose the prompt
length to match LongBench’s configuration for this model and generation length
to match the maximum in any LongBench task. For both figures, we use a batch
size of 16 and report the average time over 10 trials (std.~dev.~in measured times
was less than 0.05 percent of the mean).}
  \label{fig:compute}
\end{figure}

We now turn our attention to the computational efficiency of \method. Analyzing
Llama2-13B with Hugging Face Transformers exposed an interesting bottleneck
(Figure~\ref{fig:topkagreement}, rightmost). Regardless of the attention type
(vanilla or \method), more than 80\% of the time is consumed within the Hugging
Face framework for appending key-value pairs of the latest token to the
KV-cache. This shared bottleneck minimizes the overall performance improvement
of our optimizations. We hypothesize that using a more advanced inference
system like vLLM~\cite{woosuk2023vllm} could significantly reduce this append
time, but leave that exploration for future work. To isolate the impact of our
optimizations, the plots in Figure~\ref{fig:compute} focus solely on the
attention computation time, excluding the KV-cache append time.

In the left plot of Figure~\ref{fig:compute}, we see that \method~speeds up the
total attention compute time (excluding KV-cache appends) compared to vanilla
attention across various prompt and generation lengths. For a prompt length of
3072 and generation length of 512, \method~achieves nearly a 45\% speedup,
despite the fact that it incurs an extra matrix multiplication operation.  The
breakdowns also show that the top-$k$ operation is nearly as expensive as the
smaller matrix multiplications, which is a significant bottleneck. Replacing
PyTorch’s top-$k$ with a custom kernel could improve this. For the shorter
prompt length of 2048 we observe a speedup of around 40\% (generation length =
512), slightly lower than the speedup at 3072. This trend is expected as larger
prompts result in a bigger KV-cache, amplifying the impact of our
optimizations.

Figure~\ref{fig:compute} (Right) shows the accuracy vs. attention time
trade-off across various $k_f, d_f$ settings of \method, with accuracy measured
on LongBench and attention times from our microbenchmark. The previously
evaluated settings, $k_f = 0.25, d_f = 0.25$ and $k_f = 0.125, d_f = 0.5$,
provide a good balance between performance and accuracy, with $k_f = 0.25, d_f
= 0.25$ favoring accuracy slightly and $k_f = 0.125, d_f = 0.5$ favoring
performance.



\section{Conclusion}
\label{sec:conclusion}
In conclusion, we introduced \method, an algorithm for efficient sparse
attention that does not compromise the model quality while reducing the
computational complexity of self attention. We made a crucial observation that
key vectors in attention lie in a low-dimensional space, across different
models and datasets.
Leveraging this insight, \method uses attention scores computed in a
lower-dimensional space to rank and select the top-$k$ most relevant tokens
from the KV-cache. It then uses the full dimensionality only for the selected
tokens to compute the final attention. Our theoretical analysis shows that
\method can provide significant speedups in the attention step. To implement
this efficiently, we develop optimized kernels for the various sparse matrix
multiplications in our approach. Our empirical evaluation shows that \method
performs better than popular approximation methods on a variety of models and
tasks, with respect to preserving model quality. Finally, we show that \method
can provide speedups of up to 45\% over the vanilla attention empirically,
making it a promising approach to address the computational challenges in
transformer inference.

\vspace{0.08in}
\noindent \textbf{Limitations and Future Work:} 
\method~does not focus on reducing memory usage of the KV-cache currently. As
mentioned previously in \ref{subsec:comparison_other_sparse}, it can potentially
be combined with other sparse attention method for improved
memory-performance-accuracy trade-offs. Another direction involves storing the
KV-cache in CPU memory and transferring only the top-$k$ keys and values to the
GPU~\citep{lee2024infinigen}.

While \method~outperforms vanilla attention in our benchmarks, practical
deployment would require integration with efficient attention kernels like
FlashAttention~\citep{dao2022flashattention}. As seen in our compute
benchmarking, the top-$k$ selection operation could introduce a bottleneck
towards achieving this.  Investigating this bottleneck and integrating
\method~with optimized attention kernels is left for future work.

Our finding of the keys’ low intrinsic dimensionality suggests promising
research directions. The variation of this dimensionality across heads and
layers could further be explored. We briefly experimented with a variable $d_f$
policy per layer (see Appendix \ref{app:variable_d}), but did not observe
significant significant improvements. A more sophisticated policy could be
explored in future work.

\begin{ack}
This research used resources of the National Energy Research Scientific
Computing Center (NERSC), a Department of Energy Office of Science User
Facility using NERSC award DDR-ERCAP0029894. Soheil Feizi was supported in
part by the following grants: NSF CAREER AWARD 1942230, ONR YIP award
N00014-22-1-2271, ARO’s Early Career Program Award 310902-00001, Army Grant
No.~W911NF2120076, NSF award CCF2212458, NSF award 2229885, an Amazon Research
Award and an award from Capital One.

\end{ack}

\bibliographystyle{plain}
\bibliography{./bib/cite,./bib/pssg}

\clearpage
\appendix
\title{Appendix}
\section{Comprehensive Dimensionality Analysis}

\subsection{Complete Key-Dimensionaltiy Analysis for All Models}
\label{appendix_dimanalysis}
\begin{figure}[ht]
  \centering
    \includegraphics[width=0.24\textwidth]{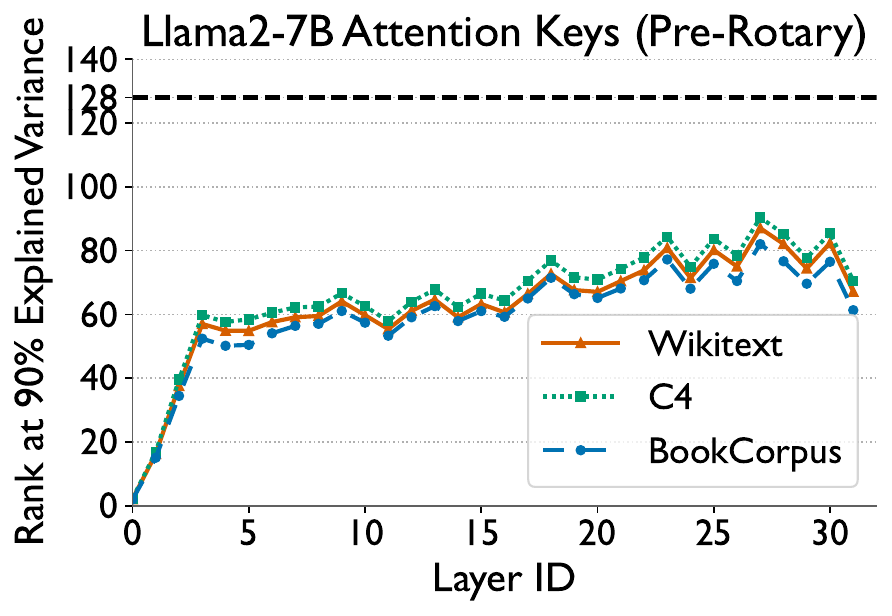}
    \includegraphics[width=0.24\textwidth]{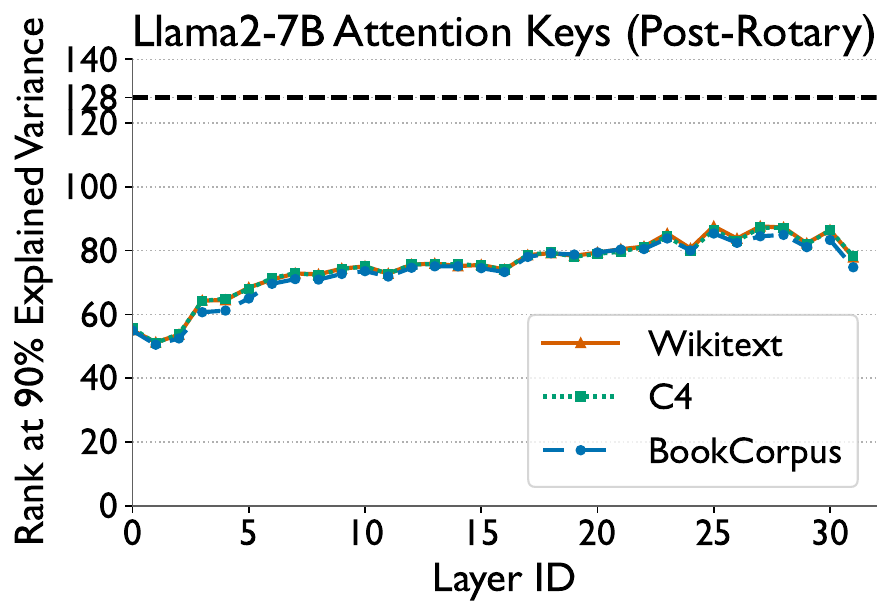}
    \includegraphics[width=0.24\textwidth]{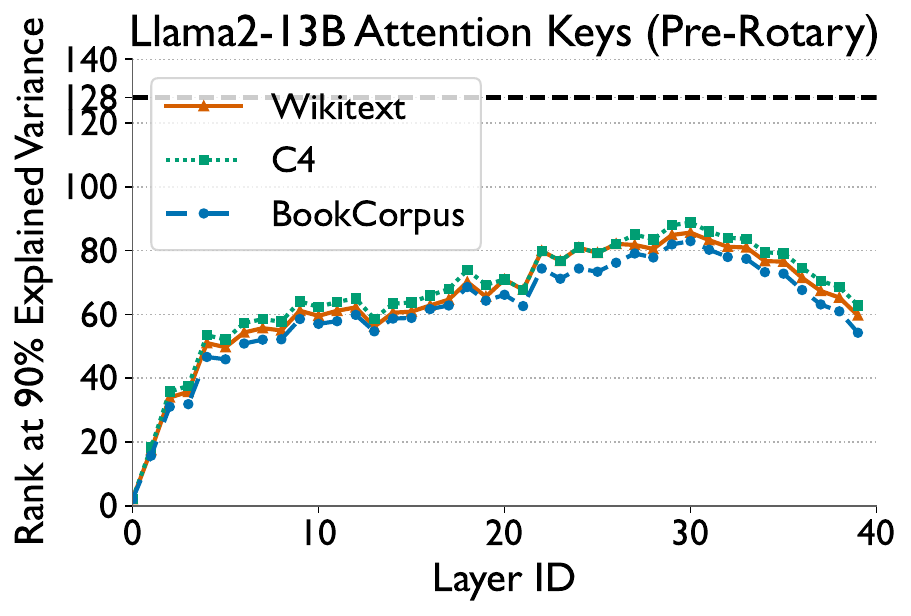}
    \includegraphics[width=0.24\textwidth]{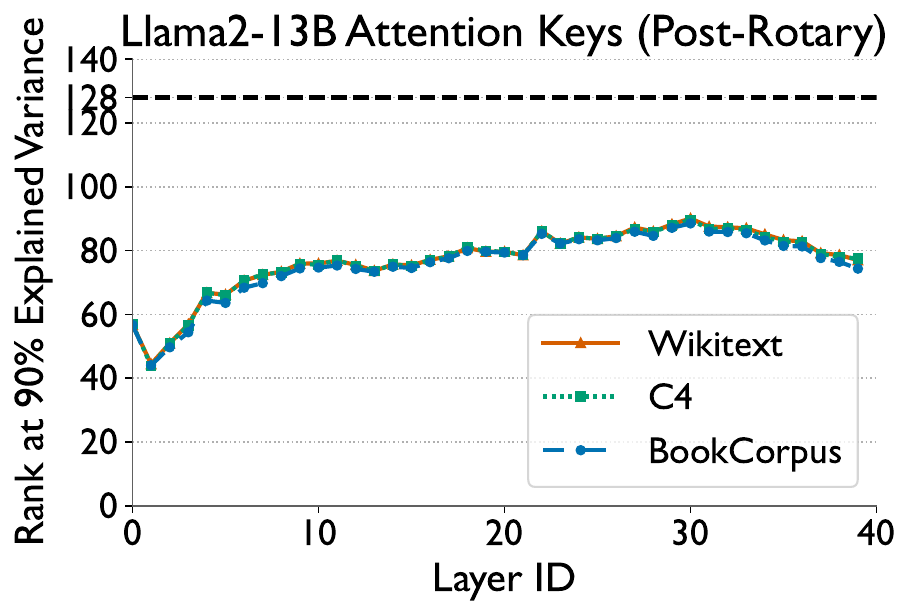}
    \includegraphics[width=0.24\textwidth]{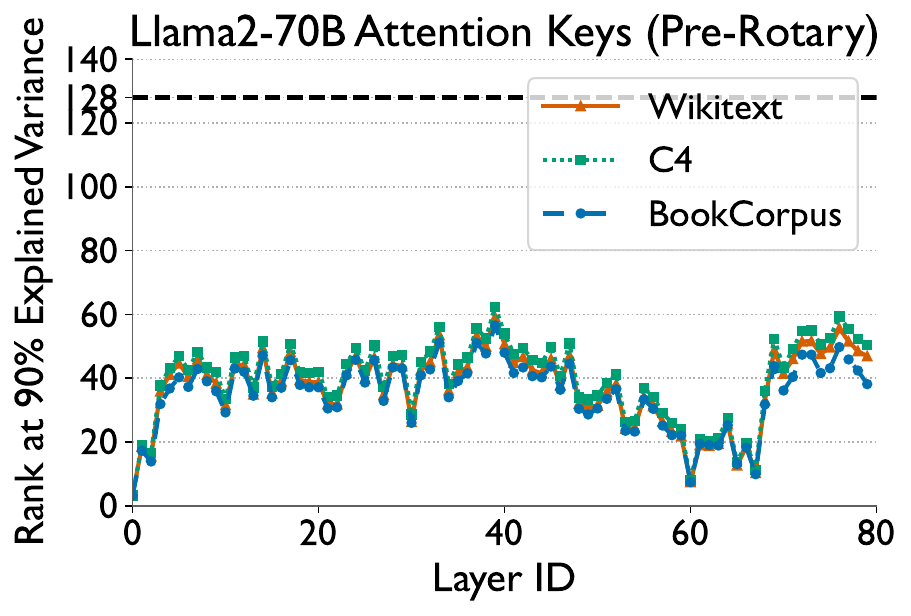}
    \includegraphics[width=0.24\textwidth]{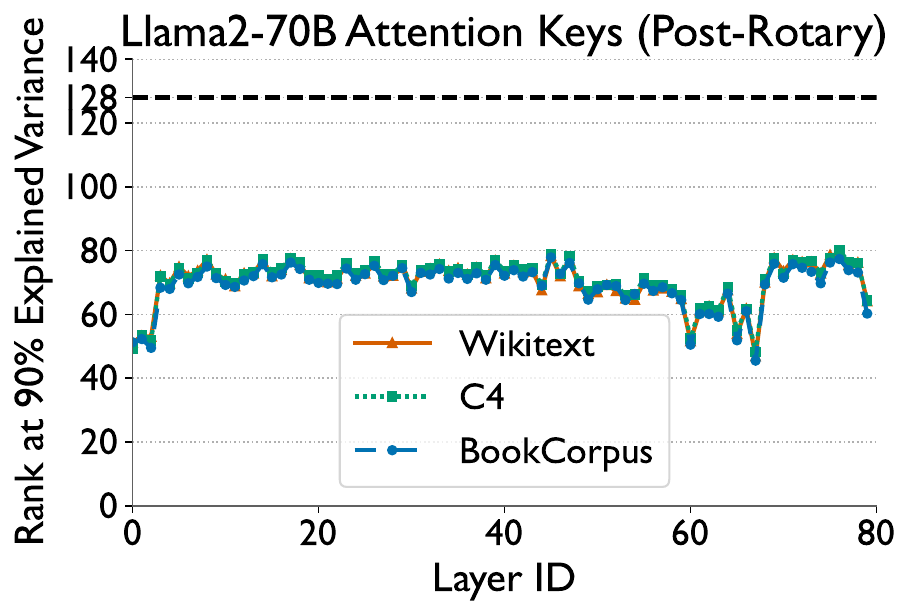}
    \includegraphics[width=0.24\textwidth]{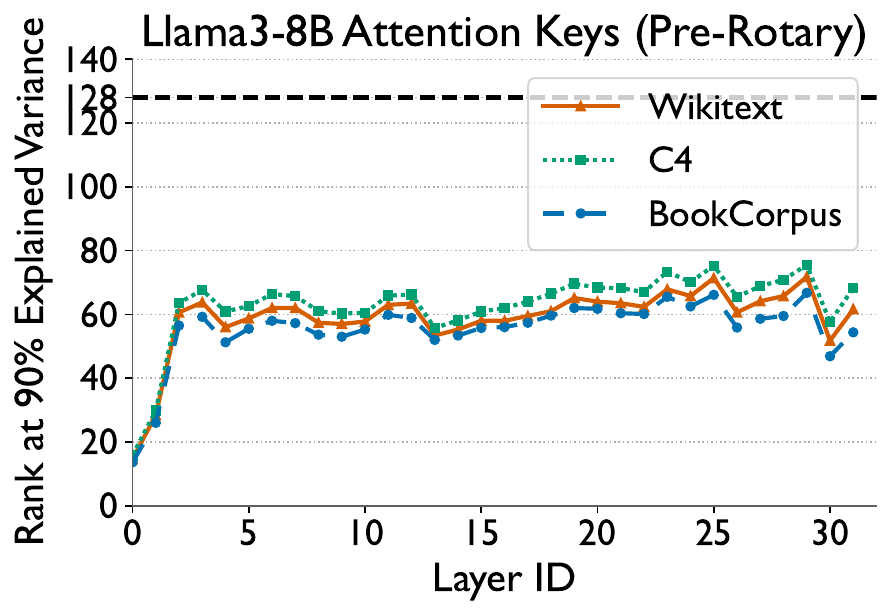}
    \includegraphics[width=0.24\textwidth]{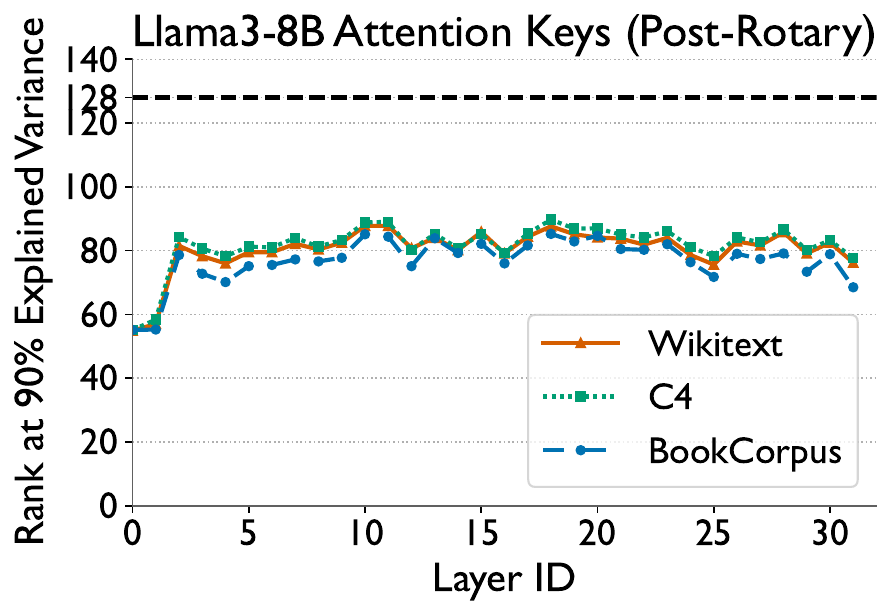}
    \includegraphics[width=0.24\textwidth]{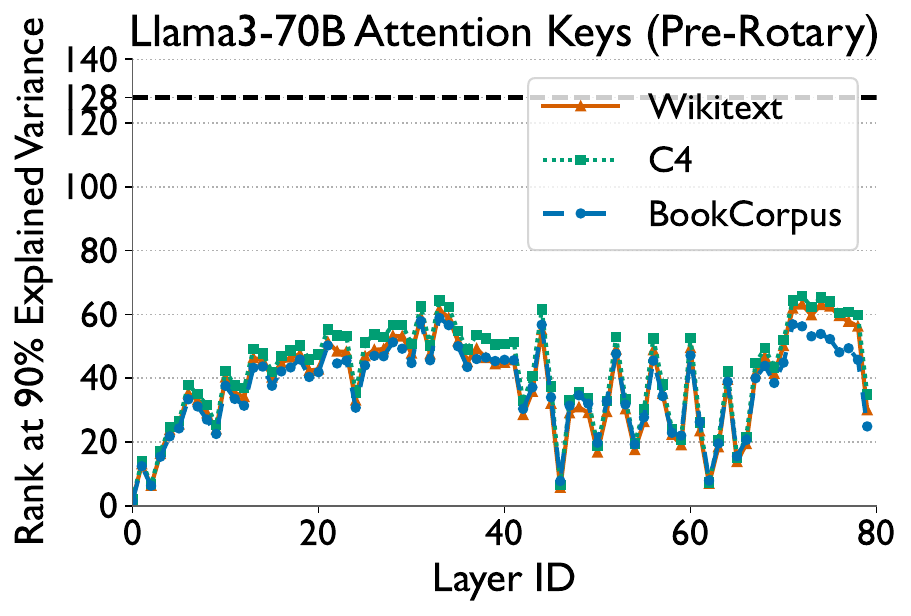}
    \includegraphics[width=0.24\textwidth]{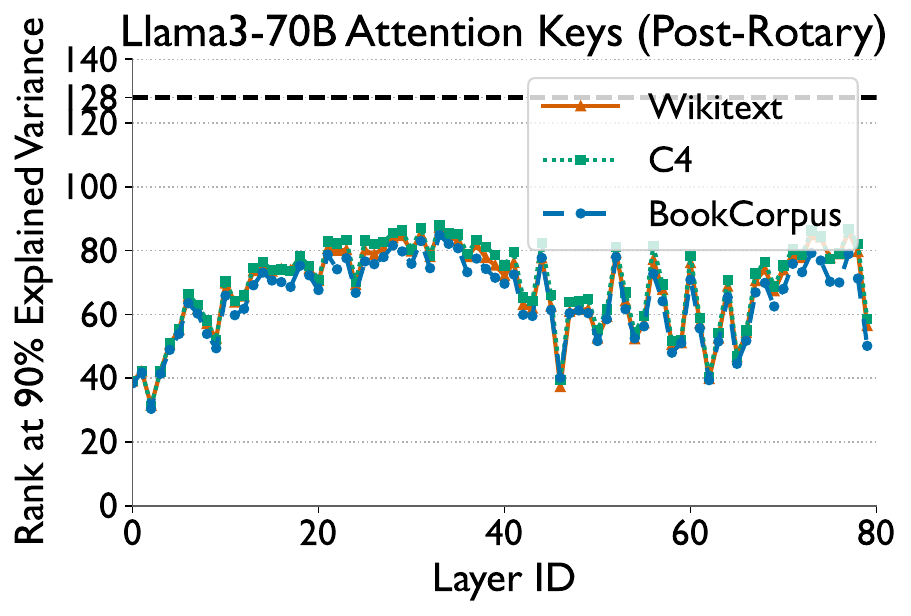}
    \includegraphics[width=0.24\textwidth]{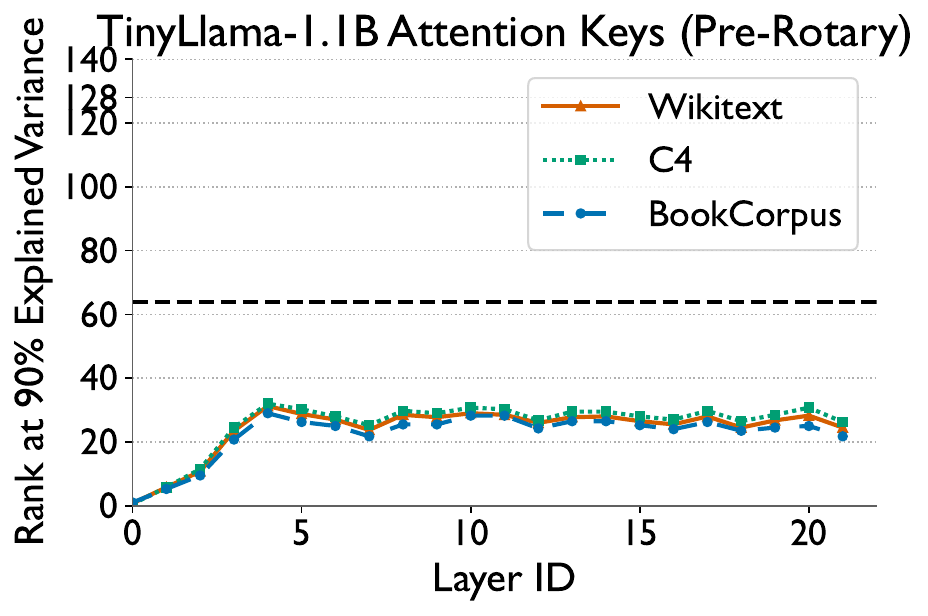}
    \includegraphics[width=0.24\textwidth]{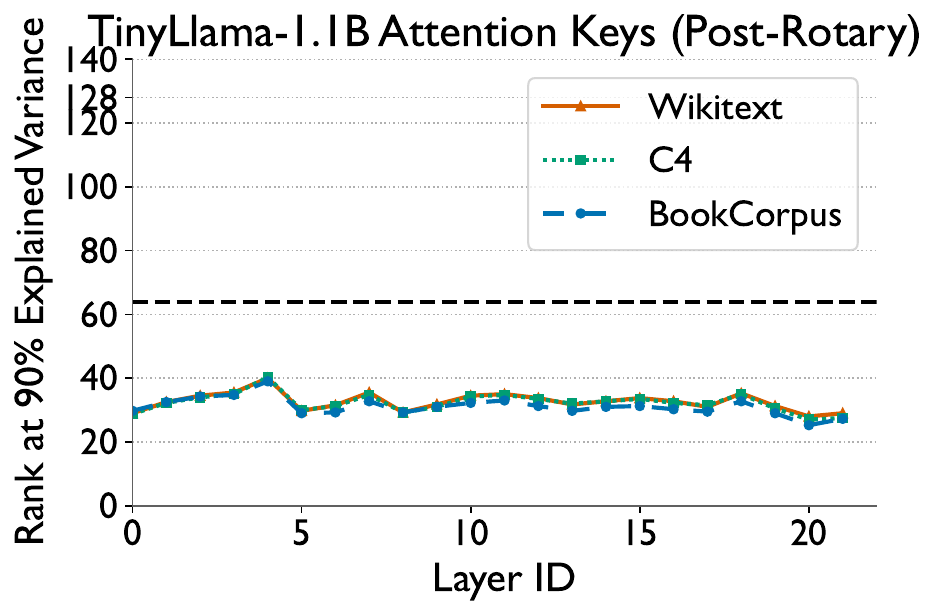}
    \includegraphics[width=0.24\textwidth]{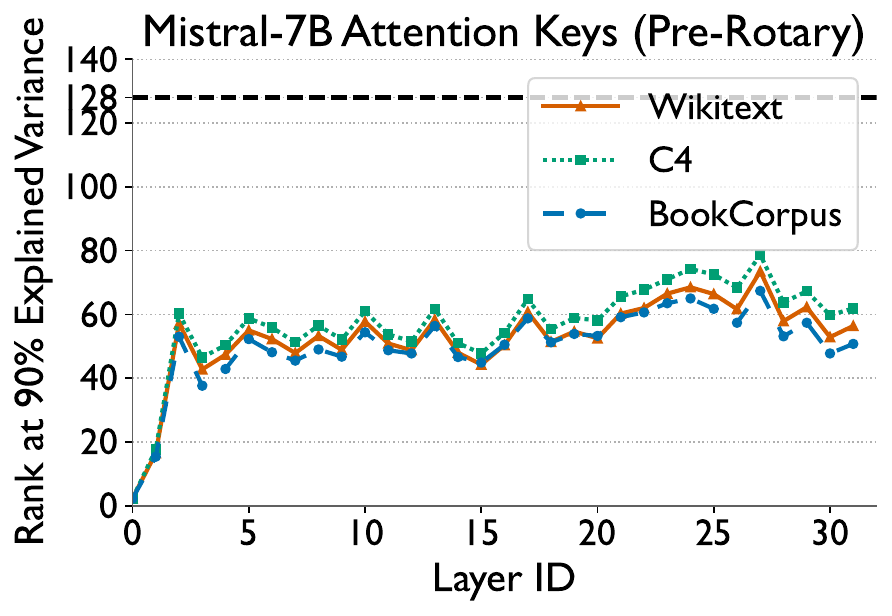}
    \includegraphics[width=0.24\textwidth]{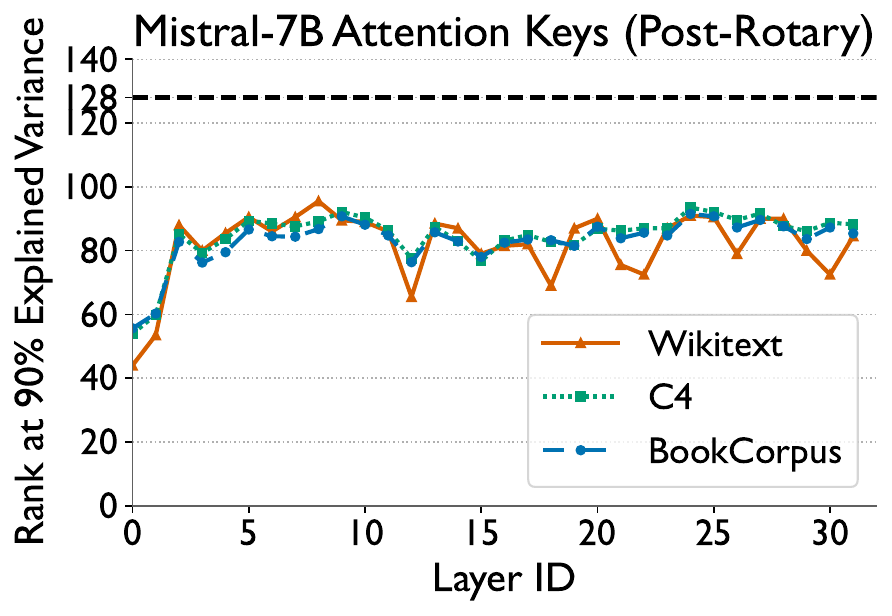}
    \includegraphics[width=0.24\textwidth]{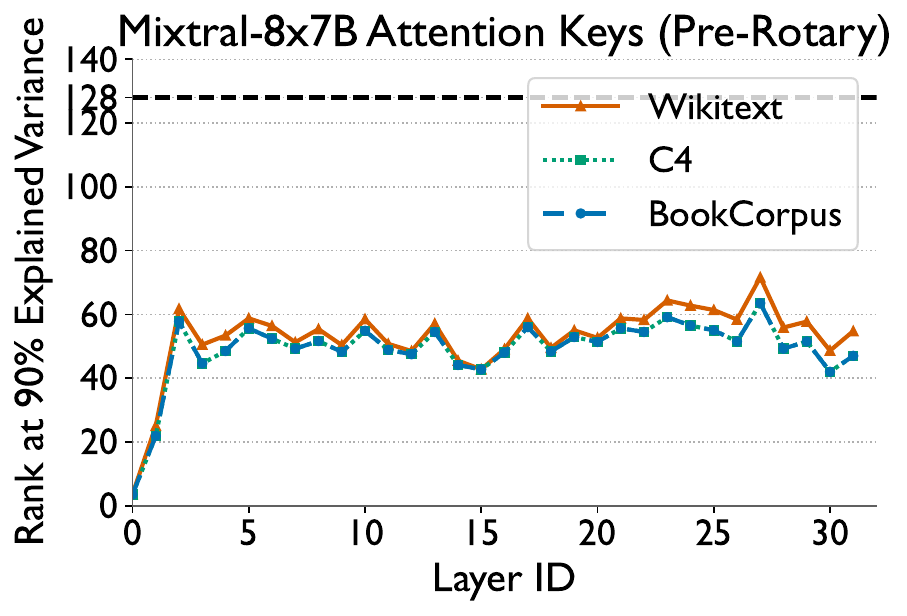}
    \includegraphics[width=0.24\textwidth]{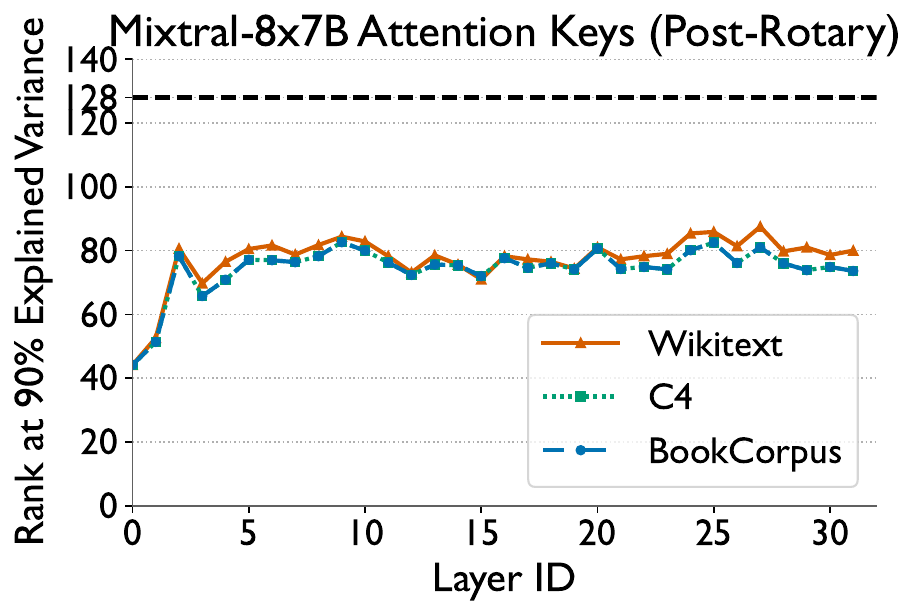}
    \includegraphics[width=0.24\textwidth]{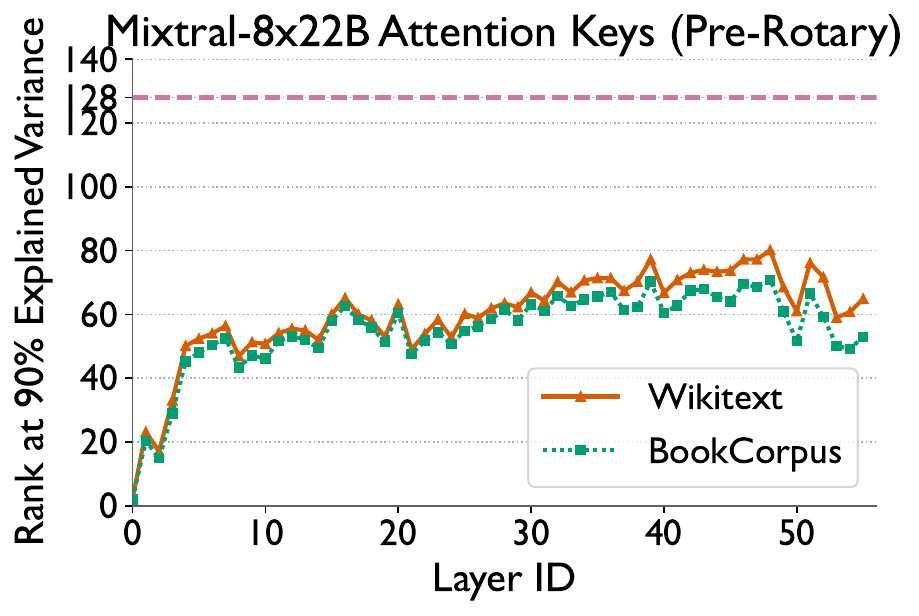}
    \includegraphics[width=0.24\textwidth]{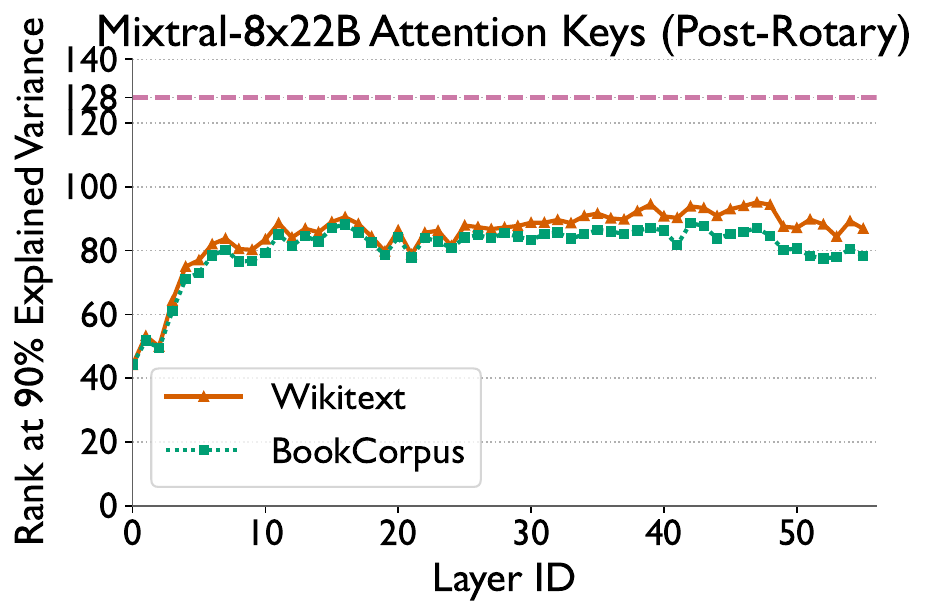}
    \includegraphics[width=0.24\textwidth]{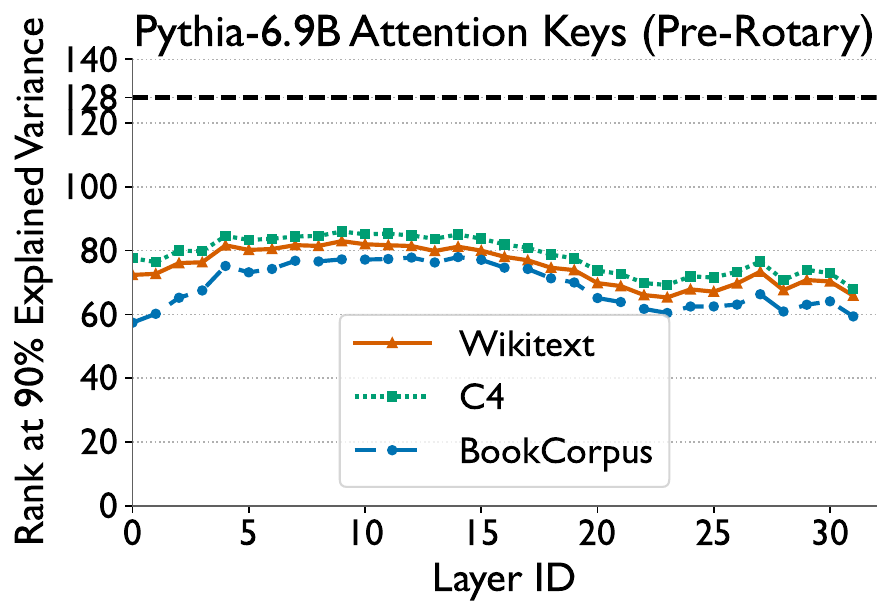}
    \includegraphics[width=0.24\textwidth]{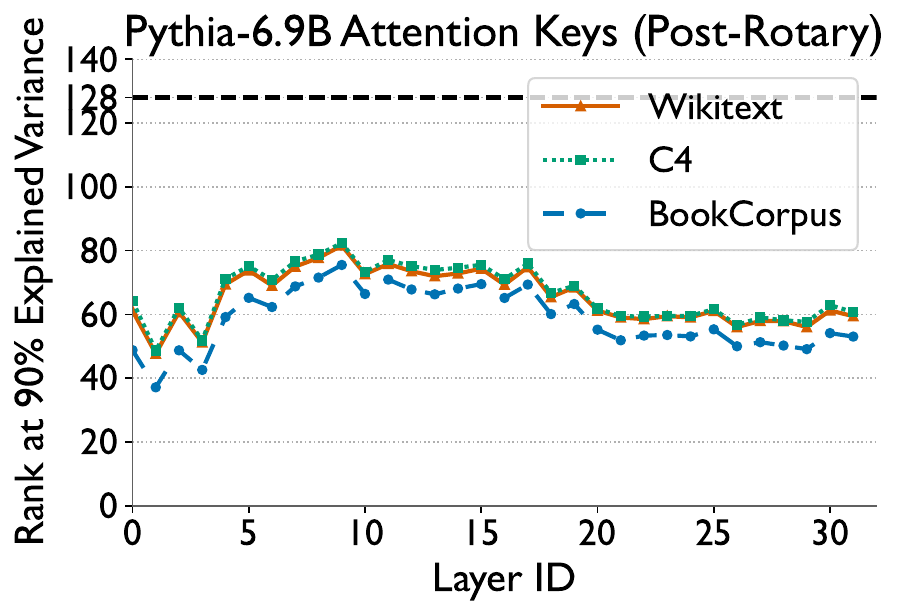}
    \includegraphics[width=0.24\textwidth]{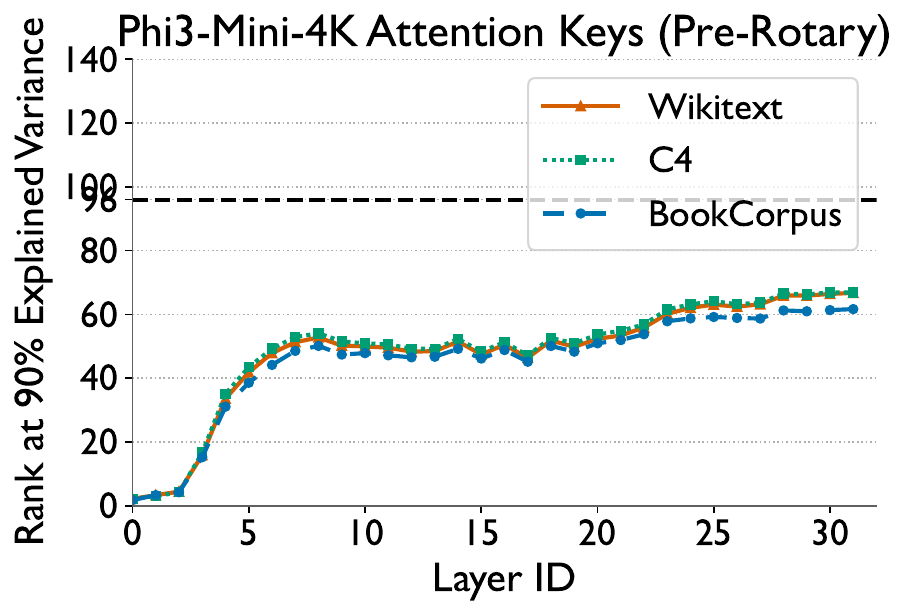}
    \includegraphics[width=0.24\textwidth]{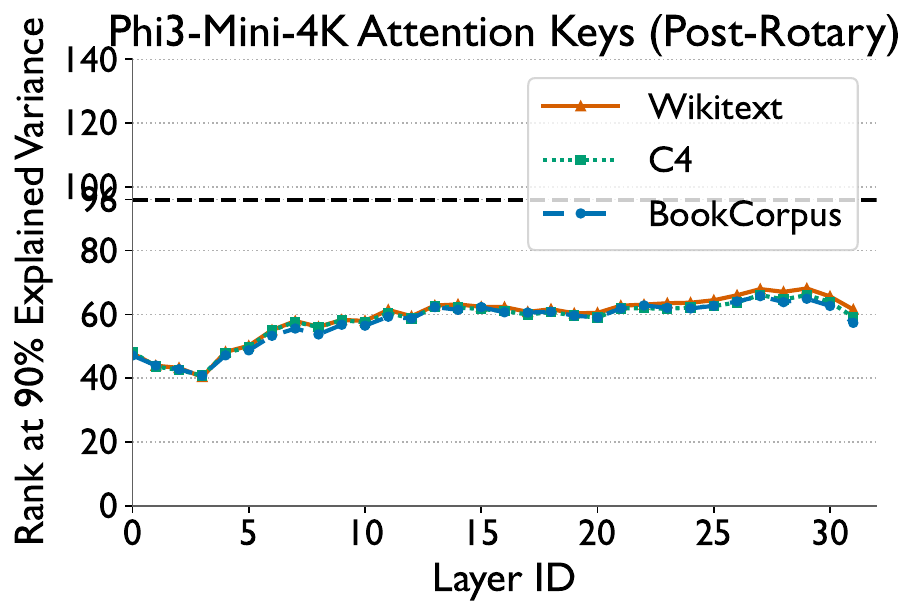}

    \caption{Rank at which 90\% of the variance is explained for pre-rotary and
    post-rotary keys produced by each layer averaged across all heads
    ($Rank_{l}@90$) for different models. We observe that all models exhibit
    significantly low rank consistently across all
    datasets.} \label{fig:rank90_appendix}
\end{figure}

In this section, we present our extended dimensionality analysis results (from
\ref{sec:dimanalysis}) for all the models we experimented with. Figure
\ref{fig:rank90_appendix} displays the $Rank_{l}@90$ values for all models
referenced in Section \ref{sec:dimanalysis}. Our analysis reveals that the low
dimensionality of the keys is consistently observed across all models and
datasets. Results for all models resemble those shown in Figure
\ref{fig:rank90} of the main text. The models we tested cover a wide range of sizes, architecture
types (dense vs. MoE), as well as older and newer architectures trained on
various datasets. Despite these differences, our main observation remains
robust.

An intriguing trend is the variation in $Rank_{l}@90$ across layers for
different models, indicating that the intrinsic dimensionality of the keys is
not uniform across model layers. A potential future direction could be to
investigate the reasons for this variation from a semantic perspective. 

\begin{figure}[h]
  \centering
    \includegraphics[width=0.32\textwidth]{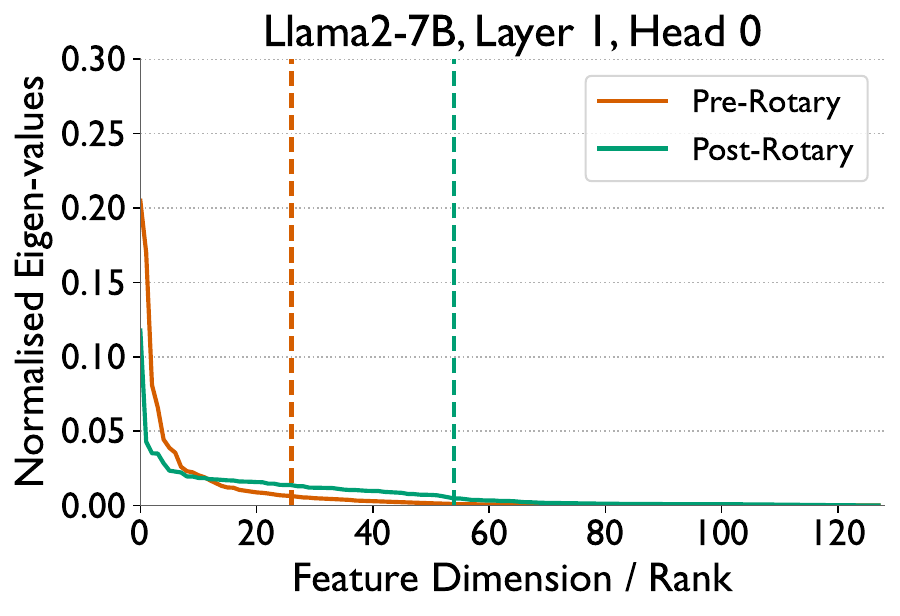}
    \includegraphics[width=0.32\textwidth]{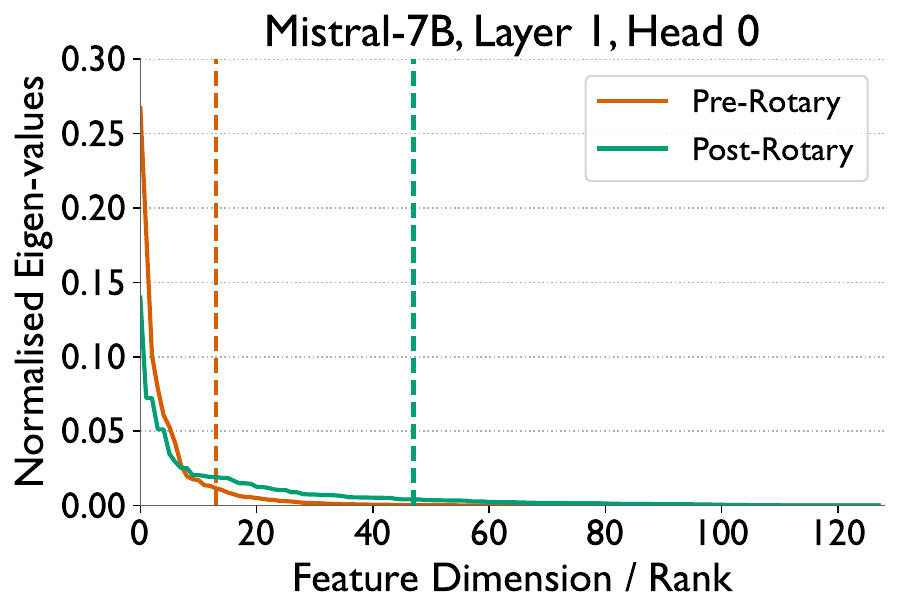}
    \includegraphics[width=0.32\textwidth]{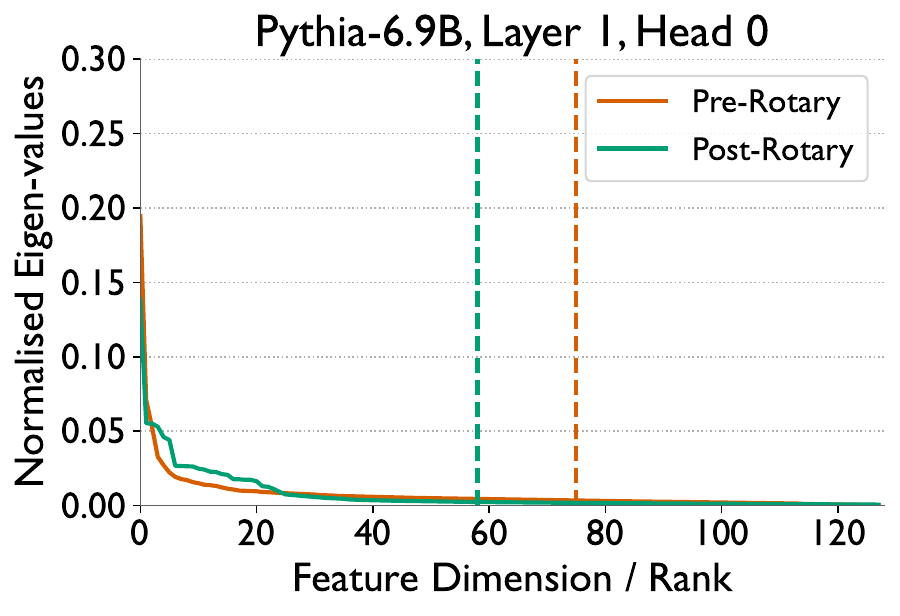}
    \includegraphics[width=0.32\textwidth]{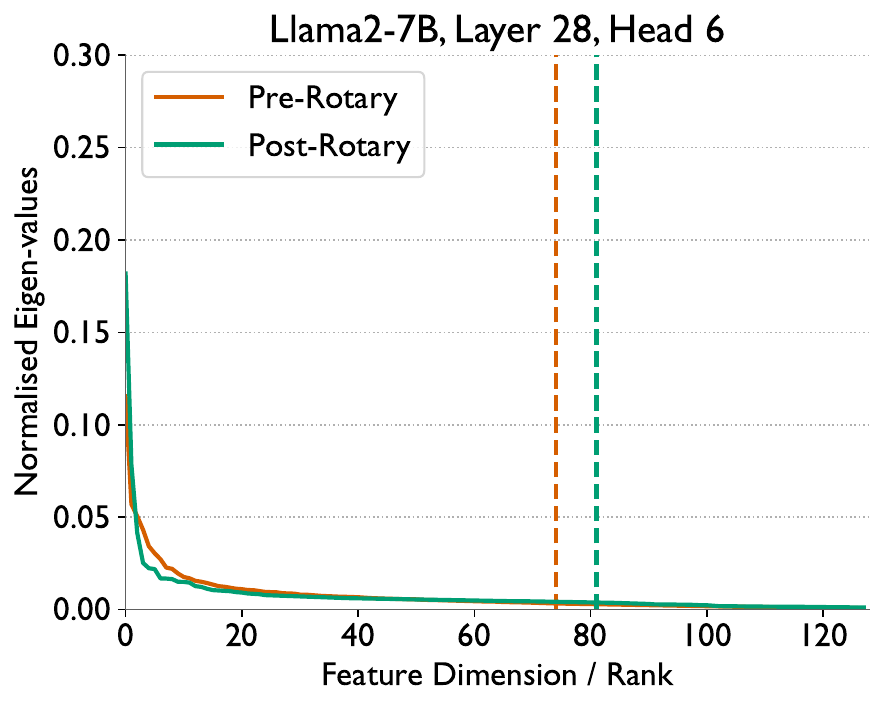}
    \includegraphics[width=0.32\textwidth]{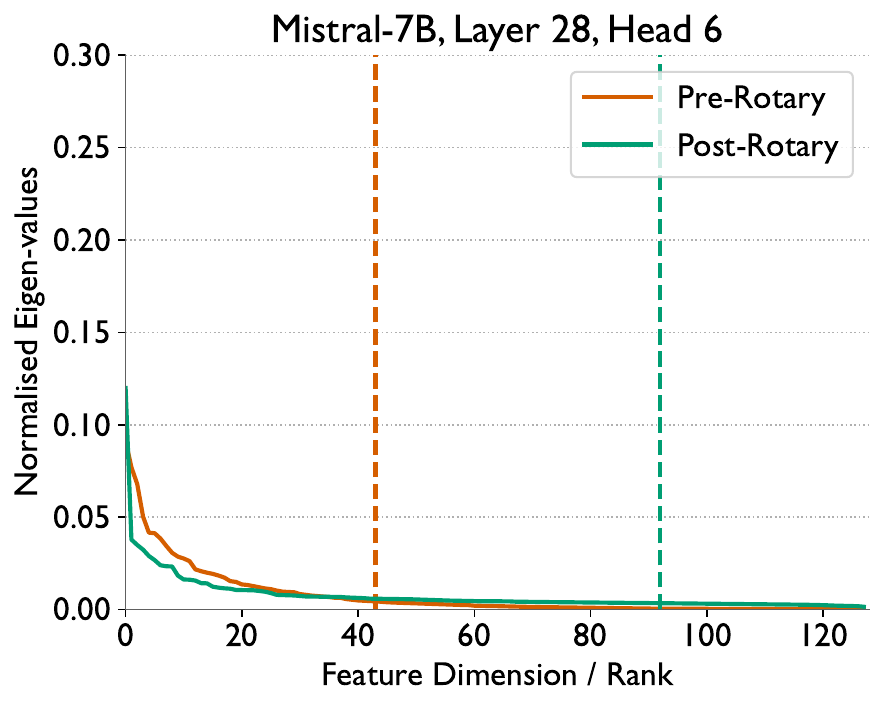}
    \includegraphics[width=0.32\textwidth]{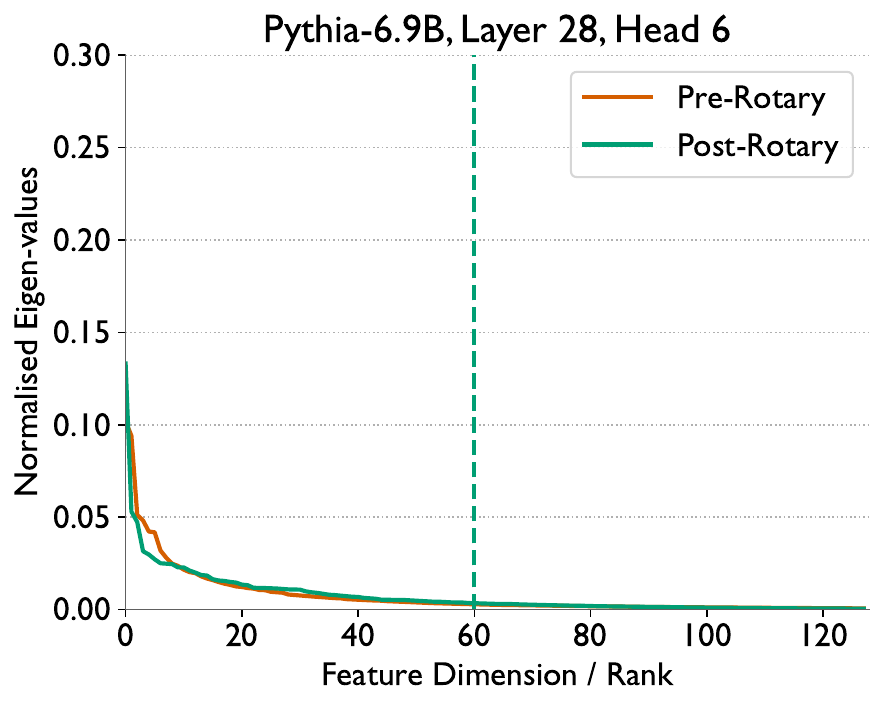}
    \caption{Normalized eigenvalues of the covariance matrix of the keys
    produced by Layer 1, Head 1 (top row), and Layer 28, Head 6 (bottom row)
    of Llama2-7B (left), Mistral-7B (middle), and Pythia-6.9B (right) on the
    WikiText-2 dataset. We observe that the explained variance significantly
    decreases after the initial principal dimensions. The dashed lines represent
    the rank at which 90\% of the variance is explained ($Rank_{i,h}@90$).
    \label{fig:variances_appendix}}
\end{figure}

For a more fine-grained analysis, we plot the normalized eigenvalues of the
covariance matrix of the keys for a few layers and heads of Llama2-7B, Mistral-7B, and Pythia-6.9B on the WikiText-2 dataset as an example 
in Figure \ref{fig:variances_appendix}. Here again, we observe that the explained variance significantly decreases after the initial principal dimensions. 
The results for the other models are similar to the ones shown here.

\subsection{Variation of Rank across Attention Heads}
\begin{figure}[t]
  \centering
  \includegraphics[width=0.95\linewidth]{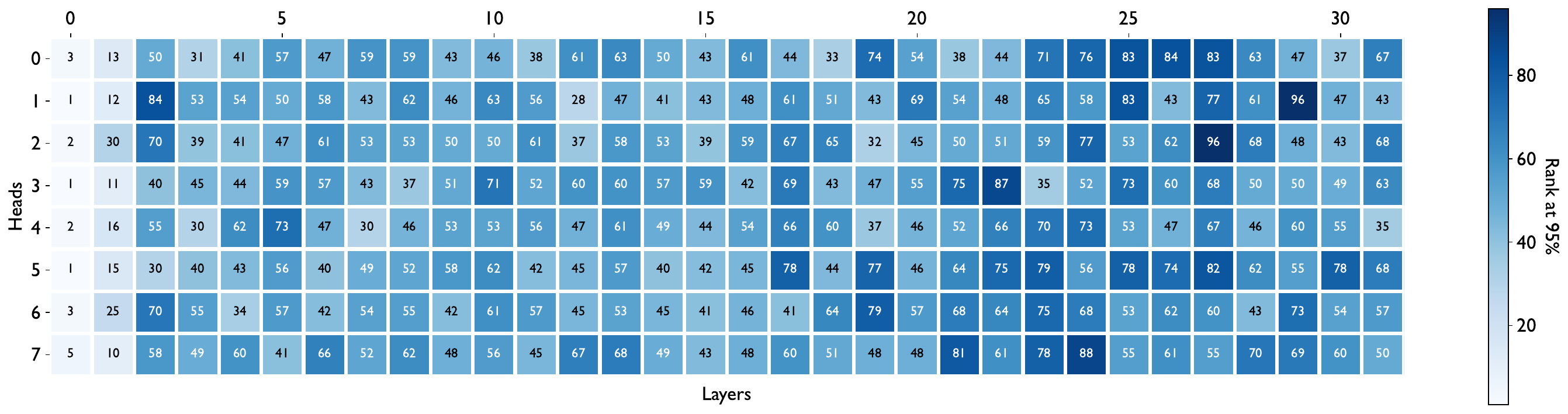}
  \includegraphics[width=0.95\linewidth]{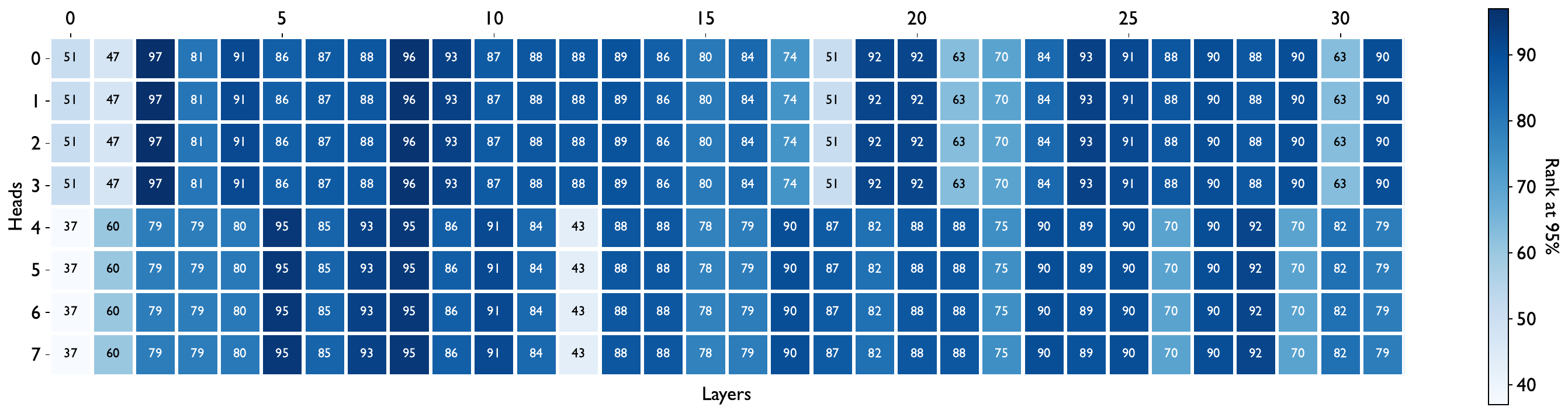}
  \caption{Heatmap showing the rank at 90\% explained variance for the pre-rotary(top) and post-rotary(bottom) 
  key vectors across  all layers and heads for Mistral-7B.}
  \label{fig:heatmap}
\end{figure}

In this section, we discuss the variation of the rank at which 90\% of the variance is explained
($Rank_{l}@90$) across different heads within a layer for two models: Llama2-7B and Mistral-7B.
Figure \ref{fig:heatmap} shows the heatmap of the $Rank_{l}@90$ for the pre-rotary (top) and 
post-rotary (bottom) keys across
all layers and heads for Mistral-7B. We observe that the $Rank_{l}@90$ is considerably lower 
for pre-rotary keys vs post-rotary keys. Focusing on the pre-rotary keys, we see that the 
initial layers have a lower rank compared to the later layers. In each layer, there are some heads
heads with high-rank values even though the median rank is low. This might 
indicate that some head in that layer is more important and uses more complex information 
about the keys. Interestingly for post-rotary keys, we see a pattern where 4 out of the 8 heads
in each layer have the same rank. This might have to do with how the rotary embeddings are
applied to Mistral-7B as we do not see this pattern in Llama2-7B. 

\begin{figure}[t]
  \centering
  \includegraphics[width=0.49\linewidth]{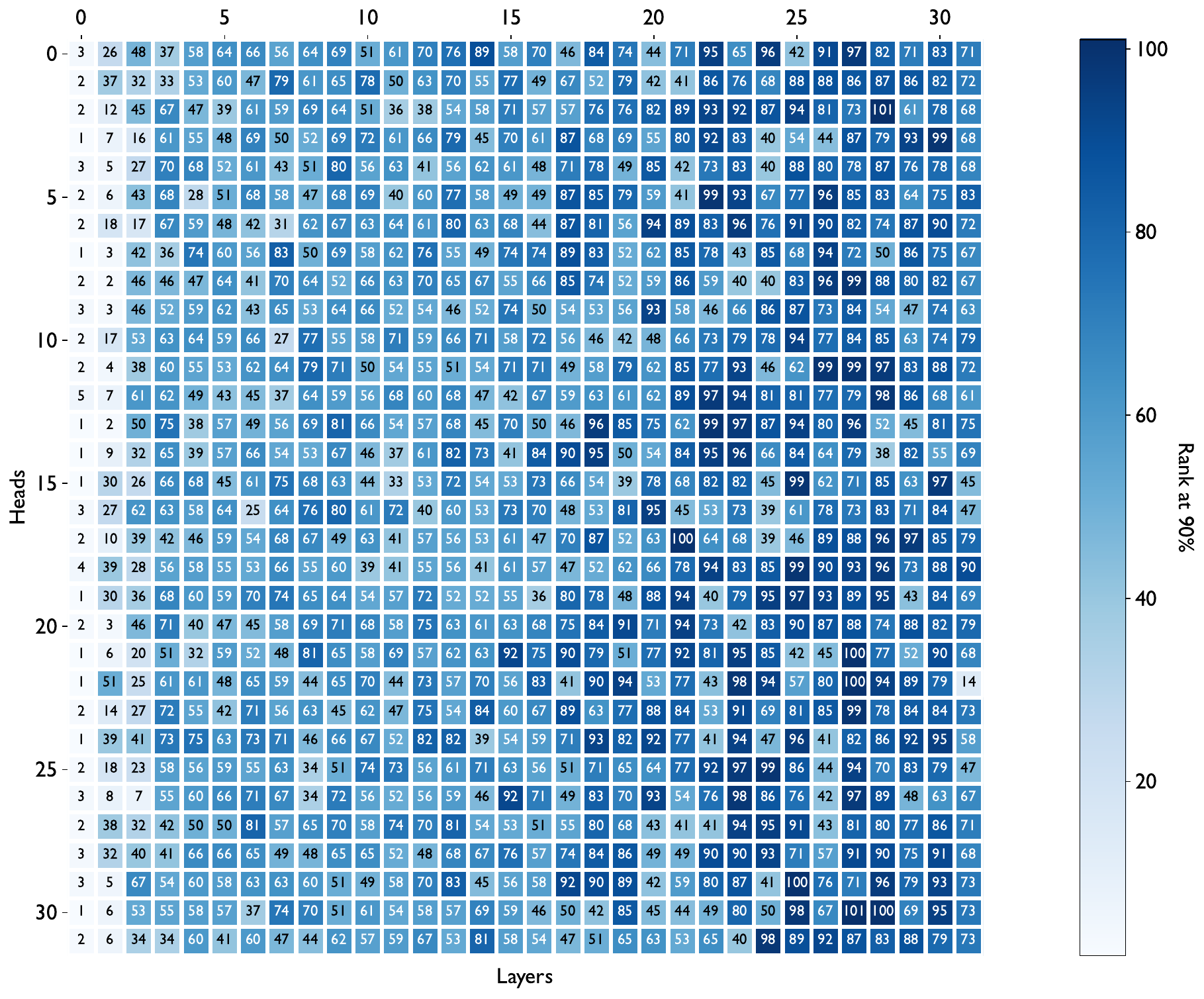}
  \includegraphics[width=0.49\linewidth]{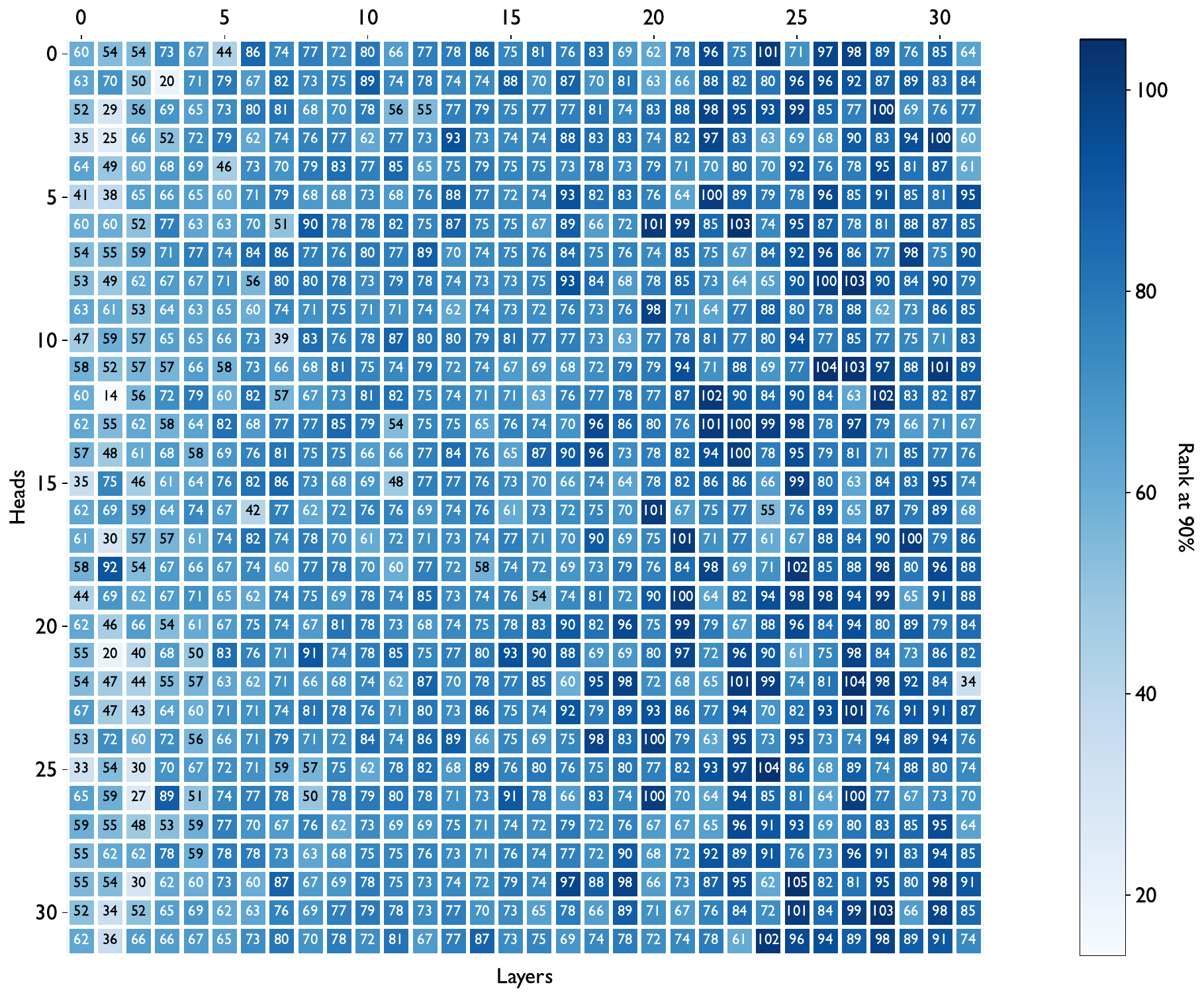}
  \caption{Heatmap showing the rank at 90\% explained variance for the pre-rotary(top) and post-rotary(bottom) 
  key vectors across  all layers and heads for Llama2-7B.}
  \label{fig:heatmap_llama}
\end{figure}

Figure \ref{fig:heatmap_llama} shows the heatmap of the $Rank_{l}@90$ for the pre-rotary (left) and 
post-rotary (right) keys across
all layers and heads for Llama2-7B. We observe a similar trend as Mistral-7B where the initial layers
have a lower rank compared to the later layers. However, we do not see the same pattern in the post-rotary
keys as we saw in Mistral-7B. This might indicate that the rotary embeddings are applied differently in
Llama2-7B compared to Mistral-7B.

In this section, we examine the variation in the rank at which 90\% of the
variance is explained ($Rank_{l}@90$) across different heads within a layer for
two models: Llama2-7B and Mistral-7B. Figure \ref{fig:heatmap} shows the
heatmap of $Rank_{l}@90$ for the pre-rotary (top) and post-rotary (bottom) keys
across all layers and heads for Mistral-7B. We observe that the $Rank_{l}@90$ is
significantly lower for pre-rotary keys compared to post-rotary keys. Focusing
on the pre-rotary keys, it is evident that the initial layers exhibit lower rank
values than the later layers. In each layer, some heads demonstrate high-rank
values, even though the median rank remains low. This suggests that certain
heads in those layers are more important and leverage more complex information
about the keys. Interestingly, for the post-rotary keys, we notice a pattern in which 
4 out of the 8 heads in each layer share the same rank. This phenomenon may be related to
how rotary embeddings are applied in Mistral-7B, as we do not observe this
pattern in Llama2-7B. Further investigation is needed to understand this trend.

Figure \ref{fig:heatmap_llama} illustrates the heatmap of $Rank_{l}@90$ for the
pre-rotary (left) and post-rotary (right) keys across all layers and heads for
Llama2-7B. A similar trend emerges as seen in Mistral-7B, where the initial
layers have lower rank values compared to the later layers. However, the same
pattern in post-rotary keys that was observed in Mistral-7B is absent here,
suggesting that rotary embeddings may be applied differently in Llama2-7B
compared to Mistral-7B.

For both the models, we can see that in each layer, there are some heads with
very high-rank values, even when the median rank is low. This might indicate
that some heads in that layer are more important and use more complex
information about the keys. Analysis into chosing the best reduced
dimensionality based upon the the distribution of ranks across heads could be a
potential future direction.

\subsection{Dimensionality Analysis for Queries and Values}
\begin{figure}[ht]
  \centering
  \includegraphics[width=0.49\linewidth]{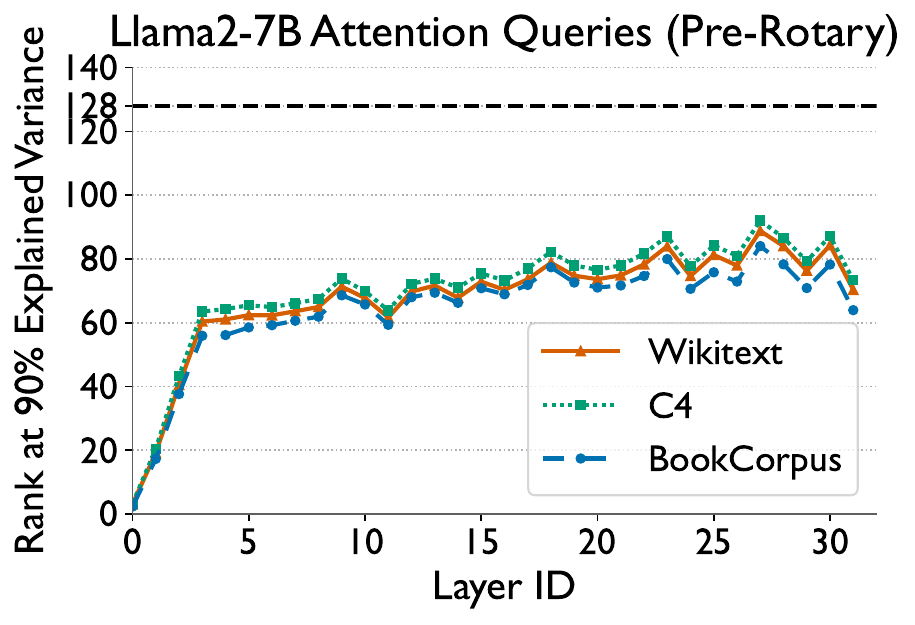}
  \includegraphics[width=0.49\linewidth]{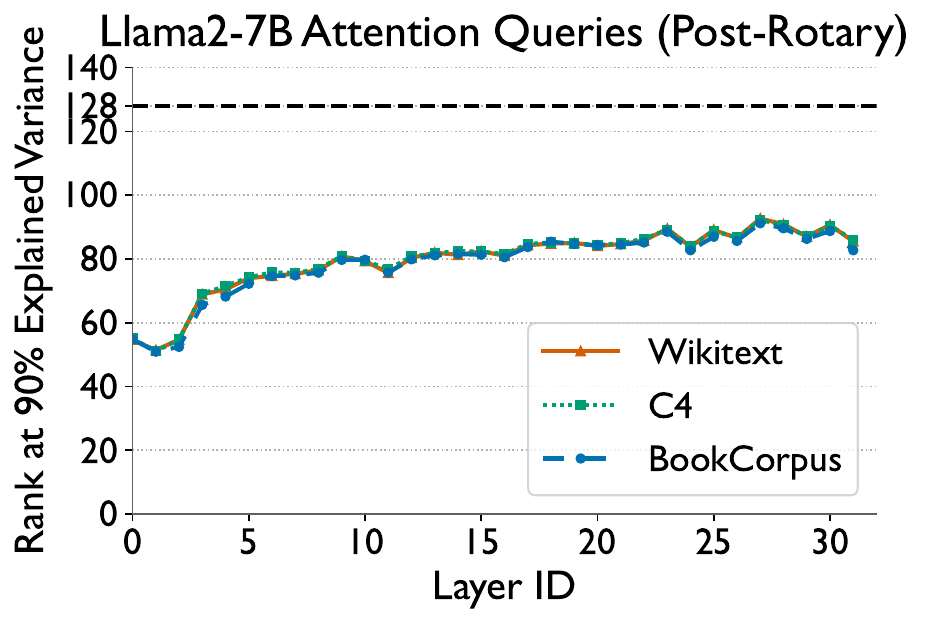}
  \includegraphics[width=0.49\linewidth]{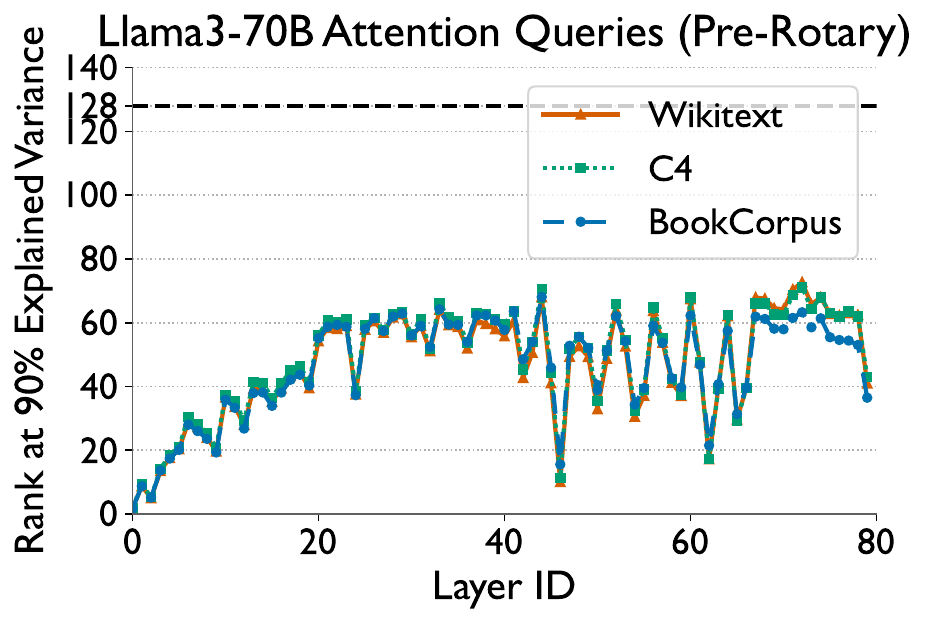}
  \includegraphics[width=0.49\linewidth]{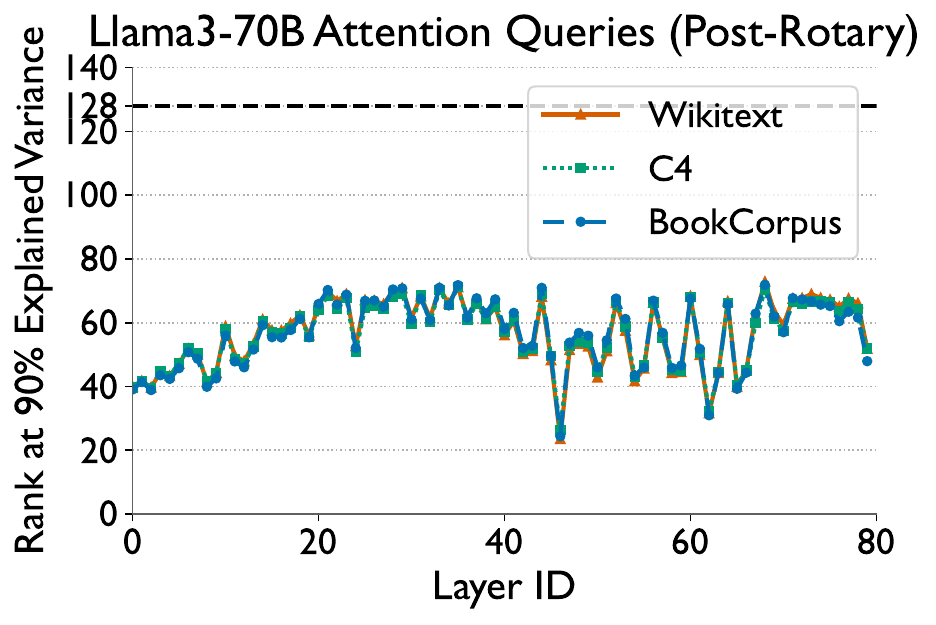}
  \caption{Rank at which 90\% of variance is explained ($Rank_{l}@90$) for the query vectors across various models and datasets.}
  \label{fig:query_dimanalysis}
\end{figure}

\begin{figure}[ht]
  \centering
  \includegraphics[width=0.49\linewidth]{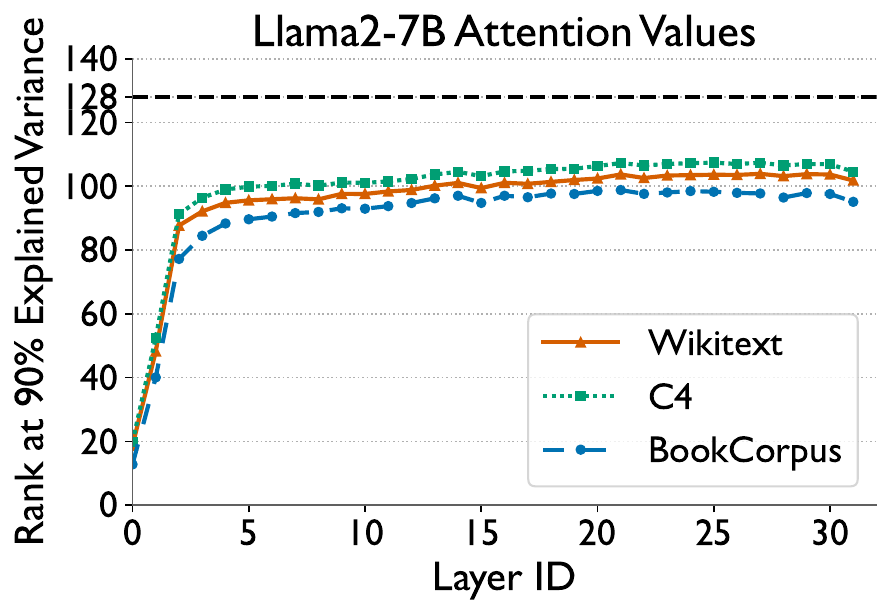}
  \includegraphics[width=0.49\linewidth]{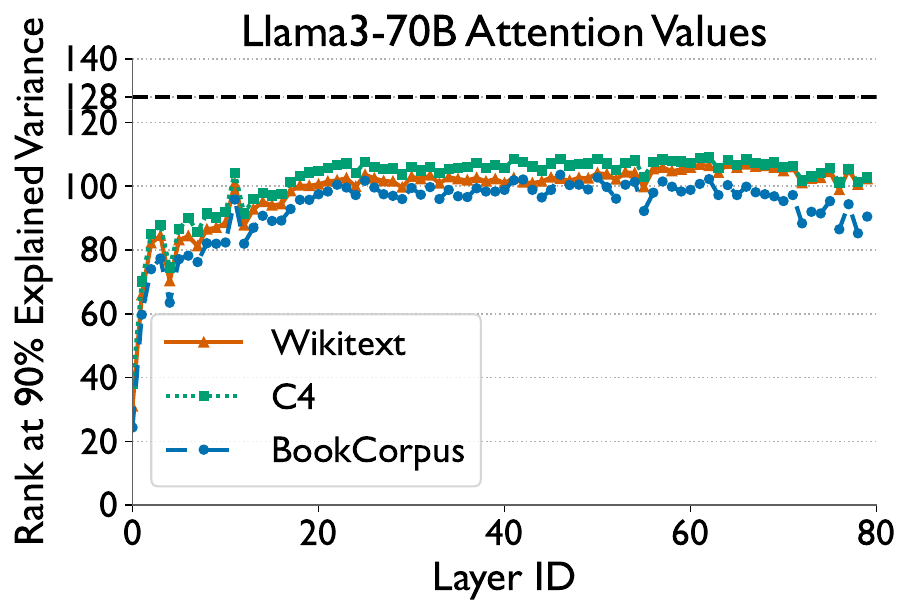}
  \caption{Rank at which 90\% of variance is explained ($Rank_{l}@90$) for the value vectors across various models and datasets.}
  \label{fig:value_dimanalysis}
\end{figure}

While our main focus has been on the dimensionality of the keys, we also
performed exploratory analysis on the dimensionality of the queries and values.
Figures \ref{fig:query_dimanalysis} and \ref{fig:value_dimanalysis} show the
$Rank_{l}@90$ for the queries and values, respectively, for Llama2-7B and
Llama3-70B. We observe that the queries and values also exhibit low
dimensionality across all layers and heads, similar to the keys, while values
tend to have a considerably higher dimensionality and close to the full
dimensionality of the value vectors. This observation can intuitively be
explained by the fact that both keys and queries are used to compute the scalar
attention scores and thus do not need to be high-dimensional, while values are
weighted by these scores and used to compute the final output, and thus need to
be high-dimensional to capture the complexity of the data.

\newpage
\clearpage 

\section{Comprehensive Evaluation Results}
\label{append:comp_pcatopk}

\subsection{Performance of \method~on Perplexity and Short-Context Downstream Tasks}
\begin{figure}[h]
  \centering
  \includegraphics[width=0.93\linewidth]{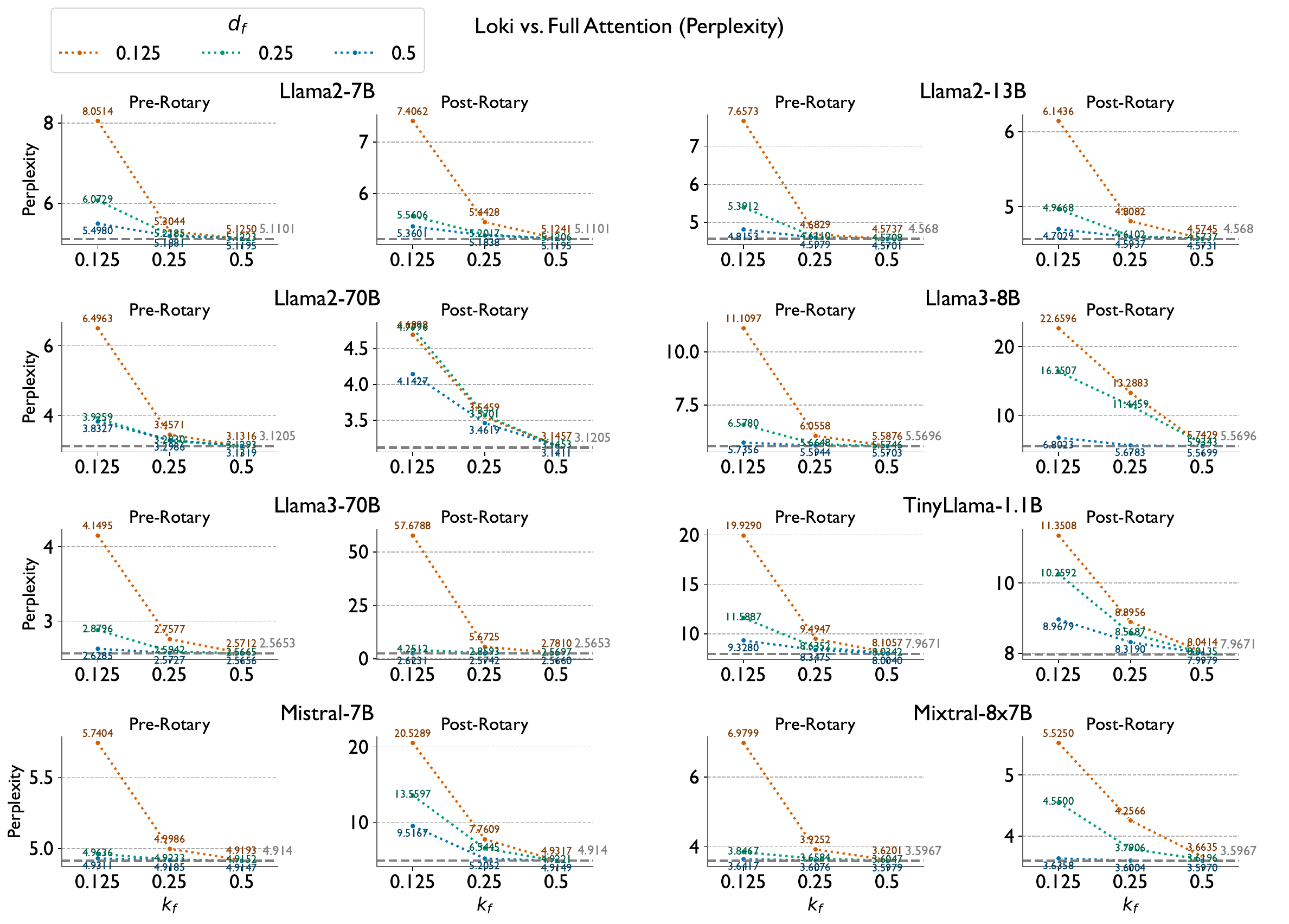}
  \includegraphics[width=0.93\linewidth]{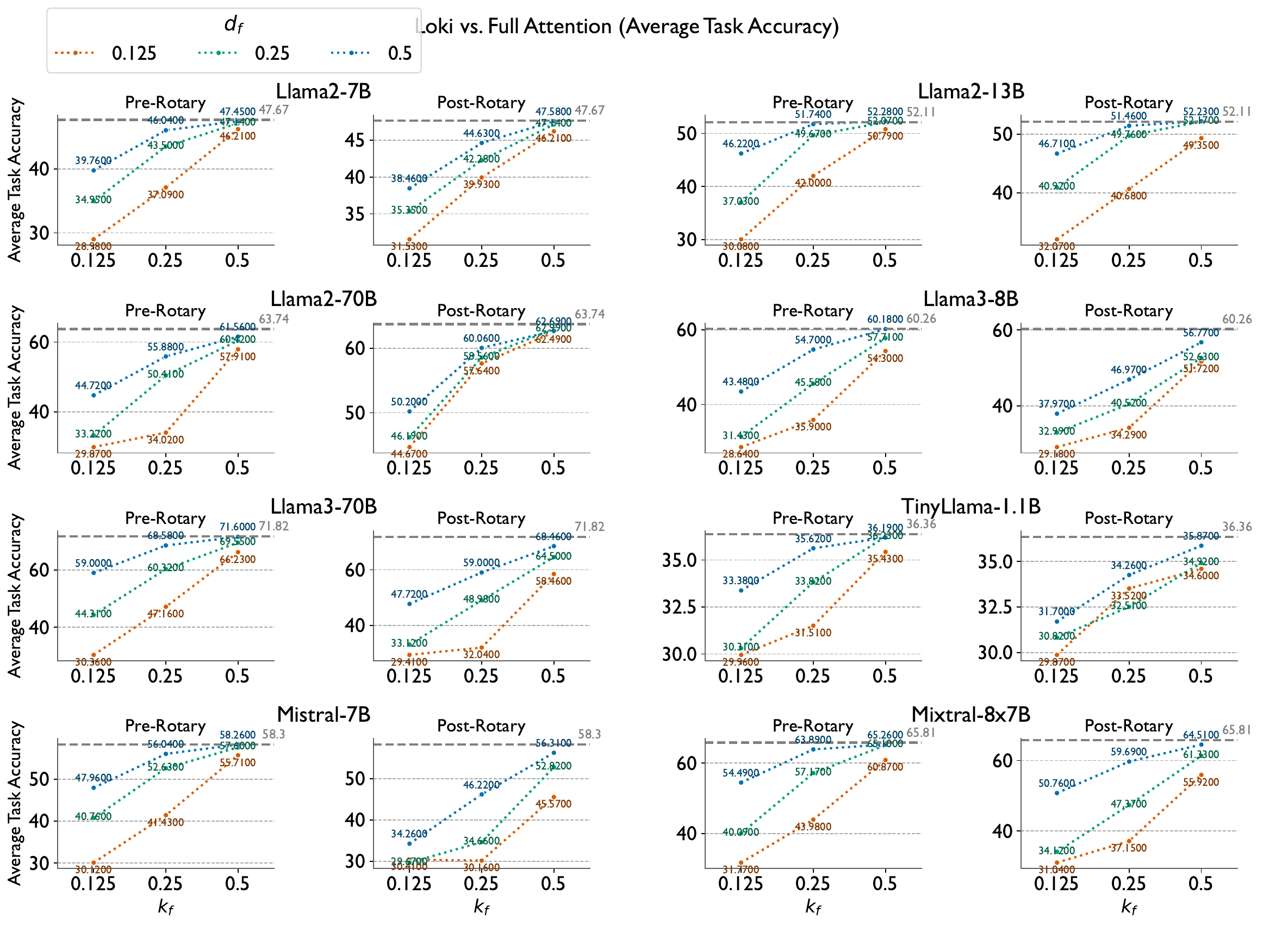}
  \caption{Performance of \method~on Perplexity (top) and Short-Context Downstream Task evaluation for different 
  models using pre-rotary and post-rotary PCA transformation. For each model and each
  transform type, we run \method~with different values of $k$ and $d$.}
  \label{fig:appendix_pcatopk}
\end{figure}


\begin{table}[h!]
  \caption{Performance of different models compared to hugging face baseline with different configurations
  of $k$ and $d$ using pre-rotary PCA transformation.}
  \centering
  \resizebox{0.90\linewidth}{!}{%
  \small 
  \begin{tabular}{@{}lcccccccccccc@{}}
  \toprule
  Model & Method & $k$ & $d$ & PPL$\downarrow$ & Hellaswag$\uparrow$ &TQA$\uparrow$ & Winogrande$\uparrow$ & ARC$\uparrow$ & GSM8K$\uparrow$ & 
  MMLU$\uparrow$ & Avg$\uparrow$ \\
  \toprule
  \toprule
  Llama2-7B & Full Attention &     - &     - &      5.1101 &               75.99 &               38.96 &           69.06 &                   46.33 &                           13.87 &     41.84 & 47.67 \\
  \midrule 
  \multirow{9}{*}{Llama2-7B}
  & \method~&   0.5 &   0.5 &      5.1195 &               75.96 &               38.85 &           69.22 &                   46.16 &                           13.19 &     41.34 & 47.45 \\
  & \method~&   0.5 &  0.25 &      5.1223 &               75.84 &               39.05 &           68.82 &                   45.82 &                           12.36 &     40.95 & 47.14 \\
  & \method~&   0.5 & 0.125 &      5.1250 &               75.09 &               38.51 &           69.53 &                   44.28 &                           10.77 &     39.07 & 46.21 \\
  & \method~&  0.25 &   0.5 &      5.1881 &               75.73 &               38.04 &           67.25 &                   44.20 &                           11.30 &     39.74 & 46.04 \\
  & \method~&  0.25 &  0.25 &      5.2185 &               73.43 &               38.35 &           63.61 &                   41.21 &                            7.96 &     36.43 & 43.50 \\
  & \method~&  0.25 & 0.125 &      5.3044 &               53.23 &               40.08 &           59.35 &                   36.09 &                            2.81 &     30.99 & 37.09 \\
  & \method~& 0.125 &   0.5 &      5.4980 &               70.42 &               39.40 &           52.49 &                   35.92 &                            7.13 &     33.22 & 39.76 \\
  & \method~& 0.125 &  0.25 &      6.0729 &               56.04 &               42.76 &           49.57 &                   31.91 &                            2.27 &     27.15 & 34.95 \\
  & \method~& 0.125 & 0.125 &      8.0514 &               31.06 &               44.46 &           49.01 &                   25.34 &                            0.38 &     23.64 & 28.98 \\
  \toprule
  \toprule
  Llama2-13B & Full Attention &     - &     - &      4.5680 &               79.38 &               36.90 &           72.22 &                   49.15 &                           22.97 &     52.06 & 52.11 \\
  \midrule 
  \multirow{9}{*}{Llama2-13B}
  & \method~&   0.5 &   0.5 &      4.5701 &               79.34 &               37.06 &           73.09 &                   48.81 &                           23.20 &     52.19 & 52.28 \\
  & \method~&   0.5 &  0.25 &      4.5708 &               79.27 &               37.14 &           72.14 &                   49.40 &                           22.44 &     52.03 & 52.07 \\
  & \method~&   0.5 & 0.125 &      4.5737 &               78.45 &               37.39 &           70.09 &                   47.95 &                           19.86 &     50.98 & 50.79 \\
  & \method~&  0.25 &   0.5 &      4.5979 &               79.19 &               37.35 &           71.90 &                   47.87 &                           22.14 &     52.02 & 51.74 \\
  & \method~&  0.25 &  0.25 &      4.6110 &               77.39 &               36.89 &           68.90 &                   46.16 &                           19.86 &     48.80 & 49.67 \\
  & \method~&  0.25 & 0.125 &      4.6829 &               71.17 &               37.21 &           58.17 &                   36.26 &                            7.88 &     41.30 & 42.00 \\
  & \method~& 0.125 &   0.5 &      4.8153 &               77.38 &               38.45 &           56.27 &                   41.64 &                           14.94 &     48.63 & 46.22 \\
  & \method~& 0.125 &  0.25 &      5.3912 &               61.85 &               36.79 &           52.09 &                   32.08 &                            2.96 &     36.40 & 37.03 \\
  & \method~& 0.125 & 0.125 &      7.6573 &               38.67 &               43.00 &           50.20 &                   24.32 &                            0.68 &     23.63 & 30.08 \\
  \toprule
  \toprule
  Llama2-70B & Full Attention &     - &     - &      3.1205 &               83.82 &               44.81 &           77.90 &                   57.34 &                           53.15 &     65.41 & 63.74 \\
  \midrule 
  \multirow{9}{*}{Llama2-70B}
  & \method~&   0.5 &   0.5 &      3.1319 &                   - &                   - &               - &                       - &                               - &         - &     - \\
  & \method~&   0.5 &  0.25 &      3.1293 &               83.65 &               39.78 &           76.95 &                   56.91 &                           41.93 &     63.32 & 60.42 \\
  & \method~&   0.5 & 0.125 &      3.1316 &               82.38 &               39.33 &           72.85 &                   54.61 &                           37.45 &     60.85 & 57.91 \\
  & \method~&  0.25 &   0.5 &      3.2986 &               80.54 &               42.46 &           75.85 &                   57.08 &                           22.21 &     57.14 & 55.88 \\
  & \method~&  0.25 &  0.25 &      3.2830 &               76.05 &               44.88 &           63.54 &                   50.26 &                           15.92 &     51.79 & 50.41 \\
  & \method~&  0.25 & 0.125 &      3.4571 &               52.25 &               44.73 &           50.36 &                   25.09 &                            2.35 &     29.37 & 34.02 \\
  & \method~& 0.125 &   0.5 &      3.8327 &               68.06 &               39.43 &           58.80 &                   46.93 &                           10.31 &     44.82 & 44.72 \\
  & \method~& 0.125 &  0.25 &      3.9259 &               46.59 &               45.88 &           46.96 &                   28.67 &                            2.35 &     28.90 & 33.22 \\
  & \method~& 0.125 & 0.125 &      6.4963 &               30.07 &               49.19 &           51.30 &                   22.78 &                            1.14 &     24.75 & 29.87 \\
  \toprule
  \toprule
  Llama3-8B & Full Attention &     - &     - &      5.5696 &               79.17 &               43.89 &           72.93 &                   53.24 &                           50.11 &     62.19 & 60.26 \\
  \midrule 
  \multirow{9}{*}{Llama3-8B}
  & \method~&   0.5 &   0.5 &      5.5703 &               78.84 &               44.21 &           73.64 &                   54.01 &                           48.90 &     61.47 & 60.18 \\
  & \method~&   0.5 &  0.25 &      5.5746 &               77.44 &               43.68 &           68.27 &                   49.15 &                           47.16 &     60.58 & 57.71 \\
  & \method~&   0.5 & 0.125 &      5.5876 &               74.83 &               44.23 &           65.43 &                   43.94 &                           40.41 &     56.97 & 54.30 \\
  & \method~&  0.25 &   0.5 &      5.5944 &               76.54 &               44.32 &           60.93 &                   43.43 &                           44.66 &     58.33 & 54.70 \\
  & \method~&  0.25 &  0.25 &      5.6648 &               69.42 &               41.50 &           50.36 &                   34.64 &                           33.06 &     44.50 & 45.58 \\
  & \method~&  0.25 & 0.125 &      6.0558 &               56.11 &               42.14 &           50.36 &                   27.13 &                            9.17 &     30.46 & 35.90 \\
  & \method~& 0.125 &   0.5 &      5.7356 &               66.13 &               44.00 &           50.04 &                   28.33 &                           31.77 &     40.61 & 43.48 \\
  & \method~& 0.125 &  0.25 &      6.5780 &               45.14 &               41.00 &           49.33 &                   23.89 &                            3.18 &     26.05 & 31.43 \\
  & \method~& 0.125 & 0.125 &     11.1097 &               32.70 &               44.31 &           47.04 &                   23.29 &                            0.68 &     23.80 & 28.64 \\
  \toprule
  \toprule
  Llama3-70B & Full Attention &     - &     - &      2.5653 &               84.89 &               45.57 &           80.43 &                   64.33 &                           80.67 &     75.03 & 71.82 \\
  \midrule 
  \multirow{9}{*}{Llama3-70B}
  & \method~&   0.5 &   0.5 &      2.5656 &               85.17 &               45.66 &           79.95 &                   63.99 &                           79.91 &     74.90 & 71.60 \\
  & \method~&   0.5 &  0.25 &      2.5665 &               84.22 &               45.78 &           75.06 &                   59.81 &                           78.77 &     73.68 & 69.55 \\
  & \method~&   0.5 & 0.125 &      2.5712 &               82.21 &               45.53 &           69.61 &                   54.78 &                           74.98 &     70.28 & 66.23 \\
  & \method~&  0.25 &   0.5 &      2.5727 &               84.09 &               45.64 &           71.35 &                   57.51 &                           79.76 &     73.12 & 68.58 \\
  & \method~&  0.25 &  0.25 &      2.5942 &               79.06 &               45.09 &           59.27 &                   43.26 &                           72.78 &     62.47 & 60.32 \\
  & \method~&  0.25 & 0.125 &      2.7577 &               67.59 &               45.46 &           50.67 &                   31.48 &                           45.56 &     42.21 & 47.16 \\
  & \method~& 0.125 &   0.5 &      2.6285 &               78.96 &               46.48 &           51.14 &                   40.70 &                           74.53 &     62.19 & 59.00 \\
  & \method~& 0.125 &  0.25 &      2.8796 &               63.93 &               41.69 &           46.33 &                   27.65 &                           50.19 &     36.08 & 44.31 \\
  & \method~& 0.125 & 0.125 &      4.1495 &               39.07 &               41.09 &           49.88 &                   23.38 &                            3.03 &     25.73 & 30.36 \\
  \toprule
  \toprule
  TinyLlama-1.1B & Full Attention &     - &     - &      7.9671 &               60.45 &               37.88 &           60.22 &                   32.85 &                            1.90 &     24.86 & 36.36 \\
  \midrule 
  \multirow{9}{*}{TinyLlama-1.1B}
  & \method~&   0.5 &   0.5 &      8.0040 &               60.39 &               38.19 &           59.98 &                   32.08 &                            1.90 &     24.62 & 36.19 \\
  & \method~&   0.5 &  0.25 &      8.0342 &               59.96 &               38.80 &           59.27 &                   32.85 &                            2.20 &     24.33 & 36.23 \\
  & \method~&   0.5 & 0.125 &      8.1057 &               57.93 &               39.10 &           57.14 &                   31.91 &                            1.52 &     24.98 & 35.43 \\
  & \method~&  0.25 &   0.5 &      8.3475 &               58.06 &               40.05 &           58.17 &                   31.06 &                            1.52 &     24.83 & 35.62 \\
  & \method~&  0.25 &  0.25 &      8.6352 &               52.69 &               42.96 &           52.01 &                   29.18 &                            1.29 &     24.76 & 33.82 \\
  & \method~&  0.25 & 0.125 &      9.4947 &               44.43 &               44.21 &           50.75 &                   23.89 &                            1.44 &     24.34 & 31.51 \\
  & \method~& 0.125 &   0.5 &      9.3280 &               51.29 &               42.27 &           53.91 &                   27.82 &                            0.83 &     24.17 & 33.38 \\
  & \method~& 0.125 &  0.25 &     11.5887 &               37.32 &               47.04 &           47.51 &                   25.00 &                            1.52 &     23.49 & 30.31 \\
  & \method~& 0.125 & 0.125 &     19.9290 &               30.13 &               48.50 &           51.30 &                   24.66 &                            1.06 &     24.11 & 29.96 \\
  \toprule
  \toprule
  Mistral-7B & Full Attention &     - &     - &      4.9140 &               81.07 &               42.62 &           73.95 &                   53.92 &                           38.59 &     59.65 & 58.30 \\
  \midrule
  \multirow{9}{*}{Mistral-7B}
  & \method~&   0.5 &   0.5 &      4.9147 &               80.84 &               42.99 &           74.27 &                   53.58 &                           38.06 &     59.83 & 58.26 \\
  & \method~&   0.5 &  0.25 &      4.9152 &               80.55 &               43.11 &           72.69 &                   53.41 &                           36.69 &     59.14 & 57.60 \\
  & \method~&   0.5 & 0.125 &      4.9193 &               79.38 &               42.29 &           70.40 &                   51.28 &                           33.59 &     57.29 & 55.71 \\
  & \method~&  0.25 &   0.5 &      4.9185 &               79.00 &               43.41 &           70.17 &                   49.23 &                           36.16 &     58.25 & 56.04 \\
  & \method~&  0.25 &  0.25 &      4.9233 &               77.65 &               42.18 &           62.98 &                   46.59 &                           32.68 &     53.70 & 52.63 \\
  & \method~&  0.25 & 0.125 &      4.9986 &               66.95 &               39.58 &           52.64 &                   36.35 &                           14.86 &     38.20 & 41.43 \\
  & \method~& 0.125 &   0.5 &      4.9311 &               72.66 &               43.89 &           52.25 &                   35.58 &                           33.36 &     50.01 & 47.96 \\
  & \method~& 0.125 &  0.25 &      4.9636 &               65.93 &               41.12 &           51.78 &                   29.18 &                           18.42 &     38.14 & 40.76 \\
  & \method~& 0.125 & 0.125 &      5.7404 &               36.32 &               43.14 &           52.17 &                   23.98 &                            0.53 &     24.60 & 30.12 \\
  \toprule
  \toprule
  Mixtral-8x7B & Full Attention &     - &     - &      3.5967 &               84.01 &               48.53 &           76.32 &                   59.73 &                           58.38 &     67.90 & 65.81 \\
  \midrule
  \multirow{9}{*}{Mixtral-8x7B}
  & \method~&   0.5 &   0.5 &      3.5979 &               83.86 &               46.86 &           75.53 &                   60.15 &                           57.32 &     67.83 & 65.26 \\
  & \method~&   0.5 &  0.25 &      3.6047 &               83.70 &               46.70 &           76.24 &                   59.73 &                           57.01 &     67.21 & 65.10 \\
  & \method~&   0.5 & 0.125 &      3.6201 &               82.91 &               42.27 &           73.48 &                   57.42 &                           43.44 &     65.71 & 60.87 \\
  & \method~&  0.25 &   0.5 &      3.6076 &               82.58 &               48.16 &           71.43 &                   58.28 &                           56.18 &     66.72 & 63.89 \\
  & \method~&  0.25 &  0.25 &      3.6584 &               81.32 &               43.49 &           62.83 &                   51.79 &                           42.76 &     60.82 & 57.17 \\
  & \method~&  0.25 & 0.125 &      3.9252 &               73.16 &               39.49 &           56.04 &                   44.80 &                            4.85 &     45.55 & 43.98 \\
  & \method~& 0.125 &   0.5 &      3.6417 &               76.93 &               48.21 &           50.91 &                   41.72 &                           50.87 &     58.30 & 54.49 \\
  & \method~& 0.125 &  0.25 &      3.8467 &               70.07 &               37.88 &           49.17 &                   32.68 &                           11.52 &     39.23 & 40.09 \\
  & \method~& 0.125 & 0.125 &      6.9799 &               42.34 &               43.80 &           54.38 &                   24.66 &                            0.45 &     24.99 & 31.77 \\
  \toprule
  \end{tabular}%
  }
  \label{table:comprehensive_results_prerotary}
\end{table}

\begin{table}[h!]
  \caption{Performance of different models compared to hugging face baseline with different configurations
  of $k$ and $d$ using post-rotary PCA transformation.}
  \centering
  \resizebox{0.9\linewidth}{!}{%
  \small 
  \begin{tabular}{@{}lcccccccccccc@{}}
  \toprule
  Model & Method & $k$ & $d$ & PPL$\downarrow$ & Hellaswag$\uparrow$ &TQA$\uparrow$ & Winogrande$\uparrow$ & ARC$\uparrow$ & GSM8K$\uparrow$ & 
  MMLU$\uparrow$ & Avg$\uparrow$ \\
  \toprule
  \toprule
  Llama2-7B & Full Attention &     - &     - &      5.1101 &               75.99 &               38.96 &           69.06 &                   46.33 &                           13.87 &     41.84 & 47.67 \\
  \midrule 
  \multirow{9}{*}{Llama2-7B}
      & \method~&   0.5 &   0.5 &      5.1195 &               75.91 &               38.87 &           68.59 &                   46.50 &                           14.10 &     41.49 & 47.58 \\
      & \method~&   0.5 &  0.25 &      5.1206 &               75.84 &               39.05 &           68.82 &                   45.82 &                           12.36 &     40.95 & 47.14 \\
      & \method~&   0.5 & 0.125 &      5.1241 &               75.48 &               38.77 &           67.64 &                   43.94 &                           12.59 &     38.85 & 46.21 \\
      & \method~&  0.25 &   0.5 &      5.1838 &               75.19 &               38.16 &           62.12 &                   41.21 &                           10.69 &     40.42 & 44.63 \\
      & \method~&  0.25 &  0.25 &      5.2017 &               72.59 &               39.16 &           56.59 &                   37.37 &                           10.24 &     37.74 & 42.28 \\
      & \method~&  0.25 & 0.125 &      5.4428 &               68.49 &               38.83 &           56.51 &                   32.17 &                           10.92 &     32.68 & 39.93 \\
      & \method~& 0.125 &   0.5 &      5.3601 &               70.42 &               39.40 &           52.49 &                   35.92 &                            7.13 &     33.22 & 39.76 \\
      & \method~& 0.125 &  0.25 &      5.5606 &               59.98 &               41.72 &           48.86 &                   26.96 &                            6.22 &     28.38 & 35.35 \\
      & \method~& 0.125 & 0.125 &      7.4062 &               40.14 &               43.84 &           49.64 &                   25.43 &                            5.99 &     24.13 & 31.53 \\
  \toprule
  \toprule
  Llama2-13B & Full Attention & - & - & 4.5680 & 79.38 & 36.90 & 72.22 & 49.15 & 22.97 & 52.06 & 52.78 \\
  \midrule 
  \multirow{9}{*}{Llama2-13B}
      & \method~&   0.5 &   0.5 &      4.5731 &               79.34 &               37.06 &           73.09 &                   48.81 &                           23.20 &     52.19 & 52.28 \\
      & \method~&   0.5 &  0.25 &      4.5737 &               79.05 &               37.46 &           72.69 &                   48.29 &                           23.58 &     51.94 & 52.17 \\
      & \method~&   0.5 & 0.125 &      4.5745 &               78.45 &               37.39 &           70.09 &                   47.95 &                           19.86 &     50.98 & 50.79 \\
      & \method~&  0.25 &   0.5 &      4.5937 &               79.19 &               37.35 &           71.90 &                   47.87 &                           22.14 &     52.02 & 51.74 \\
      & \method~&  0.25 &  0.25 &      4.6102 &               77.39 &               36.89 &           68.90 &                   46.16 &                           19.86 &     48.80 & 49.67 \\
      & \method~&  0.25 & 0.125 &      4.8082 &               61.52 &               38.10 &           52.41 &                   26.54 &                           20.85 &     44.69 & 40.68 \\
      & \method~& 0.125 &   0.5 &      4.7029 &               77.38 &               38.45 &           56.27 &                   41.64 &                           14.94 &     48.63 & 46.22 \\
      & \method~& 0.125 &  0.25 &      4.9668 &               72.71 &               40.09 &           51.14 &                   33.28 &                            9.10 &     39.20 & 40.92 \\
      & \method~& 0.125 & 0.125 &      6.1436 &               34.81 &               46.07 &           52.09 &                   24.15 &                            8.79 &     26.50 & 32.07 \\
  \toprule
  \toprule
  Llama2-70B & Full Attention &     - &     - &      3.1205 &               83.82 &               44.81 &           77.90 &                   57.34 &                           53.15 &     65.41 & 63.74 \\
  \midrule 
  \multirow{9}{*}{Llama2-70B}
      &\method~&   0.5 &   0.5 &      3.1411 &               83.89 &               41.32 &           78.06 &                   57.68 &                           50.42 &     64.75 & 62.69 \\
      &\method~&   0.5 &  0.25 &      3.1453 &               83.69 &               43.42 &           76.80 &                   56.31 &                           52.99 &     64.73 & 62.99 \\
      &\method~&   0.5 & 0.125 &      3.1457 &               83.41 &               43.51 &           75.45 &                   55.89 &                           52.54 &     64.12 & 62.49 \\
      &\method~&  0.25 &   0.5 &      3.4619 &               82.36 &               41.91 &           76.87 &                   56.48 &                           42.61 &     60.11 & 60.06 \\
      &\method~&  0.25 &  0.25 &      3.5701 &               81.42 &               45.26 &           71.11 &                   49.74 &                           44.28 &     59.56 & 58.56 \\
      &\method~&  0.25 & 0.125 &      3.5459 &               80.59 &               45.57 &           65.59 &                   49.15 &                           46.93 &     58.00 & 57.64 \\
      &\method~& 0.125 &   0.5 &      4.1427 &               71.90 &               44.59 &           58.09 &                   41.98 &                           34.34 &     50.30 & 50.20 \\
      &\method~& 0.125 &  0.25 &      4.7796 &               67.03 &               46.58 &           51.85 &                   32.17 &                           37.30 &     42.21 & 46.19 \\
      &\method~& 0.125 & 0.125 &      4.6898 &               64.84 &               44.89 &           50.51 &                   29.95 &                           38.74 &     39.08 & 44.67 \\
  \toprule
  \toprule
  Llama3-8B  & Full Attention &     - &     - &      5.5696 &               79.17 &               43.89 &           72.93 &                   53.24 &                           50.11 &     62.19 & 60.26 \\
  \midrule 
  \multirow{9}{*}{Llama3-8B}
      & \method~&   0.5 &   0.5 &      5.5699 &               76.03 &               43.83 &           67.32 &                   44.71 &                           49.36 &     59.38 & 56.77 \\
      & \method~&   0.5 &  0.25 &      5.9343 &               72.55 &               42.67 &           61.64 &                   39.93 &                           41.09 &     57.88 & 52.63 \\
      & \method~&   0.5 & 0.125 &      5.7429 &               71.38 &               43.16 &           58.64 &                   40.61 &                           39.42 &     57.14 & 51.72 \\
      & \method~&  0.25 &   0.5 &      5.6783 &               68.02 &               42.07 &           48.78 &                   31.31 &                           43.59 &     48.06 & 46.97 \\
      & \method~&  0.25 &  0.25 &     11.4459 &               57.39 &               42.13 &           48.70 &                   27.90 &                           28.28 &     38.69 & 40.52 \\
      & \method~&  0.25 & 0.125 &     13.2883 &               48.99 &               42.10 &           48.07 &                   22.87 &                           12.81 &     30.90 & 34.29 \\
      & \method~& 0.125 &   0.5 &      6.8023 &               49.68 &               41.14 &           49.25 &                   25.51 &                           31.39 &     30.86 & 37.97 \\
      & \method~& 0.125 &  0.25 &     16.3507 &               36.39 &               43.62 &           50.04 &                   25.09 &                           16.60 &     26.21 & 32.99 \\
      & \method~& 0.125 & 0.125 &     22.6596 &               31.60 &               46.38 &           49.25 &                   23.12 &                            1.14 &     23.61 & 29.18 \\
  \toprule
  \toprule
  Llama3-70B & Full Attention &     - &     - &      2.5653 &               84.89 &               45.57 &           80.43 &                   64.33 &                           80.67 &     75.03 & 71.82 \\
  \midrule 
  \multirow{9}{*}{Llama3-70B}
      & \method~&   0.5 &   0.5 &      2.5660 &               83.60 &               45.83 &           72.61 &                   56.14 &                           79.15 &     73.43 & 68.46 \\
      & \method~&   0.5 &  0.25 &      2.5697 &               79.92 &               46.22 &           62.90 &                   48.46 &                           78.39 &     71.11 & 64.50 \\
      & \method~&   0.5 & 0.125 &      2.7810 &               76.66 &               46.77 &           59.27 &                   42.66 &                           56.94 &     68.48 & 58.46 \\
      & \method~&  0.25 &   0.5 &      2.5742 &               74.91 &               47.67 &           51.54 &                   38.05 &                           77.71 &     64.14 & 59.00 \\
      & \method~&  0.25 &  0.25 &      2.8593 &               61.40 &               47.86 &           48.38 &                   27.73 &                           67.32 &     41.18 & 48.98 \\
      & \method~&  0.25 & 0.125 &      5.6725 &               41.90 &               47.18 &           47.59 &                   23.29 &                            5.31 &     26.98 & 32.04 \\
      & \method~& 0.125 &   0.5 &      2.6231 &               56.24 &               43.91 &           50.51 &                   24.66 &                           72.48 &     38.52 & 47.72 \\
      & \method~& 0.125 &  0.25 &      4.2512 &               31.91 &               47.39 &           50.43 &                   24.57 &                           19.71 &     24.72 & 33.12 \\
      & \method~& 0.125 & 0.125 &     57.6788 &               27.01 &               49.28 &           50.67 &                   24.06 &                            0.68 &     24.76 & 29.41 \\
  \toprule
  \toprule
  TinyLlama-1.1B & Full Attention &     - &     - &      7.9671 &               60.45 &               37.88 &           60.22 &                   32.85 &                            1.90 &     24.86 & 36.36 \\
  \midrule 
  \multirow{9}{*}{TinyLlama-1.1B}
      & \method~&   0.5 &   0.5 &      7.9979 &               60.17 &               38.14 &           58.33 &                   31.57 &                            1.90 &     25.12 & 35.87 \\
      & \method~&   0.5 &  0.25 &      8.0135 &               58.78 &               39.95 &           54.38 &                   30.55 &                            1.29 &     24.58 & 34.92 \\
      & \method~&   0.5 & 0.125 &      8.0414 &               57.77 &               38.20 &           54.93 &                   30.89 &                            1.44 &     24.39 & 34.60 \\
      & \method~&  0.25 &   0.5 &      8.3190 &               57.35 &               37.87 &           53.83 &                   29.69 &                            1.67 &     25.13 & 34.26 \\
      & \method~&  0.25 &  0.25 &      8.5687 &               52.40 &               40.86 &           49.33 &                   26.96 &                            2.20 &     23.34 & 32.51 \\
      & \method~&  0.25 & 0.125 &      8.8956 &               51.19 &               42.07 &           52.96 &                   28.92 &                            0.91 &     25.10 & 33.52 \\
      & \method~& 0.125 &   0.5 &      8.9679 &               51.32 &               38.24 &           50.20 &                   24.23 &                            1.29 &     24.92 & 31.70 \\
      & \method~& 0.125 &  0.25 &     10.2592 &               42.85 &               39.06 &           51.85 &                   25.60 &                            1.52 &     24.06 & 30.82 \\
      & \method~& 0.125 & 0.125 &     11.3508 &               39.27 &               41.55 &           50.67 &                   22.78 &                            0.45 &     24.50 & 29.87 \\
  \toprule
  \toprule
  Mistral-7B & Full Attention &     - &     - &      4.9140 &               81.07 &               42.62 &           73.95 &                   53.92 &                           38.59 &     59.65 & 58.30 \\
  \midrule
  \multirow{9}{*}{Mistral-7B}
      & \method~&   0.5 &   0.5 &      4.9149 &               79.89 &               42.15 &           70.56 &                   49.83 &                           37.45 &     58.00 & 56.31 \\
      & \method~&   0.5 &  0.25 &      4.9221 &               78.99 &               40.84 &           63.06 &                   45.48 &                           33.43 &     55.15 & 52.82 \\
      & \method~&   0.5 & 0.125 &      4.9317 &               73.88 &               40.58 &           57.06 &                   33.87 &                           22.06 &     45.95 & 45.57 \\
      & \method~&  0.25 &   0.5 &      5.2052 &               71.86 &               40.74 &           56.04 &                   38.91 &                           24.18 &     45.56 & 46.22 \\
      & \method~&  0.25 &  0.25 &      6.5445 &               62.62 &               38.93 &           48.62 &                   25.17 &                            1.82 &     30.80 & 34.66 \\
      & \method~&  0.25 & 0.125 &      7.7609 &               35.51 &               43.67 &           53.20 &                   23.63 &                            1.06 &     23.86 & 30.16 \\
      & \method~& 0.125 &   0.5 &      9.5167 &               51.73 &               45.44 &           51.62 &                   25.77 &                            3.03 &     27.99 & 34.26 \\
      & \method~& 0.125 &  0.25 &     13.5597 &               34.85 &               46.38 &           50.20 &                   22.53 &                            0.45 &     23.60 & 29.67 \\
      & \method~& 0.125 & 0.125 &     20.5289 &               28.52 &               51.98 &           50.91 &                   26.96 &                            0.45 &     23.64 & 30.41 \\
  \toprule
  \toprule
  Mixtral-8x7B & Full Attention &     - &     - &      3.5967 &               84.01 &               48.53 &           76.32 &                   59.73 &                           58.38 &     67.90 & 65.81 \\
  \midrule
  \multirow{9}{*}{Mixtral-8x7B}
      & \method~&   0.5 &   0.5 &      3.5970 &               83.24 &               47.32 &           74.27 &                   58.53 &                           56.48 &     67.23 & 64.51 \\
      & \method~&   0.5 &  0.25 &      3.6196 &               81.71 &               43.51 &           69.61 &                   53.67 &                           55.57 &     63.92 & 61.33 \\
      & \method~&   0.5 & 0.125 &      3.6635 &               76.18 &               41.63 &           61.72 &                   47.78 &                           49.28 &     58.94 & 55.92 \\
      & \method~&  0.25 &   0.5 &      3.6004 &               79.99 &               46.47 &           61.64 &                   49.15 &                           57.85 &     63.04 & 59.69 \\
      & \method~&  0.25 &  0.25 &      3.7906 &               71.58 &               37.77 &           53.28 &                   37.54 &                           37.38 &     46.66 & 47.37 \\
      & \method~&  0.25 & 0.125 &      4.2566 &               59.23 &               36.58 &           50.75 &                   28.67 &                           15.39 &     32.28 & 37.15 \\
      & \method~& 0.125 &   0.5 &      3.6358 &               72.29 &               45.28 &           50.67 &                   33.70 &                           55.50 &     47.15 & 50.76 \\
      & \method~& 0.125 &  0.25 &      4.5500 &               52.16 &               37.86 &           46.57 &                   23.98 &                           17.13 &     27.02 & 34.12 \\
      & \method~& 0.125 & 0.125 &      5.5250 &               46.93 &               40.33 &           49.72 &                   23.55 &                            0.91 &     24.78 & 31.04 \\
  \toprule
  \end{tabular}%
  }
  \label{table:comprehensive_results_postrotary}
\end{table}

In this section, we provide a detailed evaluation of our method across a wide
range of models and tasks. Figure \ref{fig:appendix_pcatopk} illustrates the
performance of \method~on perplexity and downstream tasks compared to the full
attention baseline. We present results for both pre-rotary and post-rotary
PCA transformations. The models evaluated include Llama2-7B,
Llama2-13B, Llama2-70B, Llama3-8B, Llama3-70B, TinyLlama-1.1B, Mistral-7B, and
Mixtral-8x7B. We assess the models using various configurations of $k$ and $d$
for \method.

As $k_f$ and $d_f$ decrease, the model’s performance deteriorates, particularly
when both are set to 0.125. Notably, the impact of $k_f$ on performance is more
pronounced than that of $d_f$. This is evident as $k_f=0.125$ and $d_f=0.5$
significantly underperform compared to $k_f=0.5$ and $d_f=0.125$ across nearly
all models. The configurations of $k_f=0.25$ and $d_f=0.25$, along with
$k_f=0.125$ and $d_f=0.5$, demonstrate relatively strong performance across all
models, striking a favorable balance between performance and accuracy, with a
theoretical speedup of 2.6x for both configurations. Settings with $k_f=0.5$
maintain model quality much more effectively but do not yield a significant
empirical speedup.

Tables \ref{table:comprehensive_results_prerotary} and
\ref{table:comprehensive_results_postrotary} present finer-grained 
results for each task and model.

\subsection{Variable Dimensionality Analysis}
\label{app:variable_d}

\begin{figure}[ht]
  \centering
  \includegraphics[width=0.32\linewidth]{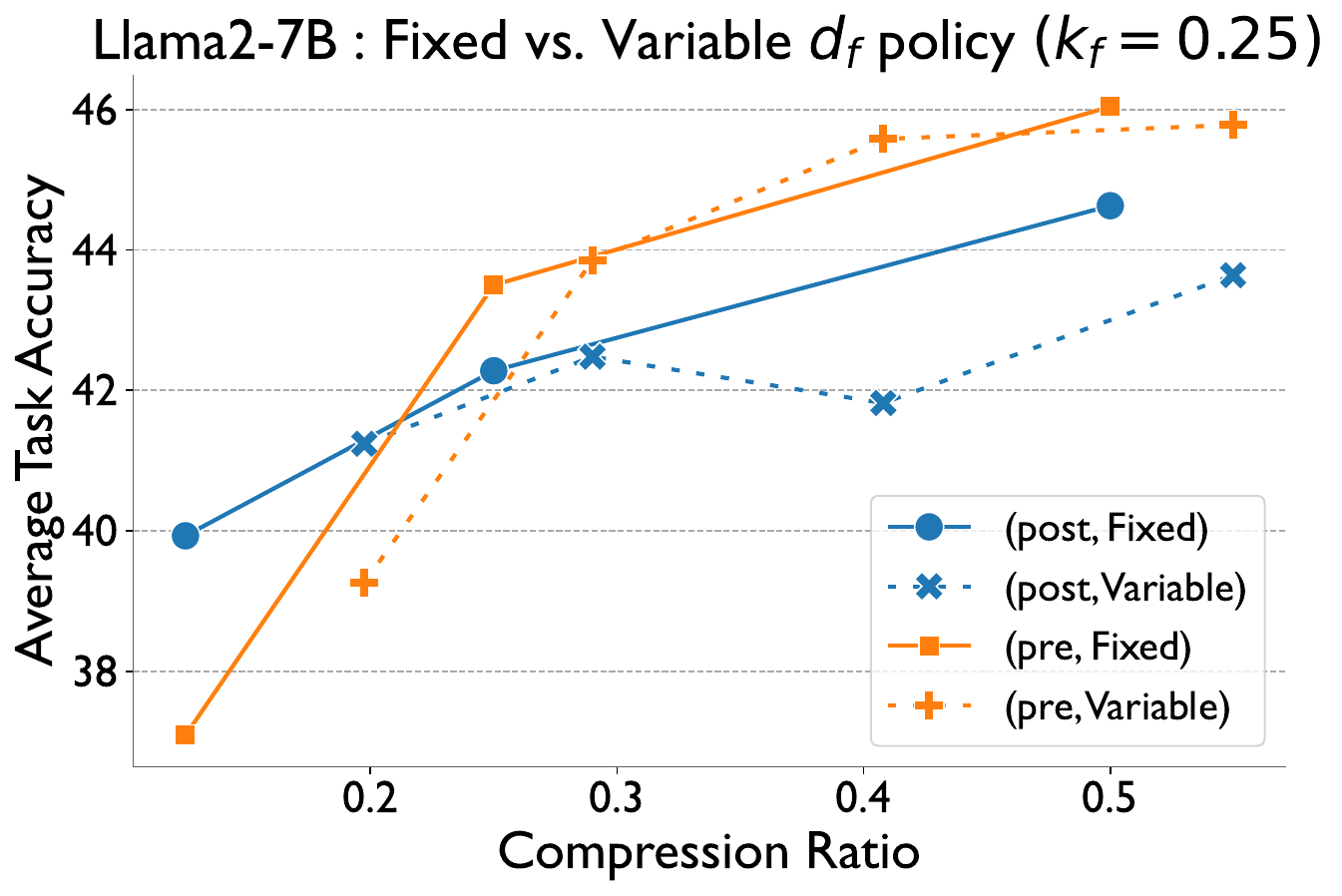}
  \includegraphics[width=0.32\linewidth]{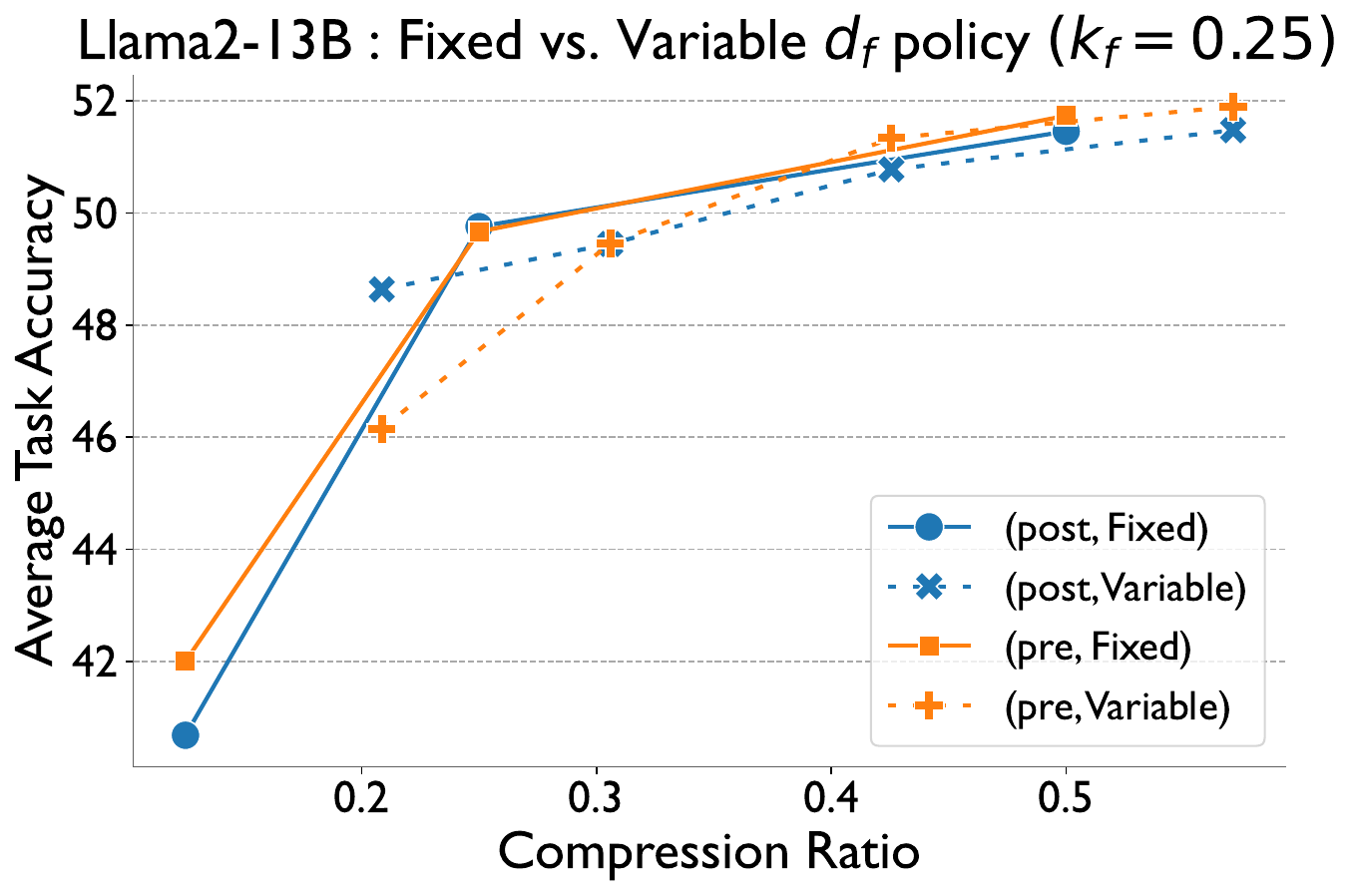}
  \includegraphics[width=0.32\linewidth]{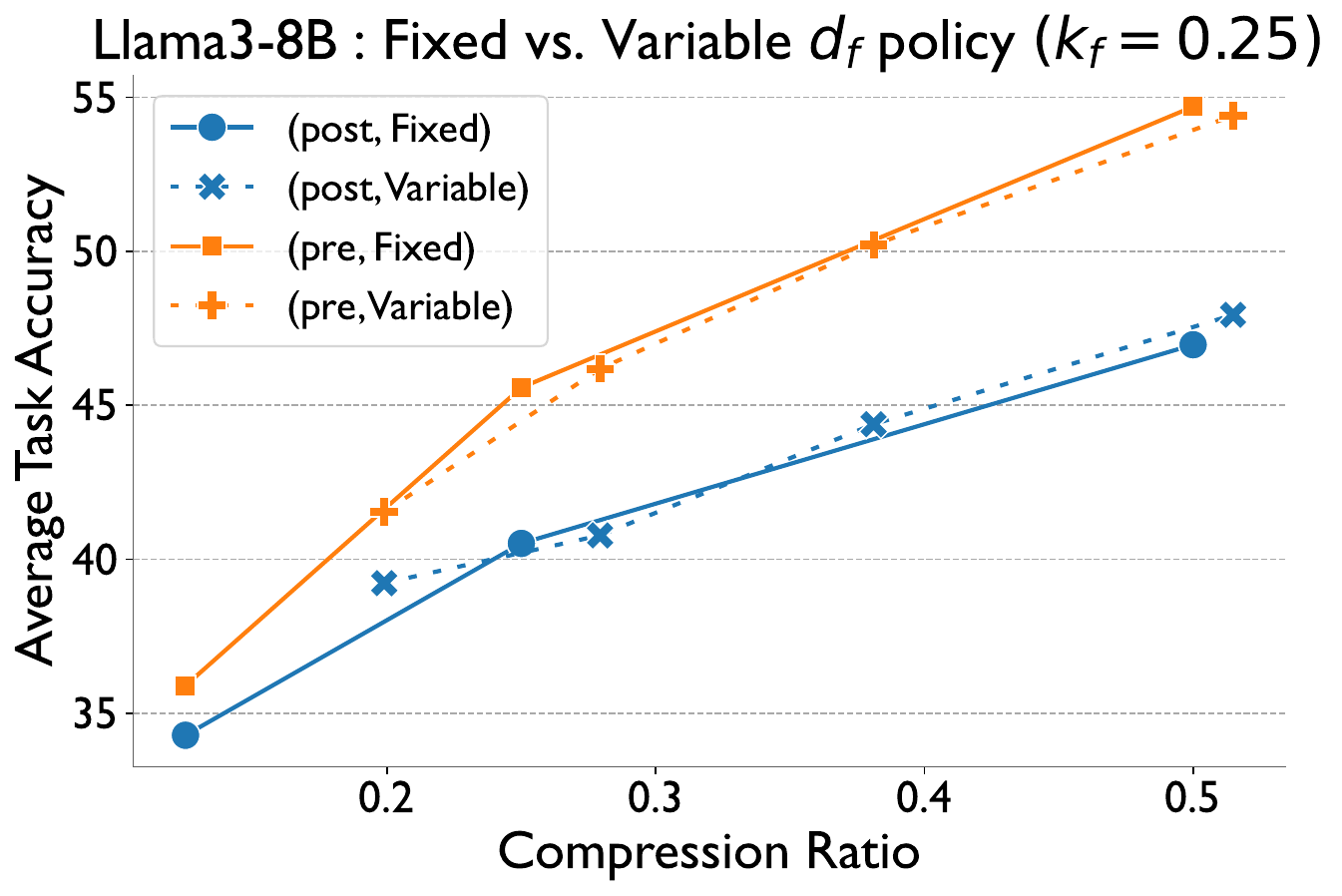}
  \caption{Average short-context task accuracies using fixed $d_f$ vs. varying $d_f$
  values across the layers for Llama2-7B (left), Llama2-13B (middle) and
  Llama3-8B (right). For the variable policy, $d_f$ is set based on per-layer
  explained variance (varied from 0.5 to 0.8). Compression ratio is the average
  $d_f/D$ across layers.}
  \label{fig:variable_d}
\end{figure}

In this section, we present our experiments utilizing a variable $d_f$ policy per
layer in \method. For this experiment, we set $d_f$ based on the per-layer
explained variance, varying from 0.5 to 0.8, and plotted the average
short-context task accuracies (refer to Section \ref{sec:setup}) against the
compression ratio, calculated as follows:
\begin{equation}
  \text{Compression Ratio} = \frac{\sum_{l=1}^{L} d_f^l}{D}
\end{equation}

Figure \ref{fig:variable_d} presents these plots for the Llama2-7B, Llama2-13B,
and Llama3-8B models. We observe that the variable $d_f$ policy does not yield
significant improvements over the fixed $d_f$ policy used in our experiments. It
is possible that different layers may require distinct explained variance
thresholds for optimal performance, indicating that further tuning is necessary.
Nevertheless, the simplicity of the fixed $d_f$ policy, combined with its
comparable performance to the variable $d_f$ policy, makes it a practical choice
for \method.

\newpage

\section{Comparison of our kernels with SparQ}
\label{append:comp_pcatopk}

\begin{figure}[ht]
    \centering
    \includegraphics[width=0.45\linewidth]{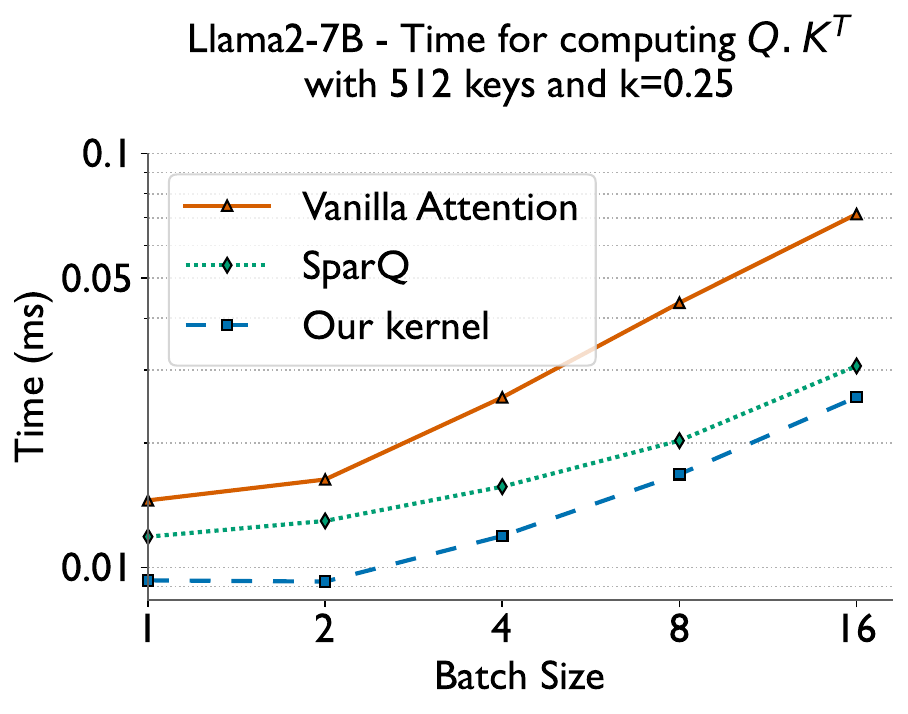}
    \includegraphics[width=0.45\linewidth]{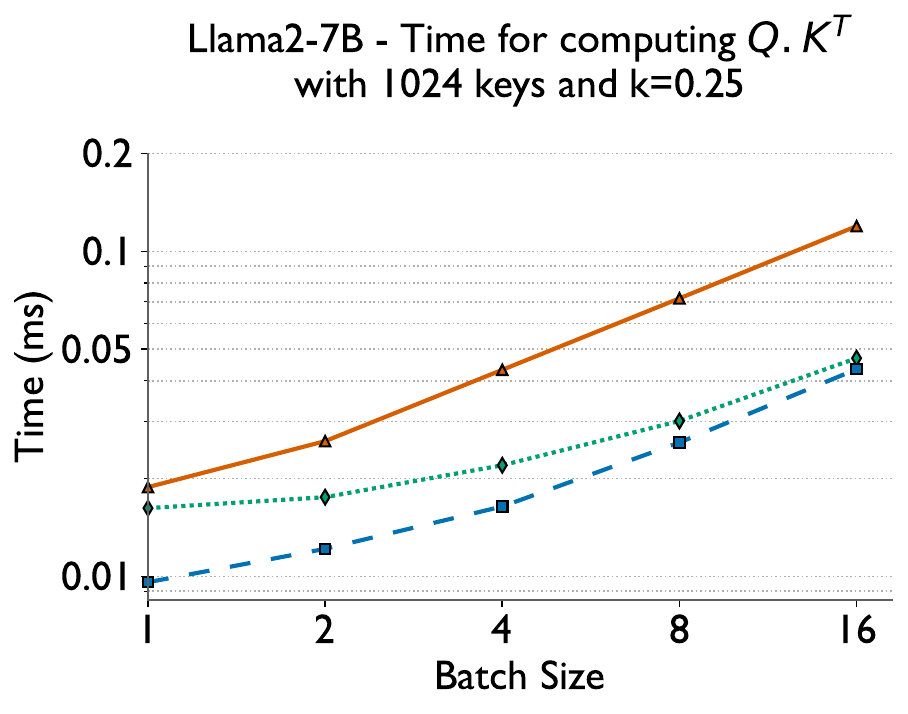}
    \includegraphics[width=0.45\linewidth]{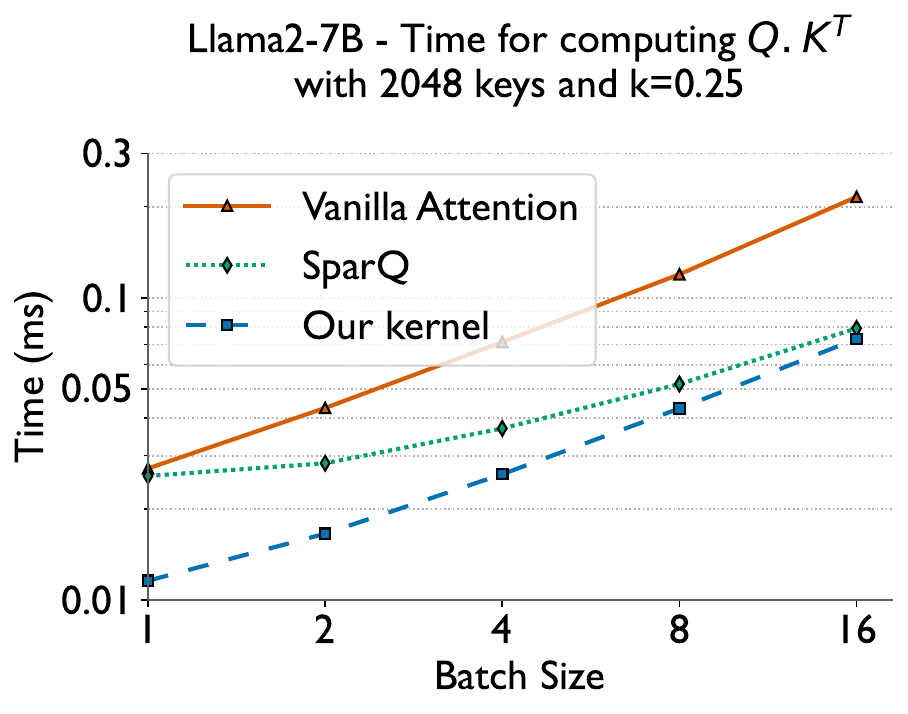}
    \includegraphics[width=0.45\linewidth]{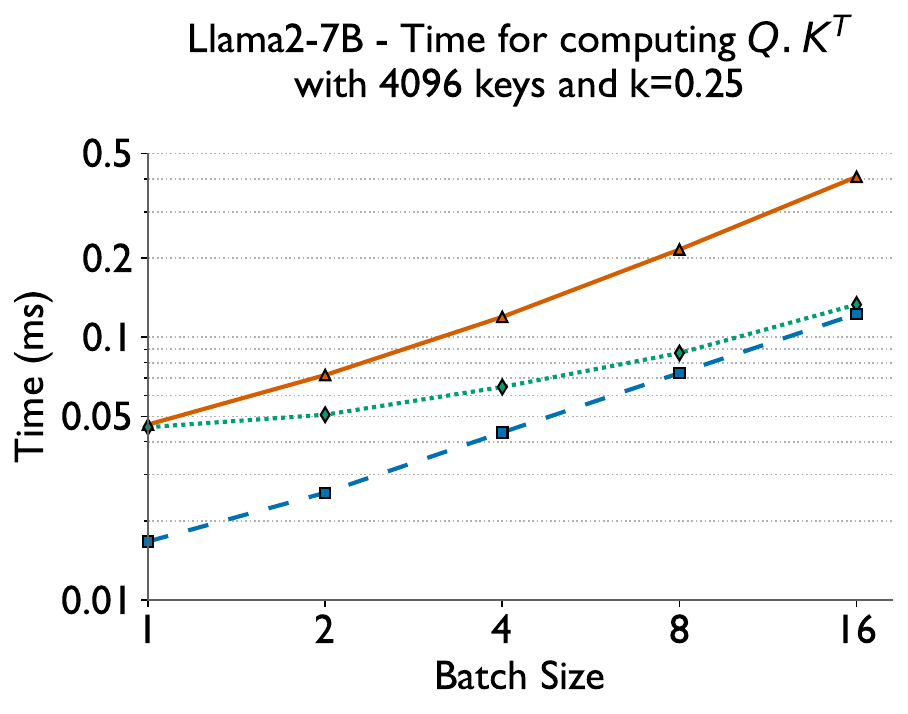}
    \caption{Comparing the performance of our proposed kernel for computing $Q.K^T$, with SparQ's~\citep{ribar2023sparq} 
    kernel for various batch sizes and number of keys in the KV-cache.  }
    \label{fig:sparq-compare}
  \end{figure}

As mentioned in Section~\ref{sec:kernel}, we create optimized kernels in Triton to efficiently compute the three 
matrix multiplications in \method~(lines 5, 8, and 9 of Algorithm~\ref{alg:topkpca}) without creating temporary 
dense copies of subsets of the KV-cache. Initially, we planned to use the implementations developed by the authors 
of SparQ~\citep{ribar2023sparq}. However, we discovered two major issues with their kernels. Let's say you are multiplying 
two matrices of sizes $m \times k$ and $k \times n$, then SparQ kernels parallelize compute along only the m dimension. However, 
it is well known that one can parallelize matrix multiplications along the n dimension as well and gain more performance. 
Thus, we add this extra dimension of parallelism to their triton kernel. Second, their kernels cannot handle non-powers of 2 
number of tokens in the KV-cache, a setting which is commonly encountered in inference since we generated keys and values 
one at a time. Therefore, we extend their kernels to handle non-powers of two number of tokens in the KV-cache successfully. 
In Figure~\ref{fig:sparq-compare}, we compare the performance of our kernel with SparQand vanilla PyTorch based attention for 
an attention layer in Llama2-7B for various sizes of the KV-cache ranging from 512 to 4096. We do this for the matmul operation 
of query and keys with top-$k$ as 0.25.

We see very high speedups over SparQ for small batch sizes. For instance, for a batch size of 1 with 4096 keys (bottom right),
our kernel is faster than SparQ by nearly $2.8 \times$! Infact, the SparQ kernel barely obtains any speedup over vanilla PyTorch 
even though it is only using 25\% of the keys (1024 out of 4096). This is because SparQ only parallelizes compute across the m-dimension as 
we discussed before. In attention the m-dimension is proportional to the batch size, and thus having a small batch size limits 
the amount of parallelism SparQ kernels can exploit. Whereas, given our 2D parallel approach to matrix multiplication, we can 
exploit parallelism along the sequence dimension and thus still obtain significant speedups over vanilla PyTorch. Note that 
smaller batch sizes are very common in inference and hence it is extremely important to design a kernel that can function 
efficiently even at smaller batch sizes. Our speedups over SparQ decrease with larger batch sizes, but overall we are always faster 
than both SparQ and vanilla PyTorch.

\newpage 

\section{Generation Examples}
\label{appendix:generation}
In this section, we look at a generation example from the Llama2-7B (Chat) model given a custom 
prompt for different settings of $k$ and $d$.

\prompt{ In a world where technology has advanced beyond imagination, society
grapples with the consequences of its own creations. The integration of
artificial intelligence into everyday life has revolutionized how we live, work,
and interact. However, with great power comes great responsibility, and ethical
dilemmas abound. Governments struggle to regulate the use of AI, while
corporations push the boundaries of what is possible in the pursuit of profit.
Meanwhile, individuals navigate a landscape where privacy is increasingly
scarce, and the line between human and machine blurs. Against this backdrop, a
new generation of thinkers emerges, questioning the very nature of consciousness
and what it means to be human. As tensions rise and alliances shift, the fate of
humanity hangs in the balance, waiting to be written by those bold enough to
seize the pen of destiny.}

\response{Will we create a future where AI and humanity coexist
in harmony, or will the pursuit of power and profit lead to our downfall? The
choice is ours, and the clock is ticking.}{Llama2-7B Chat Full}

\response{Will we chart a path towards utopia or succumb to the
dangers of unchecked progress? Can we find a balance between technological
advancement and preserving the essence of our humanity? The future is ours to
create, and the choice is ours to make. The clock ticks on, waiting for us to
take the next step forward into an uncertain tomorrow, where the only constant
is the endless possibilities of an age of AI}{Llama2-7B Chat - $k=50\%$, $d=25\%$}

\response{Will you answer the call to shape the future of AI?
Will you play a part in shaping the course of human history? The possibilities
are endless, and the outcome uncertain. Are you ready to embark on a journey
that could change the world forever?}{Llama2-7B Chat - $k=25\%$, $d=25\%$}
\newpage

\section{PCAAttn: Directly Leveraging Reduced-Dimensional Attention Scores}
\label{appendix:PCAAttn}

\begin{algorithm}
  \caption{PCAAttn}
  \label{alg:PCAAttn}
  \begin{algorithmic}[1]
  \Require At the $i^{th}$ step - Input: $\x_S \in
  \mathbb{R}^{1 \times D}$, KV-cache: $\Khat_{:S-1,:d}, \V_{:S-1} \in \mathbb{R}^{(S - 1)
  \times d}$, Projection Matrix: $\bP_{:d} \in \mathbb{R}^{D \times d}$ (first $d$ principal components)
  \Function {PCA-Attention}{$\x_S, \Khat_{:S-1,:d}, \V_{i-1}, \bP_d$} 
  \State $\q_S, \bk_S, \bv_S \gets computeQKV(\x_S)$ 
  \State $\qhat_{S,:d} \gets \q_S\bP_{:d}$, $\bkhat_{S,:d} \gets \bk_S\bP_{:d}$ 
  \State $\Khat_{:S,:d} \gets concat(\Khat_{:S-1,:d}, \bkhat_{S,:d})$ 
  \State $\V_{:S} \gets concat(\V_{:S-1}, \bv_S)$ 
  \State $\ba = softmax(\frac{\qhat_{S,:d}(\Khat_{:S,:d})^{T}}{\sqrt{D}})$ 
  \State \Return $\ba\V_{:S}$
  \EndFunction
  \end{algorithmic}
\end{algorithm}

One other approach we tried is to directly use the formulation in \ref{lemma:projection}
to compute the final attention scores. More specifically, we compute the PCA
transformed query and key vectors, projected onto the first $d$ principal
components, and then compute the attention scores. We only store the reduced
dimension key vectors in the KV-cache. We call this method PCAAttn (Algorithm \ref{alg:PCAAttn}).

\noindent \textbf{Compute and Memory Analysis:} When computing attention between
a single query $\q_S \in \mathbb{R}^{1 \times D}$ and the key vectors $\K_{:S} \in
\mathbb{R}^{S \times D}$, the matrix multiplication $\q_S\K_{:S}^{T}$ has a
complexity of $\bigO(DS)$. Using PCAAttn, the key and query vectors are reduced
to $d$ dimensions and the complexity of the matrix multiplication is reduced to
$\bigO(dS)$. Thus, we can get a speedup of $D/d$ in the attention dot product
computation. The PCA transformation of the query and key vector generated at each
step has a complexity of $\bigO(D^2)$, which is small when $S>>D$. The KV-cache
memory requirement is reduced by a factor of $0.5*D/d$ because we only reduce
the key vectors to $d$ dimensions and not the values. Additionally, the PCA adds 
a significantly small memory overhead of $\bigO(Dd)$. 

\noindent \textbf{Experimental Results:} 

\begin{table}[h!]
  \caption{Performance of PCAAttn with various cache configurations.}
  \centering
  \resizebox{0.97\linewidth}{!}{%
  \small 
  \begin{tabular}{@{}lccccccccccc@{}}
  \toprule
  Model & Method & $k_f$ & $d_f$ & Perplexity$\downarrow$ & Hellaswag$\uparrow$ & Winogrande $\uparrow$ & MathQA $\uparrow$ & OpenbookQA $\uparrow$ & 
  RTE $\uparrow$ & COPA $\uparrow$ \\
  \toprule
  \toprule
  Llama2-7B & Full Attention & - & - & 5.1102 & 57.2 & 69.1 & 28.4 & 31.4 & 62.8 & 87.0\\
  \midrule 
  \multirow{3}{*}{Llama2-7B}
  & Exact TopK & 0.5 & - & 5.1191 & 57.2	& 68.9	& 28.3	& 31.2	& 63.9	& 86.0\\
  & H$_2$O & 0.5 & - & 5.1456 & 55.5	& 61.8	& 24.4	& 27.4	& 62.8 & 77.0\\
  & PCAAttn & - & 0.5 & 38.3997 & 33.3 & 53.2 & 21.7 & 14.2 & 50.5 & 73\\
  \midrule 
  \multirow{3}{*}{Llama2-7B}
  & Exact TopK & 0.25 & - & 5.1799 & 56.9	& 68.6	& 29.4	& 29	& 66.4 & 76.0\\
  & H$_2$O & 0.25 & - & 5.2809 & 50.1	& 51.6	& 21.1	& 17.8	& 55.2 & 55.0\\
  & PCAAttn & - & 0.25 & 243.2631 & 26.9 & 48.5 & 20.5 & 11.4 & 49.1 & 65.0\\
  \toprule
  \toprule
  Mistral-7B 
  & Full Attention & - & - & 4.9140 & 61.2 & 73.9 & 35.7 & 32.2 & 66.8 & 91.0\\
  \midrule 
  \multirow{3}{*}{Mistral-7B}
  & Exact TopK & 0.5 & - & 4.9143 & 61.1	& 73.8	& 35.6	& 32.6	& 65.3	& 92.0\\
  & H$_2$O & 0.5 & - & 4.9560 & 59.4	& 58.6	& 26.4	& 23.0	& 62.4 & 71.0\\
  & PCAAttn & - & 0.5 & 396.8967 & 31.4 & 50.4 & 22.5 & 15.6 & 53.4 & 72.0 \\
  \midrule 
  \multirow{3}{*}{Mistral-7B}
  & Exact TopK & 0.25 & - & 4.9170 & 60.4	& 73.0	& 35.4	& 30.0 & 65.3 & 85.0\\  
  & H$_2$O & 0.25 & - & 5.0805 & 52.7	& 49.7	& 21.9	& 17.4	& 52.0 & 56.0 \\
  & PCAAttn & - & 0.25 & 933.6016 & 27.2 & 52.2 & 21.6 & 13.6 & 53.0 & 63.0 \\
  \bottomrule
  \end{tabular}%
  }
  \label{table:model_performance_pcaattn}
  \end{table}

Table \ref{table:model_performance_pcaattn} shows the performance of PCAAttn on
Llama2-7B and Mistral-7B models. We can see that our PCAAttn method performs
poorly compared to all the baselines and the H$_2$O method for all cache
configurations. We believe that this happens because the application of rotary
embeddings increases the dimensionality of the key vectors and using reduced
dimensionality to store the keys results in loss of information. To further
investigate this, we look back at Figure \ref{fig:heatmap} which shows the rank
at 90\% explained variance for the key vectors across all layers and heads. Even
though, the average rank per layer is around 50\% of the full dimensionality,
the rank for some layers and especially some heads within each layer is much
higher. Due to the poor performance of PCAAttn, we do not include it in the
final results and decide to focus on \method~instead in the main paper. Note, that 
we only tried the post-rotary transformations in PCAAttn, and it is possible that
pre-rotary transformations might perform better, but we leave this for future work.

\newpage

\section{Estimate of Compute Resources Required to Replicate our Experiments}
\label{append:estimate-compute}

As mentioned in Section~\ref{sec:results}, we conduct all of our experiments on Perlmutter, a 
multi-GPU cluster with 4 A100 GPUs per node. Since we do not do any training/fine-tuning, our experiments can 
be done on a very small number of GPUs. For instance, all of our runs involving models with 7B and 13B parameters were 
done on a single A100 GPU. For models larger than this (like LLama2-70B, Llama3-70B), we had to resort to running on four 
A100 GPUs (or a single node)  with tensor parallelism using the AxoNN parallel deep learning framework. All results for 7B and 
13B sized models can be compiled within 3 hours. For larger models like the 70B Llama-2 and 3 as well as Mixtral models, the 
total times for computing all results are in the ballpark of 10 hours. Our compute benchmarking runs of Llama-13B are very short 
and can be completed within 5 minutes.

%
%


\end{document}